\documentclass[sigconf]{acmart}
\acmSubmissionID{5591}

\usepackage{multirow}
\usepackage{multicol}
\usepackage{bbding}
\usepackage{enumitem}

\AtBeginDocument{%
  }

\setcopyright{acmlicensed}
\copyrightyear{2025}
\acmYear{2025}
\acmDOI{10.1145/3746027.3755727}
\acmConference[MM '25]{Proceedings of the 33rd ACM International Conference on Multimedia}{October 27--31, 2025}{Dublin, Ireland}
\acmBooktitle{Proceedings of the 33rd ACM International Conference on Multimedia (MM '25), October 27--31, 2025, Dublin, Ireland}
\acmISBN{979-8-4007-2035-2/2025/10}

\begin{document}

\title{Multi-Task Label Discovery via Hierarchical Task Tokens for Partially Annotated Dense Predictions}

\author{Jingdong Zhang}
\orcid{0009-0008-6668-1140}
\affiliation{%
  \institution{Texas A\&M University}
  \city{College Station}
  \state{Texas}
  \country{USA}
}
\email{jdzhang@tamu.edu}

\author{Hanrong Ye}
\orcid{0000-0002-7986-6143}
\affiliation{%
  \institution{Hong Kong University of Science and Technology}
  \city{Hong Kong}
  \country{China}
}
\email{hanrong.ye@connect.ust.hk}

\author{Xin Li}
\orcid{0000-0002-0144-9489}
\affiliation{%
  \institution{Texas A\&M University}
  \city{College Station}
  \state{Texas}
  \country{USA}
}
\email{xinli@tamu.edu}

\author{Wenping Wang}
\authornote{Corresponding author.}
\orcid{0000-0002-2284-3952}
\affiliation{%
  \institution{Texas A\&M University}
  \city{College Station}
  \state{Texas}
  \country{USA}
}
\email{wenping@tamu.edu}

\author{Dan Xu}
\orcid{0000-0003-0136-9603}
\affiliation{%
  \institution{Hong Kong University of Science and Technology}
  \city{Hong Kong}
  \country{China}
}
\email{danxu@cse.ust.hk}

\renewcommand{\shortauthors}{Zhang et al.}

\begin{abstract}
In recent years, simultaneous learning of multiple dense prediction tasks with partially annotated label data has emerged as an important research area. Previous works primarily focus on leveraging cross-task relations or conducting adversarial training for extra regularization, which achieve promising performance improvements, while still suffering from the lack of direct pixel-wise supervision and extra training of heavy mapping networks. To effectively tackle this challenge, we propose a novel approach to optimize a set of compact learnable hierarchical task tokens, including global and fine-grained ones, to discover consistent pixel-wise supervision signals in both feature and prediction levels. Specifically, the global task tokens are designed for effective cross-task feature interactions in a global context. Then, a group of fine-grained task-specific spatial tokens for each task is learned from the corresponding global task tokens. It is embedded to have dense interactions with each task-specific feature map. The learned global and local fine-grained task tokens are further used to discover pseudo task-specific dense labels at different levels of granularity, and they can be utilized to directly supervise the learning of the multi-task dense prediction framework. Extensive experimental results on challenging NYUD-v2, Cityscapes, and PASCAL Context datasets demonstrate significant improvements over existing state-of-the-art methods for partially annotated multi-task dense prediction.
\end{abstract}





\maketitle

\section{Introduction}
\label{sec:intro}

With the rapid development of supervised learning with deep neural networks, various pixel-wise dense prediction tasks with highly complementary properties such as semantic segmentation and depth estimation have achieved great success in multi-task learning (MTL) in recent years~\cite{misra2016cross,vandenhende2020mti,xu2018pad,ye2022inverted,zhang2018joint}. Researchers pursue the learning of them simultaneously in a unified framework, which can effectively model cross-task correlations and achieve superior results in terms of model training costs and performances.

\par
\begin{figure*}[t]
  \centering
  \includegraphics[width=0.95\linewidth]{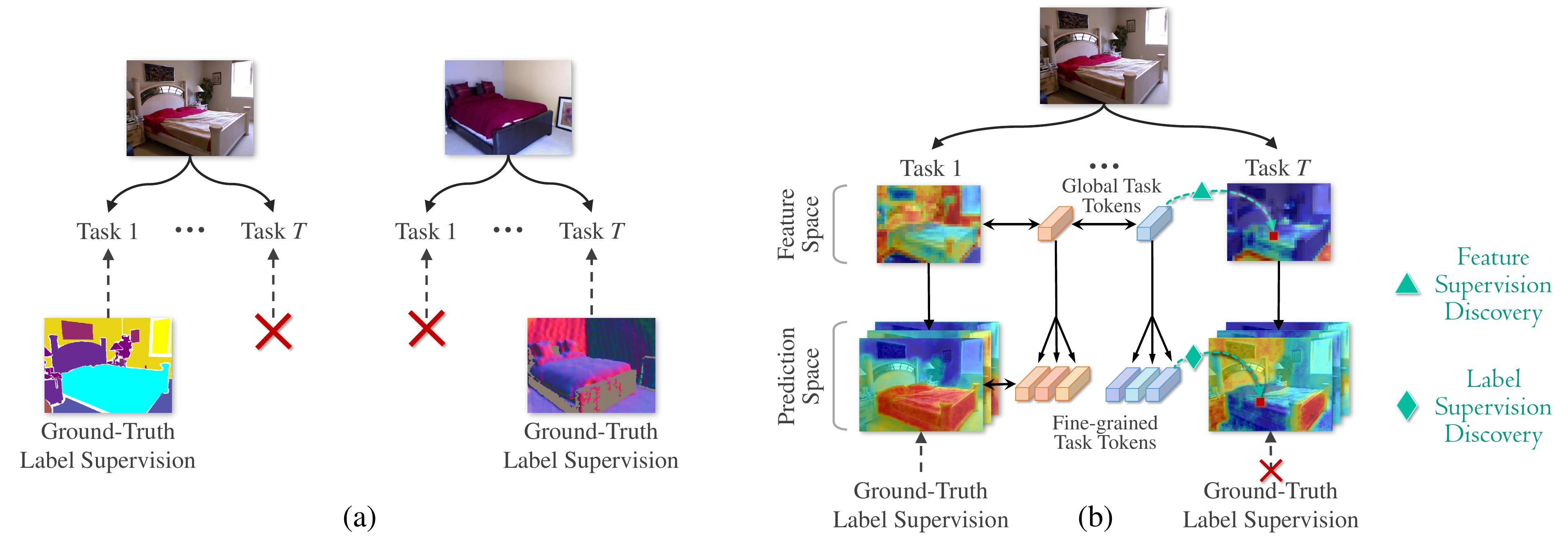}
  \vspace{-3mm}
  \caption{\textbf{(a)} Illustration of partially annotated multi-task dense prediction setting. Each input image only has partial task labels from all the tasks. \textbf{(b)} Illustration of the learning of Hierarchical Task Tokens, including global task tokens and fine-grained task tokens, by conducting feature-token interactions in feature and prediction spaces separately. The well-learned hierarchical task tokens can achieve both feature supervision discovery and task label discovery.
  }
  \vspace{-3mm}
  \label{fig:1}
\end{figure*}

 However, in real-world scenarios, obtaining pixel-level annotations is prohibitively expensive, especially when dealing with a set of distinct dense prediction tasks. Each image has to be annotated with pixel labels for all the tasks. Thus, existing works have delved into the problem of multi-task learning with only partially annotated dense labels~\cite{li2022learning,luo2021semi,wang2022semi,zamir2020robust,zeng2019joint,nishi2024joint,ye2024diffusionmtl}. 
Specifically, as illustrated in Fig.~\ref{fig:1} (a), given an input image, for $T$ dense prediction tasks, the task labels are provided partially, i.e. for at least one task and at most $T-1$ tasks. Learning a multi-task model under this setting is particularly challenging since every input image lacks some of the task supervision signals, and the performance typically drops significantly if compared to the same model trained with full task label supervisions~\cite{li2022learning}.

\par Previous works have been focusing on excavating cross-task relations by training heavy extra mapping networks~\cite{li2022learning,nishi2024joint}, however, simply applying regularization in compact latent spaces fails to address the lack of dense pixel supervision and limits the performance. On the contrary, directly discovering pseudo task labels in prediction spaces can alleviate this problem to a certain extent, while still suffering from the following two severe limitations: (i) Simply discovering task labels in the prediction spaces separately ignores the highly relevant task relations, thus leading to trivial performance under multitask scenarios. (ii) Solely discovering labels in prediction spaces cannot take advantage of the abundant task representations in feature space. For each specific task, the distributions of the task features and predictions should be consistent. Since the encoder parameters are shared among tasks, thus the produced rich task-generic features are beneficial for building cross-task relations and discovering supervision signals effectively. Therefore, it is critically important to involve hierarchical cross-task representations from feature space to prediction space to consistently boost the task label discovery process.

To effectively tackle the aforementioned challenges, we propose a novel approach that performs task label discovery from both the feature and prediction spaces via an effective design of learnable Hierarchical Task Tokens (HiTTs). HiTTs are sets of compact parameters that are learned in a hierarchical manner to model global inter-task relationships and local fine-grained intra-task relationships, which allows for discovering pixel-wise task pseudo labels straightforwardly in both feature and prediction spaces consistently. More specifically, as depicted in Fig.~\ref{fig:1} (b), we apply HiTTs during the multi-task decoding stage and jointly optimize them with the multi-task learning network. The HiTTs consist of two hierarchies. The first hierarchy is a set of global task tokens. 
The global task tokens are randomly initialized and can perform cross-task feature-token interactions with different task feature maps based on self-attention. 
These learned task tokens can be used to discover feature-level pseudo supervision by selecting highly activated pixel features correlated to each task. The second hierarchy is the fine-grained task tokens. These tokens are directly derived from the global task tokens with learnable projection layers to inherent beneficial task representations. They are subsequently utilized to perform interactions within each task-specific feature map at a finer granularity. As the fine-grained task tokens can learn pixel-to-pixel correlation with each task feature map, they thus benefit the discovery of dense spatial labels for each task. Compared with naive pseudo labeling process~\cite{lee2013pseudo}, the HiTTs can bridge effective information from supervised tasks so as to encourage highly confident predictions on unsupervised tasks. We learn both hierarchies of tokens simultaneously in an end-to-end manner incorporating the multi-task baseline network, and exploit both levels of supervision signals discovered from the two hierarchies, for optimizing multi-task dense predictions on partially annotated datasets.

\par In summary, the contribution of this work is three-fold:
\begin{itemize}[leftmargin=*, topsep=0pt, partopsep=0pt, itemsep=0pt, parsep=0pt]
\setlength{\itemsep}{0pt}
\setlength{\parskip}{2pt}
    \item Instead of discovering cross-task regularization by extra heavy mapping networks, we propose to utilize cross-task relations for high-quality task label discovery, which serves as pixel-level dense pseudo supervision under the multi-task partially supervised setting.
    \item We propose a novel design of Hierarchical Task Tokens (HiTTs), which can learn hierarchical multi-task representations for high-quality pseudo label discovery consistently in both the feature and the prediction levels.
    \item Our proposed method significantly outperforms existing state-of-the-art competitors on multi-task partially annotated benchmarks, including PASCAL-Context, NYUD-v2 and Cityscapes, and demonstrates clear effectiveness on challenging dense prediction tasks with limited annotations, including segmentation, depth estimation, normal estimation and edge detection, etc. Code is released at \url{https://github.com/Evergreen0929/EEMTL}.
\end{itemize}

\begin{figure*}[t]
  \centering
  \includegraphics[width=0.99\linewidth]{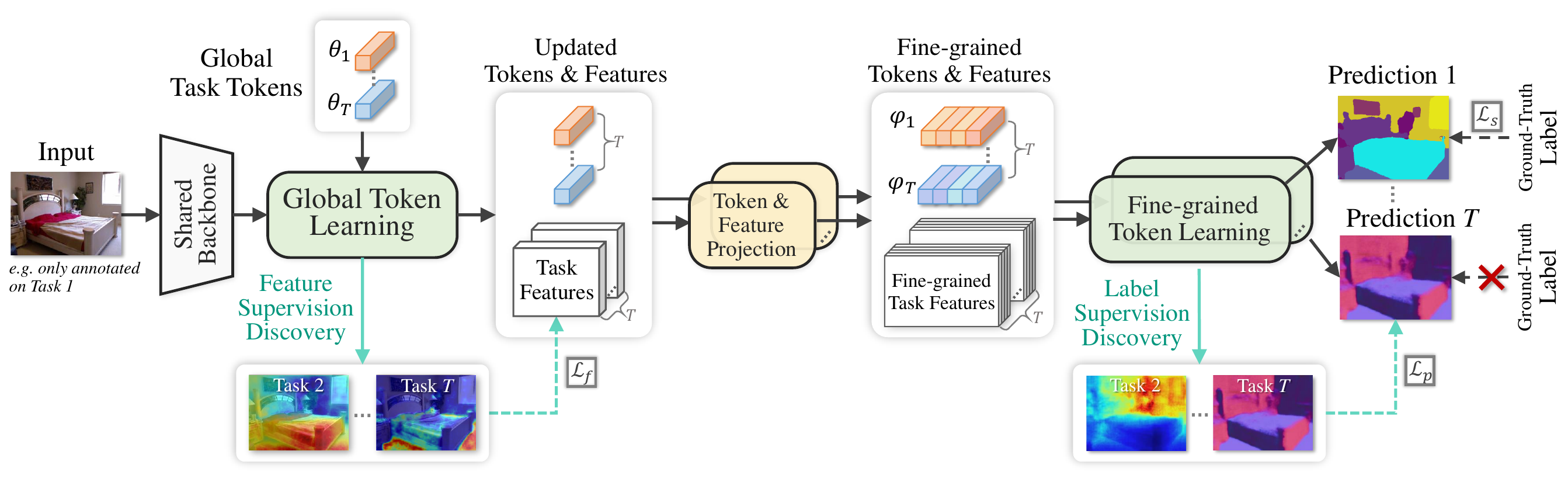}
  \vspace{-5mm}
  \caption{Illustration of our method. HiTTs consist of both global and fine-grained task tokens which learn discriminative task representations by conducting feature-token interactions with attentions in corresponding multi-task decoding stages. The global task tokens $\boldsymbol{\theta_i}$ discover feature-level pseudo supervision $\mathcal{L}_f$, while the fine-grained task tokens $\boldsymbol{\varphi_i}$ inherit the knowledge from global task tokens and directly discover pixel labels for supervision $\mathcal{L}_p$. The supervision from ground-truth label is denoted as $\mathcal{L}_s$.}
  \label{fig:main}
  \vspace{-4mm}
\end{figure*}

\section{Related Work}
\label{sec:related}
\par\noindent\textbf{Multi-task Dense Prediction.} Dense prediction tasks aim to produce pixel-wise predictions for each image. Common tasks including semantic segmentation, depth estimation, and surface normal estimation exhibit high cross-task correlations. For instance, depth discontinuity is usually aligned with semantic boundaries~\cite{vandenhende2020mti}, and surface normal distributions are aligned with spatial derivatives of the depth maps~\cite{lu2021taskology}. Thus, a number of works have been focusing on multi-task dense predictions~\cite{gao2019nddr,liu2019end,misra2016cross,vandenhende2020mti,xu2018pad,yang2023contrastive,ye2022inverted,ye2023taskexpert,taskprompter2023,zhang2023rethinking,zhang2018joint,zhang2019pattern,zhang2025bridgenet,tang2025semantic,tian2024unite,lu2024swiss,caomsm,yangmulti,chavhan2025upcycling}. They leverage parameters sharing to conduct cross-task interactions by effective attention mechanisms for task feature selection~\cite{liu2019end}, and multi-modal distillation~\cite{xu2018pad}, multi-scale cross-task interactions~\cite{vandenhende2020mti}, global pixel and task interactions~\cite{ye2022inverted} and multi-task bridge features \cite{zhang2025bridgenet}. However, these works focus on fully-supervised settings. In contrast, our work addresses the challenge of insufficient supervision signals in each task.

\noindent\textbf{Semi-supervised learning.} Obtaining pseudo labels for semi-supervised learning is a popular research direction, with several deep learning works published on the topic~\cite{iscen2019label,lee2013pseudo,li2019bidirectional,shi2018transductive,sohn2020fixmatch,tarvainen2017mean,xie2020self,zou2018unsupervised,zou2019confidence}. Among them,~\cite{lee2013pseudo} aims at picking up the class which has the maximum predicted confidence. The graph-based label propagation method~\cite{iscen2019label} is also used to infer pseudo-labels for unlabeled data.~\cite{shi2018transductive} provides a confidence level for each unlabeled sample to reduce
influences from outliers and uncertain samples, and uses MMF regularization at feature levels to make images with the same label close to each other in the feature space.~\cite{xie2020self} uses accurate pseudo labels produced by the teacher model on clean unlabeled data to train the student model with noise injected. For semi-supervised dense prediction tasks~\cite{pieropan2022dense,yang2023revisiting,hoyer2024semivl,mai2024rankmatch,ran2024pseudo,yang2025unimatch}, such as semantic segmentation, several works focus on assigning pixel-wise pseudo annotations from high-confidence predictions~\cite{li2019bidirectional,zou2018unsupervised,zou2019confidence}. However, these works target single-task learning setups. {Despite pseudo labeling, discovering consistency for regularization is also a popular direction for unlabeled data~\cite{li2022learning,luo2021semi,tang2022towards,zeng2019joint}. \cite{li2022learning,luo2021semi,zeng2019joint} focus on building cross-task consistency, while ~\cite{tang2022towards} uses image-level feature similarities to find important samples for semi-supervised learning. Differently, our work targets \textit{pixel-level} task supervision discovery by hierarchical task tokens containing multi-level multi-task representations for partially annotated dense predictions.

\noindent\textbf{Multi-task Partially Supervised Learning.} As discussed in the introduction, obtaining pixel-level annotations for every task on images is prohibitively expensive. Therefore, some recent works focus on partially annotated settings for multi-task learning~\cite{imran2020partly,li2022learning,liu2007semi,lu2021taskology,luo2021semi,wang2022semi,ye2024diffusionmtl,zamir2020robust,zeng2019joint,nishi2024joint}. Since directly recovering labels from other tasks is an ill-posed problem~\cite{li2022learning}, enforcing consistency among tasks is usually adopted. For instance, constructing a common feature space to align predictions and impose regularization~\cite{li2022learning}, and leveraging intrinsic connections of different task pairs between predictions of different tasks on unlabeled data in a mediator dataset, when jointly learning multiple models~\cite{lu2021taskology}. Adversarial training is also adopted to align the distributions between labeled and unlabeled data by discriminators~\cite{wang2022semi}, and multi-task denoising diffusion is adopted~\cite{ye2024diffusionmtl} to address the issue of noise in initial prediction maps. To the best of our knowledge, our hierarchical task tokens for both pseudo feature supervision and task label discovery are a novel exploration of the problem, and show a clear difference from existing works.

\section{Proposed Method}
\label{sec:method}
\par Our proposed approach for learning Hierarchical Task Tokens (HiTTs) primarily comprises two stages, \textit{i.e.}, the Global Token Learning and the Fine-grained Token Learning. The overall structure of HiTTs is depicted in~Fig.~\ref{fig:main}. Firstly, in the Global Token Learning stage, the global task tokens produce task features and then learn rich task-level representations by conducting inter- and intra-task attention with all task features. The global tokens are utilized to exploit rich representations in feature space and discover feature-level pseudo supervision. Subsequently, in the fine-grained stage, we project each task feature into fine-grained feature space by simple convolution layers, and derive the fine-grained tokens from the global tokens by Multi-layer Perceptrons (MLPs), to inherit well-learned global task representations and therefore achieve consistent pseudo label discovery. To perform a uniform confidence-based pseudo label discovery for different types of dense prediction tasks, we follow~\cite{bruggemann2021exploring} to conduct discrete quantization of regression task annotations (e.g. depth estimation and normal estimation), and treat all tasks as pixel-wise classification.

\begin{figure*}[t]
  \centering
  \includegraphics[width=0.98\linewidth]{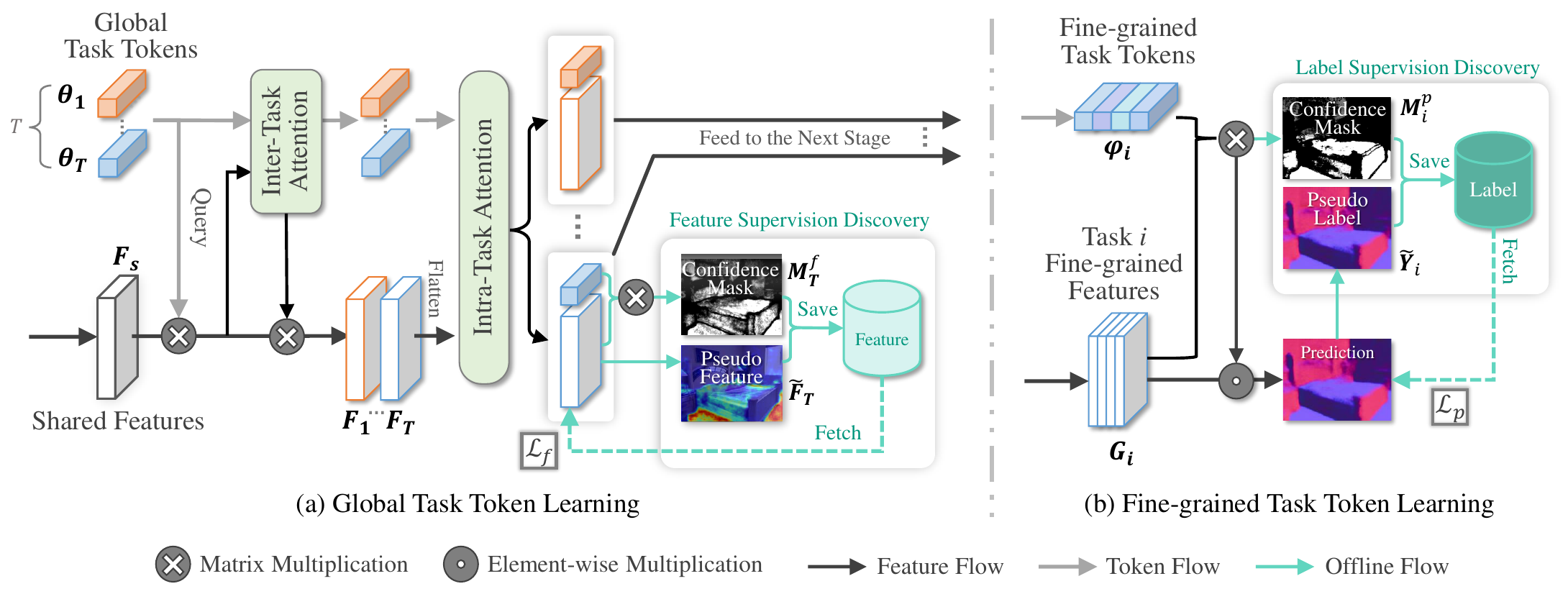}
  \vspace{-4mm}
  \caption{(a) Global Task Token Learning. It mainly contains two stages: Inter- and Intra-Task Attention for thoroughly feature-token cross-task interaction to obtain robust and representative global task tokens, and excavating pseudo feature supervision based on the learned task tokens. (b) Fine-grained Task Token Learning. The projected fine-grained tokens and feature maps (projection is shown in Fig.~\ref{fig:proj}) can be used for discovering high-quality pseudo labels in prediction spaces.}
  \vspace{-2mm}
  \label{fig:TTL}
\end{figure*}

\subsection{Global Task Token Learning}
\label{sec:GTTL}
In the global task token learning stage, we target learning global tokens representing the distributions of each task, which are further used for pseudo feature supervision discovery. The learning process is mainly achieved by inter- and intra-task attention among tokens and features, in order to exploit beneficial multi-task representations for token learning and feature supervision discovery.

Given an RGB input $\boldsymbol{X} \in \mathbb{R}^{3 \times H \times W}$, a multi-task dense prediction framework firstly produces a task-generic representation $\boldsymbol{F_{s}} \in \mathbb{R}^{C \times h \times w}$ through a shared encoder. Considering we have $T$ tasks, and we target decoding task features $\{\boldsymbol{F_{1}}, \boldsymbol{F_{2}}, \cdots, \boldsymbol{F_{T}}\}$ from $\boldsymbol{F_{s}}$ as well as learning representative global task tokens $\{\boldsymbol{\theta_{1}}, \boldsymbol{\theta_{2}}, \cdots, \boldsymbol{\theta_{T}}\}, \boldsymbol{\theta_{i}} \in \mathbb{R}^{C}$ for each task. They are randomly initialized learnable vectors serving as additional input tokens for the decoder.

As shown in Fig.~\ref{fig:TTL} (a), the proposed global task token learning process is mainly composed of two stages: i) Inter- and Intra-Task Attention, which aim to thoroughly model feature-token cross-task relations to obtain robust and representative global task tokens. ii) Feature Supervision Discovery, which excavates pseudo feature supervision with confidence maps provided by global task tokens for unsupervised task features.

Firstly, we use each global task token to query the shared feature to obtain each task feature $\boldsymbol{F_{i}}$ accordingly. Then, to conduct \textit{inter-task attention}, all global task tokens are used to calculate all-task affinities $\boldsymbol{A} \in \mathbb{R}^{T \times T}$ to represent the global task relations. After the $\boldsymbol{A}$ is calculated, it is used to conduct affine combinations of task features and global task tokens respectively. Since for each task feature $\boldsymbol{F_{i}}$, if it is not directly supervised by labels, the feature will be less representative and contain more noise. Thus, conducting affine combinations among all tasks ensures that beneficial discriminative representations from other supervised task features can fertilize the unsupervised ones. Afterward, the updated tokens and features containing cross-task information are fed to the \textit{intra-task attention} module, where they are rearranged and grouped in each task, and self-attention is applied to each group of task token and feature. The global task tokens will further learn more specific and discriminative task representations during this process, and representative task tokens will in turn enhance the feature quality as well. 

Followingly, we discover pseudo feature supervision with the aid of well-learned global task tokens, and this process will be discussed in Sec.~\ref{sec:PSSD}. In addition, for multi-scale backbone features, directly fusing them ignores the various granularity of task representations maintained at different scales. Thus, for multi-scale image backbone, we further propose Multi-scale Global Task Token Learning in order to learn comprehensive multi-scale task relations. The proposed method involves \textit{inter-task attention} separately at each scale, and then the multi-scale features and tokens are fused before \textit{intra-task attention}. In this way, the global task tokens gain richer cross-task relations at different scales and are able to maintain stronger representations. The multi-scale global task token learning, inter- and intra-task attention will be illustrated in detail in the supplementary material.

\subsection{Fine-grained Task Token Learning}
\label{sec:FTTL}
After the global task tokens are learned, we further propose to conduct feature-token interaction at a finer spatial granularity, which takes advantage of various representations in global task tokens and boosts the task label discovery process. 


Firstly, as shown in Fig.~\ref{fig:proj}, we jointly project each updated task token $\boldsymbol{\theta_{i}} \in \mathbb{R}^{C}$ and feature $\boldsymbol{F_{i}} \in \mathbb{R}^{C \times h \times w}$ into the prediction space with finer granularity. For features, this can be easily achieved by applying a linear convolution layer, and we denote the fine-grained task features as $\boldsymbol{G_{i}} \in \mathbb{R}^{C_p \times H \times W}$, where $C_p$ indicates the prediction dimension. For tokens, we denote the projected fine-grained tokens as $\boldsymbol{\varphi_{i}} \in \mathbb{R}^{C_p \times C}$, we hope every $1 \times C$ vector inside it can represent one category distribution over the spatial dimension. The simplest way is to project each $\boldsymbol{\theta_{i}}$ with a Multi-Layer Perceptron (MLP), which can be described as: $\boldsymbol{\varphi_{i}} = \operatorname{MLP_i} (\boldsymbol{\theta_{i}})^{\top}, i = 1, 2, \cdots, T$. However, since there is no direct supervision imposed to distinguish every fine-grain token during this process, the MLPs will tend to degenerate and perform linearly correlated outputs, which prevents the fine-grained tokens from learning discriminative task-specific representations. To alleviate this problem, we propose to use Orthogonal Embeddings (OE) to serve as priors and aid the learning process of fine-grained tokens.

In detail, the vectors in the fine-grained token should be far from each other to represent meaningful and distinguishable task category information, so we use a group of orthogonal basis in $\mathbb{R}^{C_p}$ to serve as the embedding for MLP input, denoting as $\boldsymbol{o} \in \mathbb{R}^{C_p \times C_p}$. These OE are projected into the feature space by linear projections, and then added with the global token before being fed into the MLP. Therefore, with these orthogonal priors, the MLPs can easily keep the distance between the vectors in $\boldsymbol{\varphi_{i}}$ far from each other in the feature space, which makes it able to learn information that distinguishes between task categories, as well as inherit the global representations from $\boldsymbol{\theta_{i}}$.

Subsequently, we exploit the fine-grained task tokens $\boldsymbol{\varphi_{i}}$ for the fine-grained token learning process. Normally, $\boldsymbol{G_{i}}$ is noisy and low-confident on unlabeled data due to the lack of supervision, which leads to inaccurate predictions. We propose to use fine-grained task tokens to encourage $\boldsymbol{G_{i}}$ to produce high-confident logits, and enhance the quality of pseudo labels:
\begin{equation}
\boldsymbol{G_{i}^{\prime}} = \operatorname{Conv_{i}^{3 \times 3}} \left( \boldsymbol{G_{i}} \odot \operatorname{Softplus} \left (\boldsymbol{\varphi_{i}} \times \boldsymbol{G_{i}} \right) \right),
\end{equation}
\noindent where  $\operatorname{Softplus}(x) =  \log (1+\exp (x))$. Since $\boldsymbol{\varphi_{i}}$ inherits global task representations from $\boldsymbol{\theta_{i}}$, which will perform more robustly on unlabeled data, and aid the production of distinguished logits score in the updated feature $\boldsymbol{G_{i}^{\prime}}$, as well as high-confident final task predictions. Similar to the previous stage, we also conduct pseudo label discovery after the fine-grained tokens are learned, which will be discussed in detail in the next section.

\begin{figure}[t]
  \centering
  \includegraphics[width=0.92\linewidth]{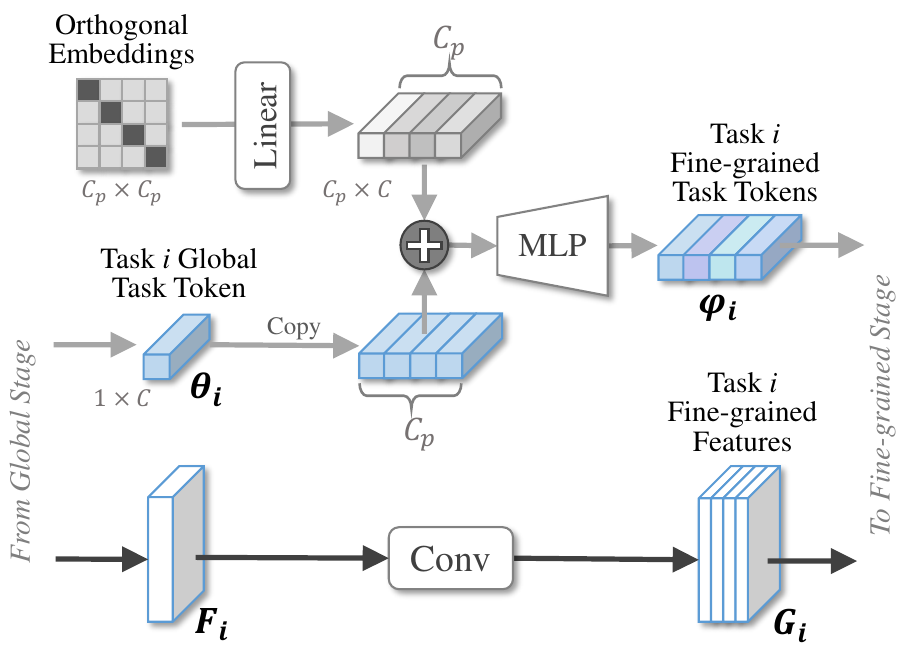}
  \vspace{-4mm}
  \caption{Illustration of Token $\&$ Feature Projection and the Fine-grained Task Token Learning. Different tasks have the same structure, and we take one as an example. We project the fine-grained feature $\boldsymbol{G_{i}}$ from the updated task feature $\boldsymbol{F_{i}}$, and derive fine-grained task tokens $\boldsymbol{\varphi_{i}}$ from the updated global task tokens $\boldsymbol{\theta_{i}}$.}
  \vspace{-2mm}
  \label{fig:proj}
\end{figure}

\subsection{Hierarchical Label Discovery and Multi-task Optimization}
\label{sec:PSSD}
For the multi-task partially annotated setting, the training loss on labeled data can be described as:
\begin{equation}
\label{eq:Ls}
\mathcal{L}_s = \frac{1}{T} \sum_{i=1}^{T}\left(\alpha_i \operatorname{L_i}\left( \boldsymbol{\hat{Y_{i}}}, {\boldsymbol{Y_{i}}} \right) \right),
\end{equation}
\noindent where where $\operatorname{L_i}(\cdot)$ is the loss function for task $i$, and $\alpha_i = 1$ if task $i$ has ground-truth label, otherwise $\alpha_i = 0$. $\boldsymbol{\hat{Y_{i}}}$ and ${\boldsymbol{Y_{i}}}$ are task prediction and ground-truth.

We first train the multi-task model with HiTTs jointly with $\operatorname{L_i}$ only to achieve convergence on labeled data, then we utilize both hierarchies of tokens to discover feature-level and prediction-level pseudo supervision. As we mentioned in Sec.~\ref{sec:GTTL}, before the updated tokens and features are fed into the next stage, we conduct the \textit{feature supervision discovery} to excavate feature-level supervision signals for unlabeled tasks. As shown in Fig.~\ref{fig:TTL} (a), we use the updated global task tokens to query each pixel feature on every task and produce confidence mask $\boldsymbol{M_{i}^{f}} = \operatorname{Sigmoid} (\boldsymbol{\theta_{i}}^{\top} \times \boldsymbol{F_{i}} )$. Since $\boldsymbol{\theta_{i}}$ is globally learned on all task features, in $\boldsymbol{M_{i}^{f}}$, higher scores indicate that the pixel features have a higher response to task $i$, which should be further used to prove task supervision. Thus we use $\boldsymbol{M_{i}^{f}}$ to serve as a soft confidence mask for pixel-wise feature supervision loss: 
\begin{equation}
\label{eq:Lf}
\mathcal{L}_f = \frac{1}{T} \sum_{i=1}^{T} \left(\alpha_i \operatorname{L_{mse}}\left( \boldsymbol{F_{i}}, \boldsymbol{\tilde{F_{i}}} \right) + (1 - \alpha_i) \boldsymbol{M_{i}^{f}} \odot \operatorname{L_{mse}}\left( \boldsymbol{F_{i}}, \boldsymbol{\tilde{F_{i}}} \right) \right),
\end{equation}
\noindent where $\tilde{\boldsymbol{F_{i}}}$ represents the offline saved features which serve as pseudo supervision signals for unsupervised task features. The $\odot$ represents element-wise multiplication, $\operatorname{L_{mse}}$ is the mean squared error loss for feature distance measurement. For the feature loss on labeled task (first item in Eq~\ref{eq:Lf}), we regard all pixel features from $\boldsymbol{\tilde{F_{i}}}$ as valid since they are supervised by ground-truth label, while for unlabeled task (second item in Eq~\ref{eq:Lf}), we use $\boldsymbol{M_{i}^{f}}$ encourage high-confidence pixel features and depress low-confidence ones.

 Afterward, we also conduct \textit{pseudo label discovery} with the aid of fine-grained task tokens $\boldsymbol{\varphi_{i}}$ as mentioned in Sec.~\ref{sec:FTTL}. We directly produce pseudo labels from $\boldsymbol{G_i^{\prime}}$: $\boldsymbol{\tilde{Y_{i}}} =  \operatorname{Argmax} \left (\operatorname{Softmax} \left ( \boldsymbol{G_i^{\prime}} \right)\right)$, along with binary masks to select high confidence pixel pseudo labels: 
$\boldsymbol{M_{i}^{p}} = \operatorname{Max} \left (\operatorname{Softmax} \left ( \boldsymbol{G_i^{\prime}} \right) \right) > \tau_i$, where $\tau_i$ is a threshold used to produce binary masks. The loss for pseudo label supervision can be written as:
\vspace{-2mm}
\begin{equation}
\label{eq:Lp}
\mathcal{L}_p = \frac{1}{T} \sum_{i=1}^{T}\left((1 - \alpha_i) \boldsymbol{M_{i}^{p}} \odot \operatorname{L_i}\left( \boldsymbol{\hat{Y_{i}}}, \boldsymbol{\tilde{Y_{i}}} \right) \right),
\end{equation}
Finally, we sum all of the losses in Eq~\ref{eq:Ls}, ~\ref{eq:Lf} and ~\ref{eq:Lp} to supervise all task features and predictions. The overall losses to optimize the model can be described as: $\mathcal{L} = \mathcal{L}_s + \mathcal{L}_p + \mathcal{L}_f$, each item is combined with the weight $1$ to form the total loss.


\section{Experiment}
\subsection{Experimental Setup}
\noindent \textbf{PASCAL-Context.} PASCAL-Context~\cite{everingham2009pascal} contains 4998 and 5105 images for training and testing respectively, which also have pixel-level annotations for semantic segmentation, human-parts segmentation and semantic edge detection. Additionally, we also consider surface normal estimation and saliency detection distilled by~\cite{maninis2019attentive}. We use Adam optimizer with learning rate $2\times 10^{-5}$, and weight decay $1\times 10^{-6}$, and train for $100$ epochs with batch size $6$. We update the learning rate with polynomial strategy and $\gamma = 0.9$ for the power factor.

\noindent\textbf{NYUD-v2.} NYUD-v2~\cite{silberman2012indoor} contains 795 and 654 RGB-D indoor scene images for training and testing respectively. We use the 13-class semantic annotations which is defined in~\cite{couprie2013indoor}, the truth depth annotations recorded by Microsoft Kinect depth camera, and surface normal annotations which are produced in~\cite{eigen2015predicting}. Following the setting in~\cite{liu2019end,li2022learning}, we use $288 \times 384$ image resolution to speed up training. We use Adam optimizer with a learning rate of $1\times 10^{-4}$, and train all models for 400 epochs with batch size $8$. We update the learning rate every $100$ epoch with $\gamma = 0.5$ as the multiplying factor.

\noindent \textbf{Cityscapes.} Cityscapes~\cite{cordts2016cityscapes} contains 2975 and 500 street-view images for training and testing respectively. We used the projected 7-class semantic annotations from~\cite{liu2019end}, and disparity maps to serve as depth annotations. Following the setting in~\cite{liu2019end,li2022learning}, we use $128 \times 256$ image resolution to speed up training. The optimizer and learning rate scheduler are set as the same as NYUD-v2.

\noindent \textbf{Model Setting.} Following~\cite{li2022learning}, we use SegNet~\cite{badrinarayanan2017segnet} for NYUD-v2 and Cityscapes, ResNet-18~\cite{he2016deep} for PASCAL-Context as the backbone of our single task learning (STL) baselines, and the multi-task baseline (MTL) is built from it, which consists of a shared backbone encoder and several task-specific decoding heads. For the learning of HiTTs, we follow~\cite{bruggemann2021exploring,li2018monocular,zeisl2014discriminatively}, and perform a discrete quantization of the label space of continuous regression tasks such as Depth. and Normal. This discrete quantization does not contribute to multi-task learning performance as analyzed in~\cite{bruggemann2021exploring}, so we ensure a fair comparison with other works. 


\noindent \textbf{Data Preparation.} We follow the setting of~\cite{ye2024diffusionmtl} to process PASCAL-Context, and the setting of~\cite{li2022learning,liu2019end,nishi2024joint} to process NYUDv2 and Cityscapes, and form two partially annotated settings~\cite{li2022learning}: \textbf{(i) one-label}: for each input image, it is only associated with one task annotation; \textbf{(ii) random-labels}: each image has at least one and at most $N-1$ tasks with corresponding task annotations, in the set of $N$ tasks. Additionally, we provide two extra settings \textbf{full-labels} and \textbf{few-shot} in the supplementary material to further validate the effectiveness of our method. 

\noindent \textbf{Training Pipeline.} We first train the multi-task model with HiTTs on all labeled task data. Then the weights of the network and tokens are fixed, and used to produce hierarchical supervision on both feature and prediction spaces. We produce the pseudo label in an offline manner according to~\cite{xie2020self}, which is labeling on clean image without data augmentation, and training on augmented images and pseudo labels to enforce consistent predictions. In~\cite{xie2020self}, this method is only applied to classification tasks while we extend the utilization to general dense prediction tasks. After the pseudo labels are saved, we use them along with the ground-truth labels to jointly train the multi-task model from scratch. Additionally, we also use the pseudo feature supervision produced by this pretrained multi-task model for feature regularization during the optimization process. Both hierarchies of the discovered supervision signals ensure that all task predictions will obtain pixel-wise supervision for multi-task optimization to gain better generalization ability on unlabeled data.

\noindent \textbf{Evaluation Metrics.} We use multiple metrics for each task to evaluate the performance. The metrics include: 
{mIoU} (mean intersection over union), {AbS / AbR} (absolute error / absolute-relative error), maxF (maximal F-measure), {mErr} (mean of angle error), {odsF} (optimal dataset scale F-measure). 
Additionally, to better evaluate the proposed method, we also use $\Delta_{M T L}$ from~\cite{vandenhende2021multi} to evaluate the overall improvement of the multi-task performances of all the tasks.

\begin{figure}[t]
  \centering
  \includegraphics[width=1.0\linewidth]{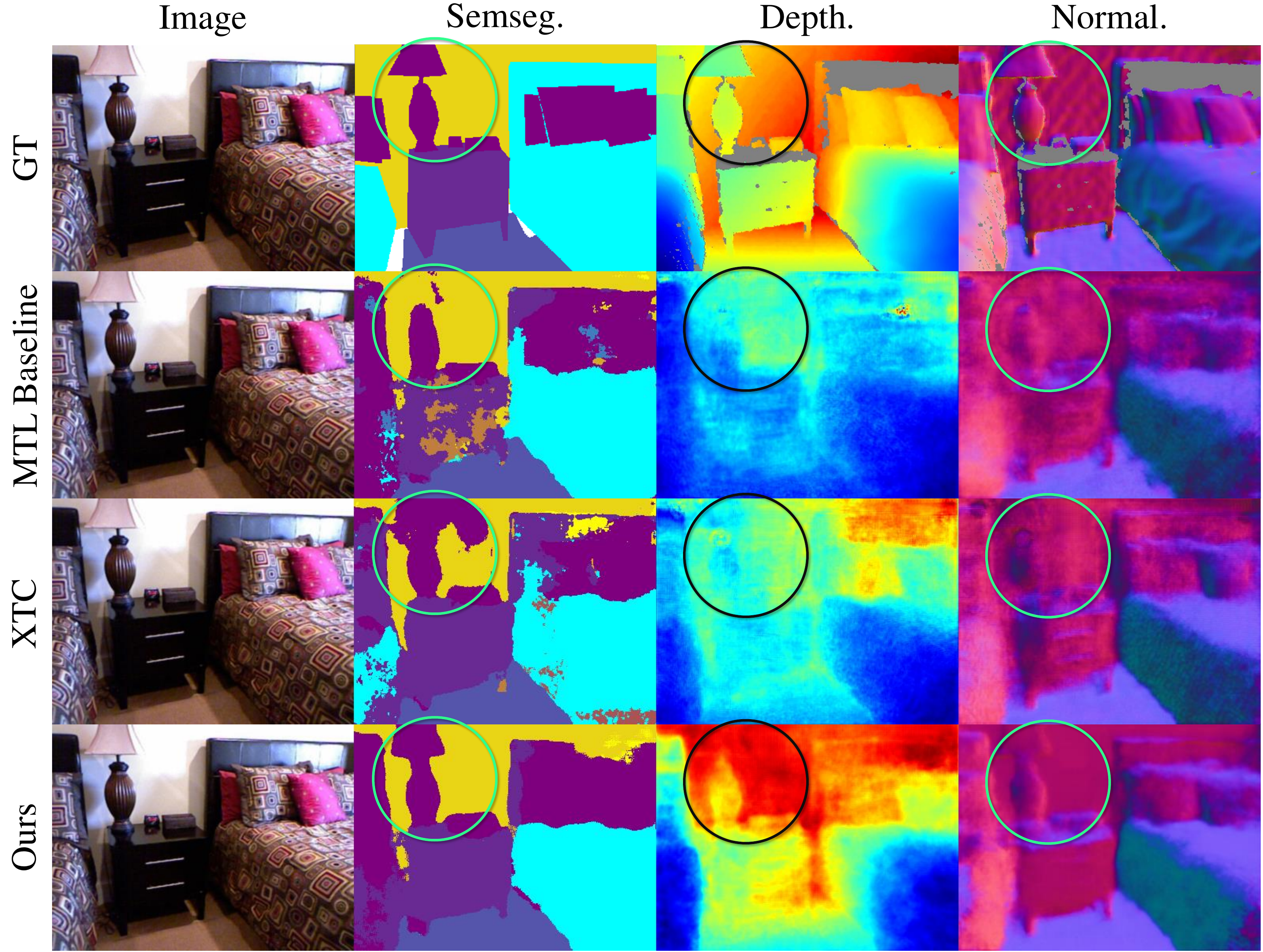}
  \vspace{-7mm}
  \caption{Comparisons with SoTA works on NYUDv2. Ours shows both clear semantic boundaries and accurate geometry estimations, indicating the effectiveness of cross-task feature-token interactions.}
  \vspace{-5mm}
  \label{fig:comSOTA}
\end{figure}

\begin{table*}[t]
 \caption{Quantitative comparison on PASCAL-Context under the \textit{one-label} and \textit{random-labels} setting. (P) and (F) represent the Prediction Diffusion and Feature Diffusion modes for DiffusionMTL~\cite{ye2024diffusionmtl}, and MTDNet is the Multi-Task Denoising Diffusion Networkt~\cite{ye2024diffusionmtl}. The Mapping Network is the extra encoder used for task mappings in~\cite{li2022learning,nishi2024joint}. ``*'' denotes the re-implemented results from~\cite{ye2024diffusionmtl}. Our method performance outperforms previous methods while using significantly fewer model parameters.}
 \vspace{-3mm}
\begin{center}
\setlength{\tabcolsep}{2.5mm}{\scalebox{0.8}{
    \begin{tabular}{c|l|cc|cc|cccccccccc}
        \toprule
        \multirow{2}{*}{\textbf{\# labels}} &  \multicolumn{1}{c|}{\multirow{2}{*}{\textbf{Method}}} & \multicolumn{1}{c}{\multirow{2}{*}{\textbf{MTDNet}}} & \textbf{Mapping}
        &\multirow{2}{*}{\textbf{\#Params}} & \multirow{2}{*}{\textbf{FLOPS}} & \textbf{Semseg}  & \textbf{Parsing}  & \textbf{Saliency} & \textbf{Normal} & \textbf{Boundary}&  \textbf{MTL Perf} \\
        & & & \textbf{Network} & & & mIoU $\mathbf{\uparrow}$  & mIoU $\mathbf{\uparrow}$
      & maxF $\mathbf{\uparrow}$ & mErr $\mathbf{\downarrow}$ & odsF $\mathbf{\uparrow}$ &$\Delta_{MTL}$(\%) $\mathbf{\uparrow}$ \\
        \midrule
        \multirow{10}{*}{\rotatebox{90}{One-Label}}
        & STL & \XSolidBrush & \XSolidBrush & 219M & 817G & 50.34 & 59.05	 & 77.43 & 16.59 & 64.40 & - \\
        & MTL \textit{baseline} & \XSolidBrush & \XSolidBrush & 157M & 608G & 49.71	& 56.00 & 74.50	& 16.85	& 62.80 & -2.85 \\
        & SS~\cite{li2022learning}  & \XSolidBrush & \XSolidBrush & -& - & 45.00 & 54.00 & 61.70 & 16.90 & 62.40 & - \\
        & XTC~\cite{li2022learning}  & \XSolidBrush & \Checkmark & -& - & 49.50 & 55.80 & 61.70 & 17.00 & {65.10} & - \\
        & XTC*~\cite{li2022learning} & \XSolidBrush & \Checkmark & 173M & 608G & 55.08 & 56.72 & 77.06 & 16.93 & 63.70 & 0.37 \\
        & JTR*~\cite{nishi2024joint} & \XSolidBrush & \Checkmark & 173M  &608G &50.29 &54.78 &78.35 &17.97 &63.66 & -3.12 \\
        & DiffusionMTL (P)~\cite{ye2024diffusionmtl} & \Checkmark & \XSolidBrush & 133M & 628G  & {59.43} & 56.79 & 77.57 & 16.20 & 64.00 & 3.23\\
        &DiffusionMTL (F)~\cite{ye2024diffusionmtl} & \Checkmark & \XSolidBrush & 133M & 676G & 57.78 & \textbf{58.98} & {77.82} & {16.11} & 64.50 & {3.65} \\
        &Ours & \XSolidBrush & \XSolidBrush & 62M & 493G & \textbf{61.24} & 57.52 & \textbf{78.35} & \textbf{15.75} & \textbf{67.70} & \textbf{6.09}\\
        \midrule
        \multirow{10}{*}{\rotatebox{90}{Random-Labels}}
        & STL & \XSolidBrush & \XSolidBrush & 219M & 817G & 51.51 & 57.90 & 80.30 & 15.24 & 67.80 & - \\
        & MTL \textit{baseline} & \XSolidBrush & \XSolidBrush & 157M & 608G & 62.23 & 55.88 & 78.67 & 15.47 & 66.70 & 2.44 \\
        & SS~\cite{li2022learning} & \XSolidBrush & \XSolidBrush & -& - & 59.00 & 55.80 & 64.00 & 15.90  &  66.90 & - \\
        & XTC~\cite{li2022learning} & \XSolidBrush  & \Checkmark &-&-& 59.00 & 55.60 & 64.00 & 15.90 & {67.80} &-\\
        & XTC*~\cite{li2022learning} & \XSolidBrush  & \Checkmark & 173M& 608G & 62.44 & 55.81 & 78.56 & 15.45 & 66.80 & 2.52 \\
        & JTR*~\cite{nishi2024joint} & \XSolidBrush  & \Checkmark & 173M  &608G &57.21 & 53.18 & 79.98 & 16.48 & 66.20 & -1.60 \\
        & DiffusionMTL (P)~\cite{ye2024diffusionmtl} & \Checkmark & \XSolidBrush & 133M & 628G & {63.68} & 55.84 & 79.87 & 15.38 & 66.80 & 3.44 \\
        & DiffusionMTL (F)~\cite{ye2024diffusionmtl} & \Checkmark & \XSolidBrush & 133M & 676G & 62.55 & \textbf{56.84} & {80.44} & {14.85} & 67.10 & {4.27} \\
        &Ours & \XSolidBrush & \XSolidBrush & 62M & 493G & \textbf{65.19} & 56.35 & \textbf{81.70} & \textbf{14.80} & \textbf{67.90} & \textbf{5.73} \\
        \bottomrule
    \end{tabular}
}}
\end{center}
\label{tab:pascal}
\end{table*}

\begin{figure*}[!t]
  \centering
  \vspace{-3mm}
  \includegraphics[width=0.92\linewidth]{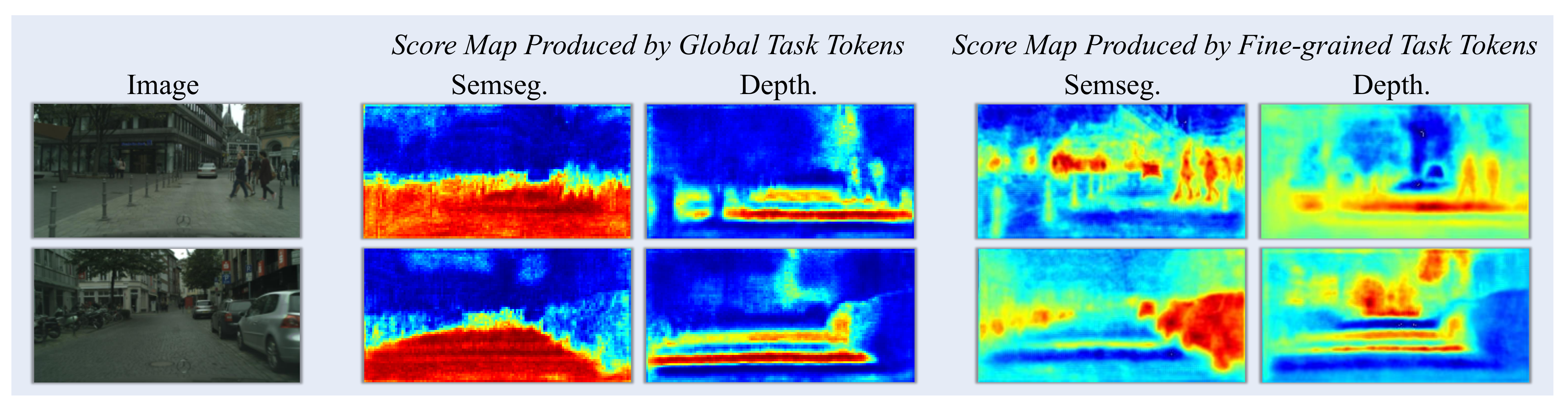}
    \vspace{-4mm}
  \caption{Comparisons of task score maps produced by global task tokens and fine-grained task tokens on Cityscapes.} 
  \vspace{-2mm}
  \label{fig:comp_score_init}
\end{figure*}

\begin{table}[t]
 \caption{Comparison on NYUD-v2 under \textit{one-label} and \textit{random-labels} settings.}
 \vspace{-3mm}
\begin{center}
\setlength{\tabcolsep}{3.4mm}{\scalebox{0.8}{
\begin{tabular}{clcccc}
 \toprule
\multirow{2}{*}{Setting}&\multirow{2}{*}{Model}&{Semseg.} &{Depth.} & {Normal.}&$\Delta_{M T L}$\\
 & &\textit{mIoU}↑ &\textit{AbS}↓ &\textit{mErr}↓ &(\%)↑ \\ \midrule
\multirow{6}*{\rotatebox{90}{One-Label}}&STL & 29.28	&0.7182	&30.1971 & - \\
&MTL \textit{baseline} & 30.92	&0.5982	&31.8509 & 5.61\\
&MTAN~\cite{liu2019end}& 30.92	&0.6196	&30.0278 &   6.63\\
&XTC~\cite{li2022learning} &33.46	&0.5728	&31.1492 &    10.46\\
&JTR~\cite{nishi2024joint} & 31.96 &0.5919 &30.8000 & 8.25 \\
&Ours &\textbf{35.81}	&\textbf{0.5540} &\textbf{28.5131} & \textbf{16.91}\\
\midrule
\multirow{6}*{\rotatebox{90}{Random-Labels}}&STL & 34.49	&0.6272	&27.9681 &-\\
&MTL \textit{baseline} & 35.49	&0.5503	&29.9541 & 2.69\\
&MTAN~\cite{liu2019end}&35.96	&0.6120	&28.6933 & 1.36\\
&XTC~\cite{li2022learning} & 38.11	&0.5387	&29.6549 & 6.19\\
&JTR~\cite{nishi2024joint} & 37.08 &0.5541 &29.4400 & 4.63 \\
&Ours &\textbf{41.78}	&\textbf{0.5177}	&\textbf{27.3488} & \textbf{13.60}\\
 \bottomrule
\end{tabular}
}}
\end{center}
\label{tab:a}
\end{table}

\begin{table}[t]
 \caption{Comparison on Cityscapes under \textit{one-label} setting. ``*'' denotes the re-implemented results to align the settings.}
 \vspace{-3mm}
\begin{center}
\setlength{\tabcolsep}{3.5mm}{\scalebox{0.8}{
\begin{tabular}{clccccc}
 \toprule
\multirow{2}*{Setting}&\multirow{2}*{Model}& {Semseg.} & {Depth.}  &$\Delta_{M T L}$\\
 & &\textit{mIoU}↑ &\textit{AbS}↓ &(\%)↑ \\ \midrule
\multirow{6}{*}{One-Label} & STL & 69.69	&0.0142	& - \\
&MTL \textit{baseline} &69.94	&0.0159	& -5.81\\
&MTAN~\cite{liu2019end} & 71.12	&0.0146	& -0.38\\
&XTC~\cite{li2022learning} & 73.23	&0.0159	& -3.45\\
&JTR~\cite{nishi2024joint} &72.33  & 0.0163 &  -5.50   \\
&DiffusionMTL (F)*~\cite{ye2024diffusionmtl} &73.19	&0.0138	 &  3.92   \\
&Ours & \textbf{73.65}	&\textbf{0.0135}	& \textbf{5.31}\\
 \bottomrule
\end{tabular}
}}
\end{center}
\label{tab:b}
\end{table}


\subsection{State-of-the-art Comparison}
\noindent \textbf{Comparison on Pascal-Context.}
For the comparison on Pascal-Context, we consider both \textit{one-label} and \textit{random-labels} settings. As shown in Table~\ref{tab:pascal}, our method achieves clear improvement over other methods on the majority of tasks under both settings. The greatest enhanced task is Semseg, which has $+3.46$ and $+2.64$ \textit{mIoU} on the two settings compared with~\cite{ye2024diffusionmtl} (F). Overall, our method is $+2.44\%$ and $+1.46\%$ higher in terms of $\Delta_{MTL}$ compared with~\cite{ye2024diffusionmtl} (F). Moreover, our HiTTs are super compact compared with previous methods which either require heavy mapping networks~\cite{li2022learning,nishi2024joint} or extra MTDNet for diffusion decoding~\cite{ye2024diffusionmtl}, while ours achieve significantly better performance with $\sim45\%$ parameter amount and $\sim70\%$ GFlops compared with the best performing~\cite{ye2024diffusionmtl}.

\noindent \textbf{Comparison on NYUD-v2.}
We compare our method with~\cite{li2022learning,liu2019end} on NYUD-v2 under both the one-label and random-labels settings, and the quantitative results are shown in Table~\ref{tab:a}. XTC~\cite{li2022learning} is the first work designed for partially annotated multi-task dense prediction. MTAN~\cite{li2022learning} is an attention-based MTL network designed for the fully supervised setting, and we train it with our setup. The quantitative results show that our method surpasses them by a large margin on all the metrics of the three tasks. More specifically, ours achieves $+6.45\%~\Delta_{MTL}$ and $+7.41\%~\Delta_{MTL}$ compared with~\cite{li2022learning} under the two partial-label settings, respectively. The qualitative comparison with the state-of-the-art method XTC~\cite{li2022learning} as shown in Fig.~\ref{fig:comSOTA} can also confirm the superior performance of our method.


\noindent \textbf{Comparison on Cityscapes.}
We also compare our results with~\cite{li2022learning,liu2019end} on Cityscapes, under the \textit{one-label} setting with both Semseg. and Depth. tasks. As shown in Table~\ref{tab:b}, our method achieves SOTA performance on both tasks, and significantly better performance on Depth ($15.09\%$ higher than~\cite{li2022learning}), resulting in an average gain of $+8.76\%$ in terms of $\Delta_{MTL}$. Additionally, it’s worth mentioning that our work is the only one that achieves balanced performance gain on both tasks compared with STL. 


\begin{table}[t]
 \caption{Investigate the effectiveness of different components on NYUD-v2 testing set under \textit{one-label} setting. $\mathcal{L}_p^{*}$ represents the navie pseudo label loss in prediction space without utilizing HiTTs.}
 \vspace{-3mm}
\begin{center}
\setlength{\tabcolsep}{2.6mm}{\scalebox{0.8}{
\begin{tabular}{lcccc}
 \toprule
\multirow{2}{*}{Method}& {Semseg.} &{Depth.} & {Normal.}&$\Delta_{M T L}$\\
 &\textit{mIoU}↑ &\textit{AbS}↓ &\textit{mErr}↓ &(\%)↑ \\ \midrule
STL & 29.28	&0.7182 &30.1971 & - \\
MTL \textit{baseline} & 30.92	&0.5982	&31.8509 & 5.61\\ \midrule
HiTTs \textit{w/o.} OE &27.38 &0.6049	&30.5904 & 2.66\\
HiTTs \textit{w/o.} Inter-Task Attention &31.26 &0.5966& 30.2911& 7.79  \\
HiTTs \textit{w/o.} Intra-Task Attention &31.44 & 0.5910 & 30.1432 & 8.42\\
HiTTs \textit{w/o.} $\boldsymbol{\theta_{i}}$ &30.03 & 0.5823 &30.0005 &7.38 \\
HiTTs \textit{w/o.} $\boldsymbol{\varphi_{i}}$ &30.53 & 0.5842 &30.0891 &7.76 \\
HiTTs &32.48 &0.5844		&30.0847 & 9.98\\ \midrule
STL \textit{w.} $\mathcal{L}_p^{*}$ & 30.78 &0.6693	&30.2420 & 3.93\\
MTL \textit{w.} $\mathcal{L}_p^{*}$ &  33.59 &0.5882	&29.8174 & 11.36\\
HiTTs \textit{w.} $\mathcal{L}_f$ &33.24	&0.5708		&29.2227 & 12.88\\
HiTTs \textit{w.} $\mathcal{L}_p$ &35.22	&0.5613		&28.8852 & 15.49\\
HiTTs \textit{w.} $\mathcal{L}_p+\mathcal{L}_f$ (\textit{full method}) &\textbf{35.81}	&\textbf{0.5540} &\textbf{28.5131} & \textbf{16.91}\\
\bottomrule
\end{tabular}
}}
\end{center}
\label{tab:d}
\end{table}

\begin{figure}[t]
  \centering
  \includegraphics[width=1.0\linewidth]{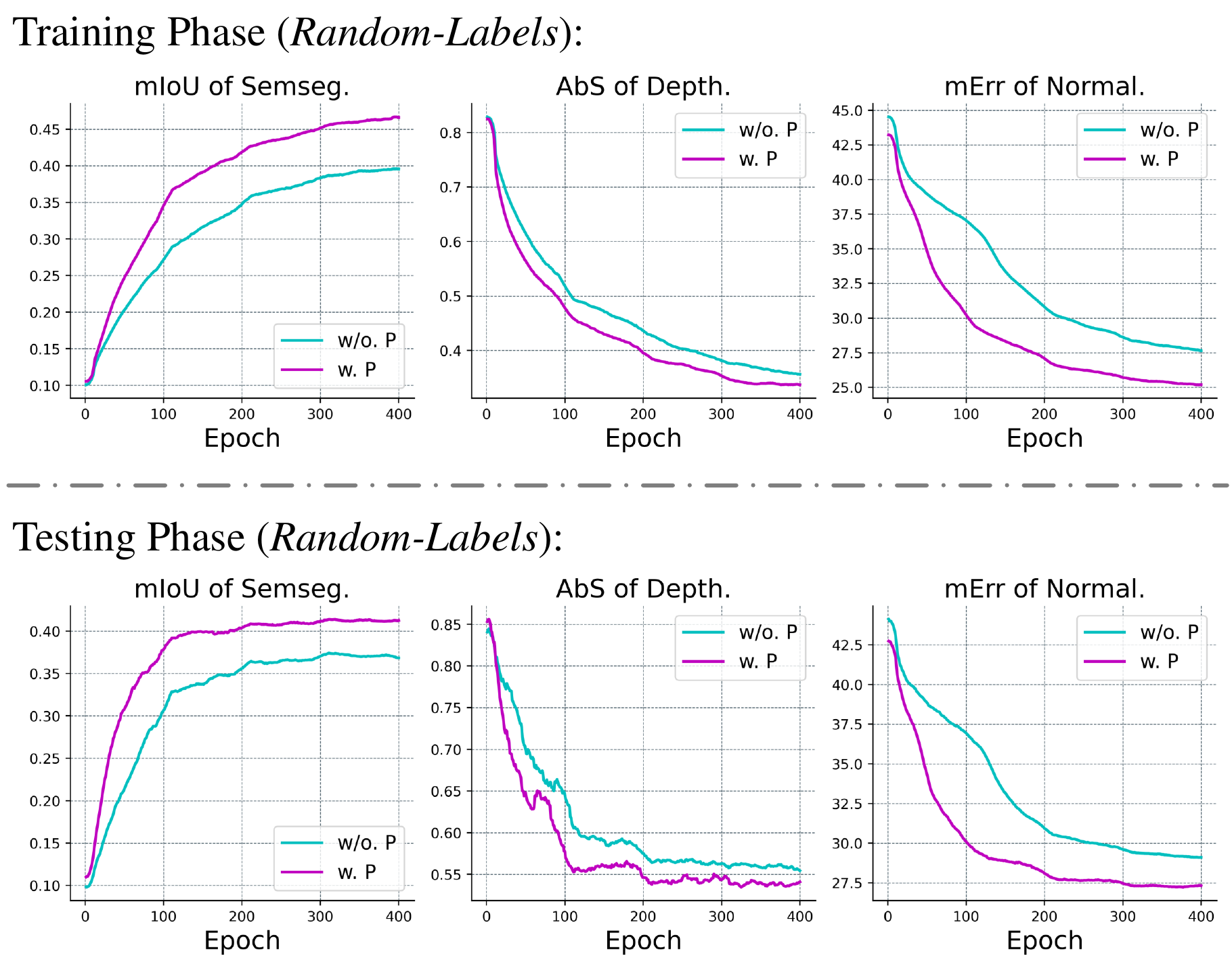}
  \vspace{-7mm}
  \caption{Comparison of the training and testing performance on each task with and without hierarchical Pseudo Supervision (P). The model trained with pseudo supervision converges faster on both train and test splits, and gains better performance.}
  \vspace{-2mm}
  \label{fig:comLoss}
\end{figure}

\begin{table}[t]
 \caption{Investigate the performance on labeled and unlabeled data on NYUD-v2 training set under the \textit{one-label} setting. \textit{GT} and \textit{pseudo} represents the ground-truth and the pseudo supervision, respectively. Our method clearly shows effective learning on unlabeled training data.}
 \vspace{-3mm}
\begin{center}
\setlength{\tabcolsep}{3.2mm}{\scalebox{0.8}{
\begin{tabular}{lccccc}
     \toprule
    \multirow{2}{*}{Method}&\multicolumn{2}{c}{Supervision}&{Semseg.} &{Depth.} &{Normal.}\\
    \cmidrule(lr){2-3}
     &\textit{GT}&\textit{Pseudo} &\textit{mIoU}↑ &\textit{AbS}↓ &\textit{mErr}↓ \\ \midrule
    \multirow{2}{*}{MTL \textit{baseline}} & \Checkmark & \XSolidBrush & 89.00	&0.2041	&25.9280\\
    & \XSolidBrush & \XSolidBrush & 34.31	&0.5823		&31.7697\\ \midrule
    \multirow{2}{*}{HiTTs} & \Checkmark & \XSolidBrush & 86.04   &	0.3016	&21.7911 \\
    & \XSolidBrush & \XSolidBrush & 34.69	&0.5699	&29.8920\\ \midrule
    \multirow{2}{*}{HiTTs \textit{w.} $\mathcal{L}_p+\mathcal{L}_f$} & \Checkmark & \XSolidBrush & 86.63	&0.3173		&20.9220 \\
    & \XSolidBrush & \Checkmark & \textbf{37.25}	&\textbf{0.5563}	&\textbf{28.5169} \\
    \bottomrule
    \end{tabular}
}}
\end{center}
\label{tab:f}
\end{table}

\subsection{Model Analysis}
\noindent \textbf{Components of Hierarchical Task Tokens.}
As shown in Table~\ref{tab:d}, under the {one-label} setting on NYUD-v2, we give an ablation study of HiTTs' key components. Generally, we analyze the role of global task tokens $\boldsymbol{\theta_{i}}$ and fine-grained tokens $\boldsymbol{\varphi_{i}}$, and core designs for token learning process, including the orthogonal embeddings (OE) (\ref{sec:FTTL}), inter- and intra-task attention (\ref{sec:GTTL}). The quantitative results clearly show that both hierarchies of tokens contribute to the multi-task performance and HiTTs boost the model performance by overall $+4.37\%~\Delta_{MTL}$ on all tasks compared with baseline. For the learning process of HiTTs, the inter- and intra-task attention both contribute to the learning process, and the orthogonal embeddings (OE) are essential for generating representative fine-grained task tokens, and without OE, the performance will significantly drop ($-7.32\%~\Delta_{MTL}$).

\par\noindent\textbf{Effect of Hierarchical Feature Supervision and Label Discovery.} To validate that our token-based label discovery is superior to the naive pseudo-labeling process, we compare $\mathcal{L}_p$ imposed on different models, including STL, MTL baselines, and ours. The HiTTs perform i) effective cross-task feature-token interactions; ii) consistent label discovery in both feature and prediction space, which yields better pseudo label quality, and significantly surpasses simply applying $\mathcal{L}_p$ to STL or MTL baselines without HiTTs. We also analyze the contributions from two types of pseudo supervision losses ($\mathcal{L}_p$ and $\mathcal{L}_f$) on different hierarchies. As shown in Table~\ref{tab:d}, both methods boost multi-task performance: $+7.27\%~\Delta_{MTL}$ for $\mathcal{L}_f$ and $+9.88\%~\Delta_{MTL}$ for $\mathcal{L}_p$ compared with MTL baseline. The combination of both methods achieves better performance ($+11.30\%~\Delta_{MTL}$) than applying them separately, which validates the importance of consistently discovering supervision signals in both hierarchies.

\par\noindent\textbf{Visualization Results.} We visualize: i) The visualization of score maps produced by $\boldsymbol{\theta_{i}}$ and $\boldsymbol{\varphi_{i}}$ in Fig.~\ref{fig:comp_score_init}. The visualizations reveal that score maps from global tokens provide a coarse, noisy overview and are biased towards common categories (e.g., focusing only on "road" in Cityscapes). In contrast, maps from fine-grained tokens are detailed, less noisy, and can identify smaller, less frequent objects (e.g., "vehicles" and "pedestrians"). This confirms our hierarchical structure is essential for learning representations at different levels of granularity. ii) The learning curves of metrics on every task in Fig.~\ref{fig:comLoss}, both training and testing performance are boosted consistently on all tasks with the discovered pseudo supervision (P) on both hierarchies.

\par\noindent\textbf{Learning Effect on Unlabeled Data.} We also study the performance of our method on the labeled and unlabeled data separately on NYUD-v2 training set under one-label setting. As shown in Table~\ref{tab:f}, for data without labels, the model with HiTTs generalizes better on them, especially on Depth. and Normal, and adding hierarchical supervision will more significantly boost the performance on unlabeled data.  

\section{Conclusion}
\label{sec:conclusion}
\par In this work, we propose to learn Hierarchical Task Tokens (HiTTs) for both pseudo feature supervision and label discovery under Multi-Task Partially Supervised Learning. The global task tokens are exploited for feature-token cross-task interactions and provide feature-level supervision, while the fine-grained tokens inherit knowledge from global tokens and excavate pixel pseudo labels. Extensive experimental results on partially annotated multi-task dense prediction benchmarks validate the effectiveness of our method.


\bibliographystyle{ACM-Reference-Format}
\balance
\bibliography{sample-base}


\begin{thebibliography}{85}


\ifx \showCODEN    \undefined \def \showCODEN     #1{\unskip}     \fi
\ifx \showISBNx    \undefined \def \showISBNx     #1{\unskip}     \fi
\ifx \showISBNxiii \undefined \def \showISBNxiii  #1{\unskip}     \fi
\ifx \showISSN     \undefined \def \showISSN      #1{\unskip}     \fi
\ifx \showLCCN     \undefined \def \showLCCN      #1{\unskip}     \fi
\ifx \shownote     \undefined \def \shownote      #1{#1}          \fi
\ifx \showarticletitle \undefined \def \showarticletitle #1{#1}   \fi
\ifx \showURL      \undefined \def \showURL       {\relax}        \fi
\providecommand\bibfield[2]{#2}
\providecommand\bibinfo[2]{#2}
\providecommand\natexlab[1]{#1}
\providecommand\showeprint[2][]{arXiv:#2}

\bibitem[Badrinarayanan et~al\mbox{.}(2017)]%
        {badrinarayanan2017segnet}
\bibfield{author}{\bibinfo{person}{Vijay Badrinarayanan}, \bibinfo{person}{Alex Kendall}, {and} \bibinfo{person}{Roberto Cipolla}.} \bibinfo{year}{2017}\natexlab{}.
\newblock \showarticletitle{Segnet: A deep convolutional encoder-decoder architecture for image segmentation}.
\newblock \bibinfo{journal}{\emph{PAMI}} \bibinfo{volume}{39}, \bibinfo{number}{12} (\bibinfo{year}{2017}), \bibinfo{pages}{2481--2495}.
\newblock


\bibitem[Bai et~al\mbox{.}(2023)]%
        {Qwen-VL}
\bibfield{author}{\bibinfo{person}{Jinze Bai}, \bibinfo{person}{Shuai Bai}, \bibinfo{person}{Shusheng Yang}, \bibinfo{person}{Shijie Wang}, \bibinfo{person}{Sinan Tan}, \bibinfo{person}{Peng Wang}, \bibinfo{person}{Junyang Lin}, \bibinfo{person}{Chang Zhou}, {and} \bibinfo{person}{Jingren Zhou}.} \bibinfo{year}{2023}\natexlab{}.
\newblock \showarticletitle{Qwen-VL: A Versatile Vision-Language Model for Understanding, Localization, Text Reading, and Beyond}.
\newblock \bibinfo{journal}{\emph{arXiv preprint arXiv:2308.12966}} (\bibinfo{year}{2023}).
\newblock


\bibitem[Bochkovskiy et~al\mbox{.}(2020)]%
        {bochkovskiy2020yolov4}
\bibfield{author}{\bibinfo{person}{Alexey Bochkovskiy}, \bibinfo{person}{Chien-Yao Wang}, {and} \bibinfo{person}{Hong-Yuan~Mark Liao}.} \bibinfo{year}{2020}\natexlab{}.
\newblock \showarticletitle{Yolov4: Optimal speed and accuracy of object detection}.
\newblock \bibinfo{journal}{\emph{arXiv preprint arXiv:2004.10934}} (\bibinfo{year}{2020}).
\newblock


\bibitem[Br{\"u}ggemann et~al\mbox{.}(2021)]%
        {bruggemann2021exploring}
\bibfield{author}{\bibinfo{person}{David Br{\"u}ggemann}, \bibinfo{person}{Menelaos Kanakis}, \bibinfo{person}{Anton Obukhov}, \bibinfo{person}{Stamatios Georgoulis}, {and} \bibinfo{person}{Luc Van~Gool}.} \bibinfo{year}{2021}\natexlab{}.
\newblock \showarticletitle{Exploring relational context for multi-task dense prediction}. In \bibinfo{booktitle}{\emph{ICCV}}. \bibinfo{pages}{15869--15878}.
\newblock


\bibitem[Cai et~al\mbox{.}(2023)]%
        {cai2023rethinking}
\bibfield{author}{\bibinfo{person}{Yancheng Cai}, \bibinfo{person}{Bo Zhang}, \bibinfo{person}{Baopu Li}, \bibinfo{person}{Tao Chen}, \bibinfo{person}{Hongliang Yan}, \bibinfo{person}{Jingdong Zhang}, {and} \bibinfo{person}{Jiahao Xu}.} \bibinfo{year}{2023}\natexlab{}.
\newblock \showarticletitle{Rethinking cross-domain pedestrian detection: A background-focused distribution alignment framework for instance-free one-stage detectors}.
\newblock \bibinfo{journal}{\emph{IEEE transactions on image processing}}  \bibinfo{volume}{32} (\bibinfo{year}{2023}), \bibinfo{pages}{4935--4950}.
\newblock


\bibitem[Cao et~al\mbox{.}({[n.\,d.]})]%
        {caomsm}
\bibfield{author}{\bibinfo{person}{Mang Cao}, \bibinfo{person}{Sanping Zhou}, \bibinfo{person}{Ye Deng}, \bibinfo{person}{Wenli Huang}, \bibinfo{person}{Le Wang}, {and} \bibinfo{person}{Jinjun Wang}.} \bibinfo{year}{[n.\,d.]}\natexlab{}.
\newblock \showarticletitle{MSM: Multi-Scale Mamba in Multi-Task Dense Prediction}.
\newblock  (\bibinfo{year}{[n.\,d.]}).
\newblock


\bibitem[Chavhan et~al\mbox{.}(2025)]%
        {chavhan2025upcycling}
\bibfield{author}{\bibinfo{person}{Ruchika Chavhan}, \bibinfo{person}{Abhinav Mehrotra}, \bibinfo{person}{Malcolm Chadwick}, \bibinfo{person}{Alberto~Gil Ramos}, \bibinfo{person}{Luca Morreale}, \bibinfo{person}{Mehdi Noroozi}, {and} \bibinfo{person}{Sourav Bhattacharya}.} \bibinfo{year}{2025}\natexlab{}.
\newblock \showarticletitle{Upcycling Text-to-Image Diffusion Models for Multi-Task Capabilities}.
\newblock \bibinfo{journal}{\emph{arXiv preprint arXiv:2503.11905}} (\bibinfo{year}{2025}).
\newblock


\bibitem[Chen et~al\mbox{.}(2023)]%
        {chen2023instance}
\bibfield{author}{\bibinfo{person}{Linwei Chen}, \bibinfo{person}{Ying Fu}, \bibinfo{person}{Kaixuan Wei}, \bibinfo{person}{Dezhi Zheng}, {and} \bibinfo{person}{Felix Heide}.} \bibinfo{year}{2023}\natexlab{}.
\newblock \showarticletitle{Instance segmentation in the dark}.
\newblock \bibinfo{journal}{\emph{International Journal of Computer Vision}} \bibinfo{volume}{131}, \bibinfo{number}{8} (\bibinfo{year}{2023}), \bibinfo{pages}{2198--2218}.
\newblock


\bibitem[Chen et~al\mbox{.}(2018b)]%
        {chen2018encoder}
\bibfield{author}{\bibinfo{person}{Liang-Chieh Chen}, \bibinfo{person}{Yukun Zhu}, \bibinfo{person}{George Papandreou}, \bibinfo{person}{Florian Schroff}, {and} \bibinfo{person}{Hartwig Adam}.} \bibinfo{year}{2018}\natexlab{b}.
\newblock \showarticletitle{Encoder-decoder with atrous separable convolution for semantic image segmentation}. In \bibinfo{booktitle}{\emph{Proceedings of the European conference on computer vision (ECCV)}}. \bibinfo{pages}{801--818}.
\newblock


\bibitem[Chen et~al\mbox{.}(2018a)]%
        {chen2018gradnorm}
\bibfield{author}{\bibinfo{person}{Zhao Chen}, \bibinfo{person}{Vijay Badrinarayanan}, \bibinfo{person}{Chen-Yu Lee}, {and} \bibinfo{person}{Andrew Rabinovich}.} \bibinfo{year}{2018}\natexlab{a}.
\newblock \showarticletitle{Gradnorm: Gradient normalization for adaptive loss balancing in deep multitask networks}. In \bibinfo{booktitle}{\emph{International conference on machine learning}}. PMLR, \bibinfo{pages}{794--803}.
\newblock


\bibitem[Cheng et~al\mbox{.}(2022)]%
        {cheng2022masked}
\bibfield{author}{\bibinfo{person}{Bowen Cheng}, \bibinfo{person}{Ishan Misra}, \bibinfo{person}{Alexander~G Schwing}, \bibinfo{person}{Alexander Kirillov}, {and} \bibinfo{person}{Rohit Girdhar}.} \bibinfo{year}{2022}\natexlab{}.
\newblock \showarticletitle{Masked-attention mask transformer for universal image segmentation}. In \bibinfo{booktitle}{\emph{Proceedings of the IEEE/CVF conference on computer vision and pattern recognition}}. \bibinfo{pages}{1290--1299}.
\newblock


\bibitem[Cordts et~al\mbox{.}(2016)]%
        {cordts2016cityscapes}
\bibfield{author}{\bibinfo{person}{Marius Cordts}, \bibinfo{person}{Mohamed Omran}, \bibinfo{person}{Sebastian Ramos}, \bibinfo{person}{Timo Rehfeld}, \bibinfo{person}{Markus Enzweiler}, \bibinfo{person}{Rodrigo Benenson}, \bibinfo{person}{Uwe Franke}, \bibinfo{person}{Stefan Roth}, {and} \bibinfo{person}{Bernt Schiele}.} \bibinfo{year}{2016}\natexlab{}.
\newblock \showarticletitle{The cityscapes dataset for semantic urban scene understanding}. In \bibinfo{booktitle}{\emph{CVPR}}. \bibinfo{pages}{3213--3223}.
\newblock
\href{https://doi.org/10.1109/CVPR.2016.350}{doi:\nolinkurl{10.1109/CVPR.2016.350}}


\bibitem[Couprie et~al\mbox{.}(2013)]%
        {couprie2013indoor}
\bibfield{author}{\bibinfo{person}{Camille Couprie}, \bibinfo{person}{Cl{\'e}ment Farabet}, \bibinfo{person}{Laurent Najman}, {and} \bibinfo{person}{Yann LeCun}.} \bibinfo{year}{2013}\natexlab{}.
\newblock \showarticletitle{Indoor semantic segmentation using depth information}.
\newblock \bibinfo{journal}{\emph{arXiv preprint arXiv:1301.3572}} (\bibinfo{year}{2013}).
\newblock


\bibitem[D{\'e}sid{\'e}ri(2012)]%
        {desideri2012multiple}
\bibfield{author}{\bibinfo{person}{Jean-Antoine D{\'e}sid{\'e}ri}.} \bibinfo{year}{2012}\natexlab{}.
\newblock \showarticletitle{Multiple-gradient descent algorithm (MGDA) for multiobjective optimization}.
\newblock \bibinfo{journal}{\emph{Comptes Rendus Mathematique}} \bibinfo{volume}{350}, \bibinfo{number}{5-6} (\bibinfo{year}{2012}), \bibinfo{pages}{313--318}.
\newblock


\bibitem[Dosovitskiy et~al\mbox{.}(2020)]%
        {dosovitskiy2020image}
\bibfield{author}{\bibinfo{person}{Alexey Dosovitskiy}, \bibinfo{person}{Lucas Beyer}, \bibinfo{person}{Alexander Kolesnikov}, \bibinfo{person}{Dirk Weissenborn}, \bibinfo{person}{Xiaohua Zhai}, \bibinfo{person}{Thomas Unterthiner}, \bibinfo{person}{Mostafa Dehghani}, \bibinfo{person}{Matthias Minderer}, \bibinfo{person}{Georg Heigold}, \bibinfo{person}{Sylvain Gelly}, {et~al\mbox{.}}} \bibinfo{year}{2020}\natexlab{}.
\newblock \showarticletitle{An image is worth 16x16 words: Transformers for image recognition at scale}.
\newblock \bibinfo{journal}{\emph{arXiv preprint arXiv:2010.11929}} (\bibinfo{year}{2020}).
\newblock


\bibitem[Eigen and Fergus(2015)]%
        {eigen2015predicting}
\bibfield{author}{\bibinfo{person}{David Eigen} {and} \bibinfo{person}{Rob Fergus}.} \bibinfo{year}{2015}\natexlab{}.
\newblock \showarticletitle{Predicting depth, surface normals and semantic labels with a common multi-scale convolutional architecture}. In \bibinfo{booktitle}{\emph{ICCV}}. \bibinfo{pages}{2650--2658}.
\newblock


\bibitem[Everingham et~al\mbox{.}(2009)]%
        {everingham2009pascal}
\bibfield{author}{\bibinfo{person}{Mark Everingham}, \bibinfo{person}{Luc Van~Gool}, \bibinfo{person}{Christopher~KI Williams}, \bibinfo{person}{John Winn}, {and} \bibinfo{person}{Andrew Zisserman}.} \bibinfo{year}{2009}\natexlab{}.
\newblock \showarticletitle{The pascal visual object classes (voc) challenge}.
\newblock \bibinfo{journal}{\emph{IJCV}}  \bibinfo{volume}{88} (\bibinfo{year}{2009}), \bibinfo{pages}{303--308}.
\newblock
\href{https://doi.org/10.1007/s11263-009-0275-4}{doi:\nolinkurl{10.1007/s11263-009-0275-4}}


\bibitem[Gao et~al\mbox{.}(2019)]%
        {gao2019nddr}
\bibfield{author}{\bibinfo{person}{Yuan Gao}, \bibinfo{person}{Jiayi Ma}, \bibinfo{person}{Mingbo Zhao}, \bibinfo{person}{Wei Liu}, {and} \bibinfo{person}{Alan~L Yuille}.} \bibinfo{year}{2019}\natexlab{}.
\newblock \showarticletitle{Nddr-cnn: Layerwise feature fusing in multi-task cnns by neural discriminative dimensionality reduction}. In \bibinfo{booktitle}{\emph{CVPR}}. \bibinfo{pages}{3205--3214}.
\newblock


\bibitem[Gu et~al\mbox{.}(2024)]%
        {gu2024diffusioninst}
\bibfield{author}{\bibinfo{person}{Zhangxuan Gu}, \bibinfo{person}{Haoxing Chen}, {and} \bibinfo{person}{Zhuoer Xu}.} \bibinfo{year}{2024}\natexlab{}.
\newblock \showarticletitle{Diffusioninst: Diffusion model for instance segmentation}. In \bibinfo{booktitle}{\emph{ICASSP 2024-2024 IEEE International Conference on Acoustics, Speech and Signal Processing (ICASSP)}}. IEEE, \bibinfo{pages}{2730--2734}.
\newblock


\bibitem[He et~al\mbox{.}(2016)]%
        {he2016deep}
\bibfield{author}{\bibinfo{person}{Kaiming He}, \bibinfo{person}{Xiangyu Zhang}, \bibinfo{person}{Shaoqing Ren}, {and} \bibinfo{person}{Jian Sun}.} \bibinfo{year}{2016}\natexlab{}.
\newblock \showarticletitle{Deep residual learning for image recognition}. In \bibinfo{booktitle}{\emph{CVPR}}. \bibinfo{pages}{770--778}.
\newblock


\bibitem[Hoyer et~al\mbox{.}(2024)]%
        {hoyer2024semivl}
\bibfield{author}{\bibinfo{person}{Lukas Hoyer}, \bibinfo{person}{David~Joseph Tan}, \bibinfo{person}{Muhammad~Ferjad Naeem}, \bibinfo{person}{Luc Van~Gool}, {and} \bibinfo{person}{Federico Tombari}.} \bibinfo{year}{2024}\natexlab{}.
\newblock \showarticletitle{SemiVL: semi-supervised semantic segmentation with vision-language guidance}. In \bibinfo{booktitle}{\emph{European Conference on Computer Vision}}. Springer, \bibinfo{pages}{257--275}.
\newblock


\bibitem[Imran et~al\mbox{.}(2020)]%
        {imran2020partly}
\bibfield{author}{\bibinfo{person}{Abdullah-Al-Zubaer Imran}, \bibinfo{person}{Chao Huang}, \bibinfo{person}{Hui Tang}, \bibinfo{person}{Wei Fan}, \bibinfo{person}{Yuan Xiao}, \bibinfo{person}{Dingjun Hao}, \bibinfo{person}{Zhen Qian}, {and} \bibinfo{person}{Demetri Terzopoulos}.} \bibinfo{year}{2020}\natexlab{}.
\newblock \showarticletitle{Partly Supervised Multitask Learning}.
\newblock \bibinfo{journal}{\emph{arXiv preprint arXiv:2005.02523}} (\bibinfo{year}{2020}).
\newblock


\bibitem[Iscen et~al\mbox{.}(2019)]%
        {iscen2019label}
\bibfield{author}{\bibinfo{person}{Ahmet Iscen}, \bibinfo{person}{Giorgos Tolias}, \bibinfo{person}{Yannis Avrithis}, {and} \bibinfo{person}{Ondrej Chum}.} \bibinfo{year}{2019}\natexlab{}.
\newblock \showarticletitle{Label propagation for deep semi-supervised learning}. In \bibinfo{booktitle}{\emph{CVPR}}. \bibinfo{pages}{5070--5079}.
\newblock


\bibitem[Jaritz et~al\mbox{.}(2019)]%
        {jaritz2019multi}
\bibfield{author}{\bibinfo{person}{Maximilian Jaritz}, \bibinfo{person}{Jiayuan Gu}, {and} \bibinfo{person}{Hao Su}.} \bibinfo{year}{2019}\natexlab{}.
\newblock \showarticletitle{Multi-view pointnet for 3d scene understanding}. In \bibinfo{booktitle}{\emph{Proceedings of the IEEE/CVF international conference on computer vision workshops}}. \bibinfo{pages}{0--0}.
\newblock


\bibitem[Kendall et~al\mbox{.}(2018)]%
        {kendall2018multi}
\bibfield{author}{\bibinfo{person}{Alex Kendall}, \bibinfo{person}{Yarin Gal}, {and} \bibinfo{person}{Roberto Cipolla}.} \bibinfo{year}{2018}\natexlab{}.
\newblock \showarticletitle{Multi-task learning using uncertainty to weigh losses for scene geometry and semantics}. In \bibinfo{booktitle}{\emph{Proceedings of the IEEE conference on computer vision and pattern recognition}}. \bibinfo{pages}{7482--7491}.
\newblock


\bibitem[Lee et~al\mbox{.}(2013)]%
        {lee2013pseudo}
\bibfield{author}{\bibinfo{person}{Dong-Hyun Lee} {et~al\mbox{.}}} \bibinfo{year}{2013}\natexlab{}.
\newblock \showarticletitle{Pseudo-label: The simple and efficient semi-supervised learning method for deep neural networks}. In \bibinfo{booktitle}{\emph{ICML}}, Vol.~\bibinfo{volume}{3}. \bibinfo{pages}{896}.
\newblock


\bibitem[Li et~al\mbox{.}(2018)]%
        {li2018monocular}
\bibfield{author}{\bibinfo{person}{Bo Li}, \bibinfo{person}{Yuchao Dai}, {and} \bibinfo{person}{Mingyi He}.} \bibinfo{year}{2018}\natexlab{}.
\newblock \showarticletitle{Monocular depth estimation with hierarchical fusion of dilated cnns and soft-weighted-sum inference}.
\newblock \bibinfo{journal}{\emph{Pattern Recognition}}  \bibinfo{volume}{83} (\bibinfo{year}{2018}), \bibinfo{pages}{328--339}.
\newblock


\bibitem[Li et~al\mbox{.}(2022)]%
        {li2022learning}
\bibfield{author}{\bibinfo{person}{Wei-Hong Li}, \bibinfo{person}{Xialei Liu}, {and} \bibinfo{person}{Hakan Bilen}.} \bibinfo{year}{2022}\natexlab{}.
\newblock \showarticletitle{Learning multiple dense prediction tasks from partially annotated data}. In \bibinfo{booktitle}{\emph{CVPR}}. \bibinfo{pages}{18879--18889}.
\newblock


\bibitem[Li et~al\mbox{.}(2019)]%
        {li2019bidirectional}
\bibfield{author}{\bibinfo{person}{Yunsheng Li}, \bibinfo{person}{Lu Yuan}, {and} \bibinfo{person}{Nuno Vasconcelos}.} \bibinfo{year}{2019}\natexlab{}.
\newblock \showarticletitle{Bidirectional learning for domain adaptation of semantic segmentation}. In \bibinfo{booktitle}{\emph{CVPR}}. \bibinfo{pages}{6936--6945}.
\newblock


\bibitem[Lin et~al\mbox{.}(2023)]%
        {lin2023video}
\bibfield{author}{\bibinfo{person}{Bin Lin}, \bibinfo{person}{Yang Ye}, \bibinfo{person}{Bin Zhu}, \bibinfo{person}{Jiaxi Cui}, \bibinfo{person}{Munan Ning}, \bibinfo{person}{Peng Jin}, {and} \bibinfo{person}{Li Yuan}.} \bibinfo{year}{2023}\natexlab{}.
\newblock \showarticletitle{Video-llava: Learning united visual representation by alignment before projection}.
\newblock \bibinfo{journal}{\emph{arXiv preprint arXiv:2311.10122}} (\bibinfo{year}{2023}).
\newblock


\bibitem[Liu et~al\mbox{.}(2023)]%
        {liu2023visual}
\bibfield{author}{\bibinfo{person}{Haotian Liu}, \bibinfo{person}{Chunyuan Li}, \bibinfo{person}{Qingyang Wu}, {and} \bibinfo{person}{Yong~Jae Lee}.} \bibinfo{year}{2023}\natexlab{}.
\newblock \showarticletitle{Visual instruction tuning}.
\newblock \bibinfo{journal}{\emph{Advances in neural information processing systems}}  \bibinfo{volume}{36} (\bibinfo{year}{2023}), \bibinfo{pages}{34892--34916}.
\newblock


\bibitem[Liu et~al\mbox{.}(2007)]%
        {liu2007semi}
\bibfield{author}{\bibinfo{person}{Qiuhua Liu}, \bibinfo{person}{Xuejun Liao}, {and} \bibinfo{person}{Lawrence Carin}.} \bibinfo{year}{2007}\natexlab{}.
\newblock \showarticletitle{Semi-supervised multitask learning}.
\newblock \bibinfo{journal}{\emph{NIPS}}  \bibinfo{volume}{20} (\bibinfo{year}{2007}).
\newblock


\bibitem[Liu et~al\mbox{.}(2019)]%
        {liu2019end}
\bibfield{author}{\bibinfo{person}{Shikun Liu}, \bibinfo{person}{Edward Johns}, {and} \bibinfo{person}{Andrew~J Davison}.} \bibinfo{year}{2019}\natexlab{}.
\newblock \showarticletitle{End-to-end multi-task learning with attention}. In \bibinfo{booktitle}{\emph{CVPR}}. \bibinfo{pages}{1871--1880}.
\newblock


\bibitem[Lu et~al\mbox{.}(2024)]%
        {lu2024swiss}
\bibfield{author}{\bibinfo{person}{Yuxiang Lu}, \bibinfo{person}{Shengcao Cao}, {and} \bibinfo{person}{Yu-Xiong Wang}.} \bibinfo{year}{2024}\natexlab{}.
\newblock \showarticletitle{Swiss army knife: Synergizing biases in knowledge from vision foundation models for multi-task learning}.
\newblock \bibinfo{journal}{\emph{arXiv preprint arXiv:2410.14633}} (\bibinfo{year}{2024}).
\newblock


\bibitem[Lu et~al\mbox{.}(2021)]%
        {lu2021taskology}
\bibfield{author}{\bibinfo{person}{Yao Lu}, \bibinfo{person}{Soren Pirk}, \bibinfo{person}{Jan Dlabal}, \bibinfo{person}{Anthony Brohan}, \bibinfo{person}{Ankita Pasad}, \bibinfo{person}{Zhao Chen}, \bibinfo{person}{Vincent Casser}, \bibinfo{person}{Anelia Angelova}, {and} \bibinfo{person}{Ariel Gordon}.} \bibinfo{year}{2021}\natexlab{}.
\newblock \showarticletitle{Taskology: Utilizing task relations at scale}. In \bibinfo{booktitle}{\emph{CVPR}}. \bibinfo{pages}{8700--8709}.
\newblock


\bibitem[Luo et~al\mbox{.}(2021)]%
        {luo2021semi}
\bibfield{author}{\bibinfo{person}{Xiangde Luo}, \bibinfo{person}{Jieneng Chen}, \bibinfo{person}{Tao Song}, {and} \bibinfo{person}{Guotai Wang}.} \bibinfo{year}{2021}\natexlab{}.
\newblock \showarticletitle{Semi-supervised medical image segmentation through dual-task consistency}. In \bibinfo{booktitle}{\emph{AAAI}}, Vol.~\bibinfo{volume}{35}. \bibinfo{pages}{8801--8809}.
\newblock


\bibitem[Mai et~al\mbox{.}(2024)]%
        {mai2024rankmatch}
\bibfield{author}{\bibinfo{person}{Huayu Mai}, \bibinfo{person}{Rui Sun}, \bibinfo{person}{Tianzhu Zhang}, {and} \bibinfo{person}{Feng Wu}.} \bibinfo{year}{2024}\natexlab{}.
\newblock \showarticletitle{Rankmatch: Exploring the better consistency regularization for semi-supervised semantic segmentation}. In \bibinfo{booktitle}{\emph{Proceedings of the IEEE/CVF conference on computer vision and pattern recognition}}. \bibinfo{pages}{3391--3401}.
\newblock


\bibitem[Maninis et~al\mbox{.}(2019)]%
        {maninis2019attentive}
\bibfield{author}{\bibinfo{person}{Kevis-Kokitsi Maninis}, \bibinfo{person}{Ilija Radosavovic}, {and} \bibinfo{person}{Iasonas Kokkinos}.} \bibinfo{year}{2019}\natexlab{}.
\newblock \showarticletitle{Attentive single-tasking of multiple tasks}. In \bibinfo{booktitle}{\emph{Proceedings of the IEEE/CVF conference on computer vision and pattern recognition}}. \bibinfo{pages}{1851--1860}.
\newblock


\bibitem[Martin et~al\mbox{.}(2004)]%
        {martin2004learning}
\bibfield{author}{\bibinfo{person}{David~R Martin}, \bibinfo{person}{Charless~C Fowlkes}, {and} \bibinfo{person}{Jitendra Malik}.} \bibinfo{year}{2004}\natexlab{}.
\newblock \showarticletitle{Learning to detect natural image boundaries using local brightness, color, and texture cues}.
\newblock \bibinfo{journal}{\emph{PAMI}} \bibinfo{volume}{26}, \bibinfo{number}{5} (\bibinfo{year}{2004}), \bibinfo{pages}{530--549}.
\newblock


\bibitem[Misra et~al\mbox{.}(2016)]%
        {misra2016cross}
\bibfield{author}{\bibinfo{person}{Ishan Misra}, \bibinfo{person}{Abhinav Shrivastava}, \bibinfo{person}{Abhinav Gupta}, {and} \bibinfo{person}{Martial Hebert}.} \bibinfo{year}{2016}\natexlab{}.
\newblock \showarticletitle{Cross-stitch networks for multi-task learning}. In \bibinfo{booktitle}{\emph{CVPR}}. \bibinfo{pages}{3994--4003}.
\newblock


\bibitem[Nishi et~al\mbox{.}(2024)]%
        {nishi2024joint}
\bibfield{author}{\bibinfo{person}{Kento Nishi}, \bibinfo{person}{Junsik Kim}, \bibinfo{person}{Wanhua Li}, {and} \bibinfo{person}{Hanspeter Pfister}.} \bibinfo{year}{2024}\natexlab{}.
\newblock \showarticletitle{Joint-Task Regularization for Partially Labeled Multi-Task Learning}. In \bibinfo{booktitle}{\emph{Proceedings of the IEEE/CVF Conference on Computer Vision and Pattern Recognition}}. \bibinfo{pages}{16152--16162}.
\newblock


\bibitem[Peng et~al\mbox{.}(2023)]%
        {peng2023openscene}
\bibfield{author}{\bibinfo{person}{Songyou Peng}, \bibinfo{person}{Kyle Genova}, \bibinfo{person}{Chiyu Jiang}, \bibinfo{person}{Andrea Tagliasacchi}, \bibinfo{person}{Marc Pollefeys}, \bibinfo{person}{Thomas Funkhouser}, {et~al\mbox{.}}} \bibinfo{year}{2023}\natexlab{}.
\newblock \showarticletitle{Openscene: 3d scene understanding with open vocabularies}. In \bibinfo{booktitle}{\emph{Proceedings of the IEEE/CVF conference on computer vision and pattern recognition}}. \bibinfo{pages}{815--824}.
\newblock


\bibitem[Pieropan et~al\mbox{.}(2022)]%
        {pieropan2022dense}
\bibfield{author}{\bibinfo{person}{Alessandro Pieropan}, \bibinfo{person}{Hossein Azizpour}, \bibinfo{person}{Atsuto Maki}, {et~al\mbox{.}}} \bibinfo{year}{2022}\natexlab{}.
\newblock \showarticletitle{Dense FixMatch: a simple semi-supervised learning method for pixel-wise prediction tasks}.
\newblock \bibinfo{journal}{\emph{arXiv preprint arXiv:2210.09919}} (\bibinfo{year}{2022}).
\newblock


\bibitem[Ran et~al\mbox{.}(2024)]%
        {ran2024pseudo}
\bibfield{author}{\bibinfo{person}{Lingyan Ran}, \bibinfo{person}{Yali Li}, \bibinfo{person}{Guoqiang Liang}, {and} \bibinfo{person}{Yanning Zhang}.} \bibinfo{year}{2024}\natexlab{}.
\newblock \showarticletitle{Pseudo labeling methods for semi-supervised semantic segmentation: A review and future perspectives}.
\newblock \bibinfo{journal}{\emph{IEEE Transactions on Circuits and Systems for Video Technology}} (\bibinfo{year}{2024}).
\newblock


\bibitem[Ren et~al\mbox{.}(2024)]%
        {ren2024dino}
\bibfield{author}{\bibinfo{person}{Tianhe Ren}, \bibinfo{person}{Yihao Chen}, \bibinfo{person}{Qing Jiang}, \bibinfo{person}{Zhaoyang Zeng}, \bibinfo{person}{Yuda Xiong}, \bibinfo{person}{Wenlong Liu}, \bibinfo{person}{Zhengyu Ma}, \bibinfo{person}{Junyi Shen}, \bibinfo{person}{Yuan Gao}, \bibinfo{person}{Xiaoke Jiang}, {et~al\mbox{.}}} \bibinfo{year}{2024}\natexlab{}.
\newblock \showarticletitle{Dino-x: A unified vision model for open-world object detection and understanding}.
\newblock \bibinfo{journal}{\emph{arXiv preprint arXiv:2411.14347}} (\bibinfo{year}{2024}).
\newblock


\bibitem[Schult et~al\mbox{.}(2022)]%
        {schult2022mask3d}
\bibfield{author}{\bibinfo{person}{Jonas Schult}, \bibinfo{person}{Francis Engelmann}, \bibinfo{person}{Alexander Hermans}, \bibinfo{person}{Or Litany}, \bibinfo{person}{Siyu Tang}, {and} \bibinfo{person}{Bastian Leibe}.} \bibinfo{year}{2022}\natexlab{}.
\newblock \showarticletitle{Mask3d: Mask transformer for 3d semantic instance segmentation}.
\newblock \bibinfo{journal}{\emph{arXiv preprint arXiv:2210.03105}} (\bibinfo{year}{2022}).
\newblock


\bibitem[Shi et~al\mbox{.}(2018)]%
        {shi2018transductive}
\bibfield{author}{\bibinfo{person}{Weiwei Shi}, \bibinfo{person}{Yihong Gong}, \bibinfo{person}{Chris Ding}, \bibinfo{person}{Zhiheng~MaXiaoyu Tao}, {and} \bibinfo{person}{Nanning Zheng}.} \bibinfo{year}{2018}\natexlab{}.
\newblock \showarticletitle{Transductive semi-supervised deep learning using min-max features}. In \bibinfo{booktitle}{\emph{ECCV}}. \bibinfo{pages}{299--315}.
\newblock


\bibitem[Silberman et~al\mbox{.}(2012)]%
        {silberman2012indoor}
\bibfield{author}{\bibinfo{person}{Nathan Silberman}, \bibinfo{person}{Derek Hoiem}, \bibinfo{person}{Pushmeet Kohli}, {and} \bibinfo{person}{Rob Fergus}.} \bibinfo{year}{2012}\natexlab{}.
\newblock \showarticletitle{Indoor segmentation and support inference from rgbd images}. In \bibinfo{booktitle}{\emph{ECCV}}. Springer, \bibinfo{pages}{746--760}.
\newblock
\showISBNx{978-3-642-33715-4}
\href{https://doi.org/10.1007/978-3-642-33715-4_54}{doi:\nolinkurl{10.1007/978-3-642-33715-4_54}}


\bibitem[Sohn et~al\mbox{.}(2020)]%
        {sohn2020fixmatch}
\bibfield{author}{\bibinfo{person}{Kihyuk Sohn}, \bibinfo{person}{David Berthelot}, \bibinfo{person}{Nicholas Carlini}, \bibinfo{person}{Zizhao Zhang}, \bibinfo{person}{Han Zhang}, \bibinfo{person}{Colin~A Raffel}, \bibinfo{person}{Ekin~Dogus Cubuk}, \bibinfo{person}{Alexey Kurakin}, {and} \bibinfo{person}{Chun-Liang Li}.} \bibinfo{year}{2020}\natexlab{}.
\newblock \showarticletitle{Fixmatch: Simplifying semi-supervised learning with consistency and confidence}.
\newblock \bibinfo{journal}{\emph{NIPS}}  \bibinfo{volume}{33} (\bibinfo{year}{2020}), \bibinfo{pages}{596--608}.
\newblock


\bibitem[Stein et~al\mbox{.}(2019)]%
        {stein2019analysis}
\bibfield{author}{\bibinfo{person}{Roger~Alan Stein}, \bibinfo{person}{Patricia~A Jaques}, {and} \bibinfo{person}{Joao~Francisco Valiati}.} \bibinfo{year}{2019}\natexlab{}.
\newblock \showarticletitle{An analysis of hierarchical text classification using word embeddings}.
\newblock \bibinfo{journal}{\emph{Information Sciences}}  \bibinfo{volume}{471} (\bibinfo{year}{2019}), \bibinfo{pages}{216--232}.
\newblock


\bibitem[Sun and Lim(2001)]%
        {sun2001hierarchical}
\bibfield{author}{\bibinfo{person}{Aixin Sun} {and} \bibinfo{person}{Ee-Peng Lim}.} \bibinfo{year}{2001}\natexlab{}.
\newblock \showarticletitle{Hierarchical text classification and evaluation}. In \bibinfo{booktitle}{\emph{Proceedings 2001 IEEE International Conference on Data Mining}}. IEEE, \bibinfo{pages}{521--528}.
\newblock


\bibitem[Tang and Jia(2022)]%
        {tang2022towards}
\bibfield{author}{\bibinfo{person}{Hui Tang} {and} \bibinfo{person}{Kui Jia}.} \bibinfo{year}{2022}\natexlab{}.
\newblock \showarticletitle{Towards Discovering the Effectiveness of Moderately Confident Samples for Semi-Supervised Learning}. In \bibinfo{booktitle}{\emph{CVPR}}. \bibinfo{pages}{14658--14667}.
\newblock


\bibitem[Tang et~al\mbox{.}(2025)]%
        {tang2025semantic}
\bibfield{author}{\bibinfo{person}{Yingjie Tang}, \bibinfo{person}{Shou Feng}, \bibinfo{person}{Chunhui Zhao}, \bibinfo{person}{Yongqi Chen}, \bibinfo{person}{Zhiyong Lv}, {and} \bibinfo{person}{Weiwei Sun}.} \bibinfo{year}{2025}\natexlab{}.
\newblock \showarticletitle{A Semantic Change Detection Network Based on Boundary Detection and Task Interaction for High-Resolution Remote Sensing Images}.
\newblock \bibinfo{journal}{\emph{IEEE Transactions on Neural Networks and Learning Systems}} (\bibinfo{year}{2025}).
\newblock


\bibitem[Tarvainen and Valpola(2017)]%
        {tarvainen2017mean}
\bibfield{author}{\bibinfo{person}{Antti Tarvainen} {and} \bibinfo{person}{Harri Valpola}.} \bibinfo{year}{2017}\natexlab{}.
\newblock \showarticletitle{Mean teachers are better role models: Weight-averaged consistency targets improve semi-supervised deep learning results}.
\newblock \bibinfo{journal}{\emph{NIPS}}  \bibinfo{volume}{30} (\bibinfo{year}{2017}).
\newblock


\bibitem[Tian et~al\mbox{.}(2024)]%
        {tian2024unite}
\bibfield{author}{\bibinfo{person}{Yuxin Tian}, \bibinfo{person}{Yijie Lin}, \bibinfo{person}{Qing Ye}, \bibinfo{person}{Jian Wang}, \bibinfo{person}{Xi Peng}, {and} \bibinfo{person}{Jiancheng Lv}.} \bibinfo{year}{2024}\natexlab{}.
\newblock \showarticletitle{UNITE: multitask learning with sufficient feature for dense prediction}.
\newblock \bibinfo{journal}{\emph{IEEE Transactions on Systems, Man, and Cybernetics: Systems}} \bibinfo{volume}{54}, \bibinfo{number}{8} (\bibinfo{year}{2024}), \bibinfo{pages}{5012--5024}.
\newblock


\bibitem[Vandenhende et~al\mbox{.}(2021)]%
        {vandenhende2021multi}
\bibfield{author}{\bibinfo{person}{Simon Vandenhende}, \bibinfo{person}{Stamatios Georgoulis}, \bibinfo{person}{Wouter Van~Gansbeke}, \bibinfo{person}{Marc Proesmans}, \bibinfo{person}{Dengxin Dai}, {and} \bibinfo{person}{Luc Van~Gool}.} \bibinfo{year}{2021}\natexlab{}.
\newblock \showarticletitle{Multi-task learning for dense prediction tasks: A survey}.
\newblock \bibinfo{journal}{\emph{PAMI}} (\bibinfo{year}{2021}).
\newblock


\bibitem[Vandenhende et~al\mbox{.}(2020)]%
        {vandenhende2020mti}
\bibfield{author}{\bibinfo{person}{Simon Vandenhende}, \bibinfo{person}{Stamatios Georgoulis}, {and} \bibinfo{person}{Luc Van~Gool}.} \bibinfo{year}{2020}\natexlab{}.
\newblock \showarticletitle{Mti-net: Multi-scale task interaction networks for multi-task learning}. In \bibinfo{booktitle}{\emph{ECCV}}. Springer, \bibinfo{pages}{527--543}.
\newblock


\bibitem[Wang et~al\mbox{.}(2025a)]%
        {wang2025towards}
\bibfield{author}{\bibinfo{person}{Yizhou Wang}, \bibinfo{person}{Kuan-Chuan Peng}, {and} \bibinfo{person}{Yun Fu}.} \bibinfo{year}{2025}\natexlab{a}.
\newblock \showarticletitle{Towards zero-shot 3d anomaly localization}. In \bibinfo{booktitle}{\emph{2025 IEEE/CVF Winter Conference on Applications of Computer Vision (WACV)}}. IEEE, \bibinfo{pages}{1447--1456}.
\newblock


\bibitem[Wang et~al\mbox{.}(2022a)]%
        {wang2022making}
\bibfield{author}{\bibinfo{person}{Yizhou Wang}, \bibinfo{person}{Can Qin}, \bibinfo{person}{Yue Bai}, \bibinfo{person}{Yi Xu}, \bibinfo{person}{Xu Ma}, {and} \bibinfo{person}{Yun Fu}.} \bibinfo{year}{2022}\natexlab{a}.
\newblock \showarticletitle{Making reconstruction-based method great again for video anomaly detection}. In \bibinfo{booktitle}{\emph{2022 IEEE International Conference on Data Mining (ICDM)}}. IEEE, \bibinfo{pages}{1215--1220}.
\newblock


\bibitem[Wang et~al\mbox{.}(2022b)]%
        {wang2022self}
\bibfield{author}{\bibinfo{person}{Yizhou Wang}, \bibinfo{person}{Can Qin}, \bibinfo{person}{Rongzhe Wei}, \bibinfo{person}{Yi Xu}, \bibinfo{person}{Yue Bai}, {and} \bibinfo{person}{Yun Fu}.} \bibinfo{year}{2022}\natexlab{b}.
\newblock \showarticletitle{Self-supervision meets adversarial perturbation: A novel framework for anomaly detection}. In \bibinfo{booktitle}{\emph{Proceedings of the 31st ACM International Conference on Information \& Knowledge Management}}. \bibinfo{pages}{4555--4559}.
\newblock


\bibitem[Wang et~al\mbox{.}(2024)]%
        {wang2024sla}
\bibfield{author}{\bibinfo{person}{Yizhou Wang}, \bibinfo{person}{Can Qin}, \bibinfo{person}{Rongzhe Wei}, \bibinfo{person}{Yi Xu}, \bibinfo{person}{Yue Bai}, {and} \bibinfo{person}{Yun Fu}.} \bibinfo{year}{2024}\natexlab{}.
\newblock \showarticletitle{Sla ˆ2 p: Self-supervised anomaly detection with adversarial perturbation}.
\newblock \bibinfo{journal}{\emph{IEEE Transactions on Knowledge and Data Engineering}} (\bibinfo{year}{2024}).
\newblock


\bibitem[Wang et~al\mbox{.}(2022c)]%
        {wang2022semi}
\bibfield{author}{\bibinfo{person}{Yufeng Wang}, \bibinfo{person}{Yi-Hsuan Tsai}, \bibinfo{person}{Wei-Chih Hung}, \bibinfo{person}{Wenrui Ding}, \bibinfo{person}{Shuo Liu}, {and} \bibinfo{person}{Ming-Hsuan Yang}.} \bibinfo{year}{2022}\natexlab{c}.
\newblock \showarticletitle{Semi-supervised multi-task learning for semantics and depth}. In \bibinfo{booktitle}{\emph{WACV}}. \bibinfo{pages}{2505--2514}.
\newblock


\bibitem[Wang et~al\mbox{.}(2025b)]%
        {wang2025cautious}
\bibfield{author}{\bibinfo{person}{Yizhou Wang}, \bibinfo{person}{Lingzhi Zhang}, \bibinfo{person}{Yue Bai}, \bibinfo{person}{Mang~Tik Chiu}, \bibinfo{person}{Zhengmian Hu}, \bibinfo{person}{Mingyuan Zhang}, \bibinfo{person}{Qihua Dong}, \bibinfo{person}{Yu Yin}, \bibinfo{person}{Sohrab Amirghodsi}, {and} \bibinfo{person}{Yun Fu}.} \bibinfo{year}{2025}\natexlab{b}.
\newblock \showarticletitle{Cautious Next Token Prediction}.
\newblock \bibinfo{journal}{\emph{arXiv preprint arXiv:2507.03038}} (\bibinfo{year}{2025}).
\newblock


\bibitem[Wang et~al\mbox{.}(2023)]%
        {wang2023vaquita}
\bibfield{author}{\bibinfo{person}{Yizhou Wang}, \bibinfo{person}{Ruiyi Zhang}, \bibinfo{person}{Haoliang Wang}, \bibinfo{person}{Uttaran Bhattacharya}, \bibinfo{person}{Yun Fu}, {and} \bibinfo{person}{Gang Wu}.} \bibinfo{year}{2023}\natexlab{}.
\newblock \showarticletitle{Vaquita: Enhancing alignment in llm-assisted video understanding}.
\newblock \bibinfo{journal}{\emph{arXiv preprint arXiv:2312.02310}} (\bibinfo{year}{2023}).
\newblock


\bibitem[Xie et~al\mbox{.}(2020)]%
        {xie2020self}
\bibfield{author}{\bibinfo{person}{Qizhe Xie}, \bibinfo{person}{Minh-Thang Luong}, \bibinfo{person}{Eduard Hovy}, {and} \bibinfo{person}{Quoc~V Le}.} \bibinfo{year}{2020}\natexlab{}.
\newblock \showarticletitle{Self-training with noisy student improves imagenet classification}. In \bibinfo{booktitle}{\emph{Proceedings of the IEEE/CVF conference on computer vision and pattern recognition}}. \bibinfo{pages}{10687--10698}.
\newblock


\bibitem[Xu et~al\mbox{.}(2018)]%
        {xu2018pad}
\bibfield{author}{\bibinfo{person}{Dan Xu}, \bibinfo{person}{Wanli Ouyang}, \bibinfo{person}{Xiaogang Wang}, {and} \bibinfo{person}{Nicu Sebe}.} \bibinfo{year}{2018}\natexlab{}.
\newblock \showarticletitle{Pad-net: Multi-tasks guided prediction-and-distillation network for simultaneous depth estimation and scene parsing}. In \bibinfo{booktitle}{\emph{CVPR}}. \bibinfo{pages}{675--684}.
\newblock


\bibitem[Yang et~al\mbox{.}(2023a)]%
        {yang2023revisiting}
\bibfield{author}{\bibinfo{person}{Lihe Yang}, \bibinfo{person}{Lei Qi}, \bibinfo{person}{Litong Feng}, \bibinfo{person}{Wayne Zhang}, {and} \bibinfo{person}{Yinghuan Shi}.} \bibinfo{year}{2023}\natexlab{a}.
\newblock \showarticletitle{Revisiting weak-to-strong consistency in semi-supervised semantic segmentation}. In \bibinfo{booktitle}{\emph{Proceedings of the IEEE/CVF conference on computer vision and pattern recognition}}. \bibinfo{pages}{7236--7246}.
\newblock


\bibitem[Yang et~al\mbox{.}(2025)]%
        {yang2025unimatch}
\bibfield{author}{\bibinfo{person}{Lihe Yang}, \bibinfo{person}{Zhen Zhao}, {and} \bibinfo{person}{Hengshuang Zhao}.} \bibinfo{year}{2025}\natexlab{}.
\newblock \showarticletitle{Unimatch v2: Pushing the limit of semi-supervised semantic segmentation}.
\newblock \bibinfo{journal}{\emph{IEEE Transactions on Pattern Analysis and Machine Intelligence}} (\bibinfo{year}{2025}).
\newblock


\bibitem[Yang et~al\mbox{.}(2023b)]%
        {yang2023contrastive}
\bibfield{author}{\bibinfo{person}{Siwei Yang}, \bibinfo{person}{Hanrong Ye}, {and} \bibinfo{person}{Dan Xu}.} \bibinfo{year}{2023}\natexlab{b}.
\newblock \showarticletitle{Contrastive multi-task dense prediction}. In \bibinfo{booktitle}{\emph{Proceedings of the AAAI Conference on Artificial Intelligence (AAAI)}}.
\newblock


\bibitem[Yang et~al\mbox{.}({[n.\,d.]})]%
        {yangmulti}
\bibfield{author}{\bibinfo{person}{Yuqi Yang}, \bibinfo{person}{Peng-Tao Jiang}, \bibinfo{person}{Qibin Hou}, \bibinfo{person}{Hao Zhang}, \bibinfo{person}{Jinwei Chen}, {and} \bibinfo{person}{Bo Li}.} \bibinfo{year}{[n.\,d.]}\natexlab{}.
\newblock \showarticletitle{Multi-Task Dense Predictions via Unleashing the Power of Diffusion}. In \bibinfo{booktitle}{\emph{The Thirteenth International Conference on Learning Representations}}.
\newblock


\bibitem[Ye and Xu(2022)]%
        {ye2022inverted}
\bibfield{author}{\bibinfo{person}{Hanrong Ye} {and} \bibinfo{person}{Dan Xu}.} \bibinfo{year}{2022}\natexlab{}.
\newblock \showarticletitle{Inverted Pyramid Multi-task Transformer for Dense Scene Understanding}.
\newblock \bibinfo{journal}{\emph{ECCV}} (\bibinfo{year}{2022}).
\newblock


\bibitem[Ye and Xu(2023a)]%
        {ye2023taskexpert}
\bibfield{author}{\bibinfo{person}{Hanrong Ye} {and} \bibinfo{person}{Dan Xu}.} \bibinfo{year}{2023}\natexlab{a}.
\newblock \showarticletitle{TaskExpert: Dynamically Assembling Multi-Task Representations with Memorial Mixture-of-Experts}. In \bibinfo{booktitle}{\emph{Proceedings of the IEEE/CVF International Conference on Computer Vision}}. \bibinfo{pages}{21828--21837}.
\newblock


\bibitem[Ye and Xu(2023b)]%
        {taskprompter2023}
\bibfield{author}{\bibinfo{person}{Hanrong Ye} {and} \bibinfo{person}{Dan Xu}.} \bibinfo{year}{2023}\natexlab{b}.
\newblock \showarticletitle{TaskPrompter: Spatial-Channel Multi-Task Prompting for Dense Scene Understanding}. In \bibinfo{booktitle}{\emph{ICLR}}.
\newblock


\bibitem[Ye and Xu(2024)]%
        {ye2024diffusionmtl}
\bibfield{author}{\bibinfo{person}{Hanrong Ye} {and} \bibinfo{person}{Dan Xu}.} \bibinfo{year}{2024}\natexlab{}.
\newblock \showarticletitle{DiffusionMTL: Learning Multi-Task Denoising Diffusion Model from Partially Annotated Data}. In \bibinfo{booktitle}{\emph{CVPR}}.
\newblock


\bibitem[Zamir et~al\mbox{.}(2020)]%
        {zamir2020robust}
\bibfield{author}{\bibinfo{person}{Amir~R Zamir}, \bibinfo{person}{Alexander Sax}, \bibinfo{person}{Nikhil Cheerla}, \bibinfo{person}{Rohan Suri}, \bibinfo{person}{Zhangjie Cao}, \bibinfo{person}{Jitendra Malik}, {and} \bibinfo{person}{Leonidas~J Guibas}.} \bibinfo{year}{2020}\natexlab{}.
\newblock \showarticletitle{Robust learning through cross-task consistency}. In \bibinfo{booktitle}{\emph{CVPR}}. \bibinfo{pages}{11197--11206}.
\newblock


\bibitem[Zeisl et~al\mbox{.}(2014)]%
        {zeisl2014discriminatively}
\bibfield{author}{\bibinfo{person}{Bernhard Zeisl}, \bibinfo{person}{Marc Pollefeys}, {et~al\mbox{.}}} \bibinfo{year}{2014}\natexlab{}.
\newblock \showarticletitle{Discriminatively trained dense surface normal estimation}. In \bibinfo{booktitle}{\emph{ECCV}}. Springer, \bibinfo{pages}{468--484}.
\newblock


\bibitem[Zeng et~al\mbox{.}(2019)]%
        {zeng2019joint}
\bibfield{author}{\bibinfo{person}{Yu Zeng}, \bibinfo{person}{Yunzhi Zhuge}, \bibinfo{person}{Huchuan Lu}, {and} \bibinfo{person}{Lihe Zhang}.} \bibinfo{year}{2019}\natexlab{}.
\newblock \showarticletitle{Joint learning of saliency detection and weakly supervised semantic segmentation}. In \bibinfo{booktitle}{\emph{ICCV}}. \bibinfo{pages}{7223--7233}.
\newblock


\bibitem[Zhang et~al\mbox{.}(2023)]%
        {zhang2023rethinking}
\bibfield{author}{\bibinfo{person}{Jingdong Zhang}, \bibinfo{person}{Jiayuan Fan}, \bibinfo{person}{Peng Ye}, \bibinfo{person}{Bo Zhang}, \bibinfo{person}{Hancheng Ye}, \bibinfo{person}{Baopu Li}, \bibinfo{person}{Yancheng Cai}, {and} \bibinfo{person}{Tao Chen}.} \bibinfo{year}{2023}\natexlab{}.
\newblock \showarticletitle{Rethinking of Feature Interaction for Multi-task Learning on Dense Prediction}.
\newblock \bibinfo{journal}{\emph{arXiv preprint arXiv:2312.13514}} (\bibinfo{year}{2023}).
\newblock


\bibitem[Zhang et~al\mbox{.}(2025)]%
        {zhang2025bridgenet}
\bibfield{author}{\bibinfo{person}{Jingdong Zhang}, \bibinfo{person}{Jiayuan Fan}, \bibinfo{person}{Peng Ye}, \bibinfo{person}{Bo Zhang}, \bibinfo{person}{Hancheng Ye}, \bibinfo{person}{Baopu Li}, \bibinfo{person}{Yancheng Cai}, {and} \bibinfo{person}{Tao Chen}.} \bibinfo{year}{2025}\natexlab{}.
\newblock \showarticletitle{BridgeNet: Comprehensive and Effective Feature Interactions via Bridge Feature for Multi-Task Dense Predictions}.
\newblock \bibinfo{journal}{\emph{IEEE Transactions on Pattern Analysis and Machine Intelligence}} (\bibinfo{year}{2025}).
\newblock


\bibitem[Zhang et~al\mbox{.}(2018)]%
        {zhang2018joint}
\bibfield{author}{\bibinfo{person}{Zhenyu Zhang}, \bibinfo{person}{Zhen Cui}, \bibinfo{person}{Chunyan Xu}, \bibinfo{person}{Zequn Jie}, \bibinfo{person}{Xiang Li}, {and} \bibinfo{person}{Jian Yang}.} \bibinfo{year}{2018}\natexlab{}.
\newblock \showarticletitle{Joint task-recursive learning for semantic segmentation and depth estimation}. In \bibinfo{booktitle}{\emph{ECCV}}. \bibinfo{pages}{235--251}.
\newblock


\bibitem[Zhang et~al\mbox{.}(2019)]%
        {zhang2019pattern}
\bibfield{author}{\bibinfo{person}{Zhenyu Zhang}, \bibinfo{person}{Zhen Cui}, \bibinfo{person}{Chunyan Xu}, \bibinfo{person}{Yan Yan}, \bibinfo{person}{Nicu Sebe}, {and} \bibinfo{person}{Jian Yang}.} \bibinfo{year}{2019}\natexlab{}.
\newblock \showarticletitle{Pattern-affinitive propagation across depth, surface normal and semantic segmentation}. In \bibinfo{booktitle}{\emph{CVPR}}. \bibinfo{pages}{4106--4115}.
\newblock


\bibitem[Zhou et~al\mbox{.}(2020)]%
        {zhou2020hierarchy}
\bibfield{author}{\bibinfo{person}{Jie Zhou}, \bibinfo{person}{Chunping Ma}, \bibinfo{person}{Dingkun Long}, \bibinfo{person}{Guangwei Xu}, \bibinfo{person}{Ning Ding}, \bibinfo{person}{Haoyu Zhang}, \bibinfo{person}{Pengjun Xie}, {and} \bibinfo{person}{Gongshen Liu}.} \bibinfo{year}{2020}\natexlab{}.
\newblock \showarticletitle{Hierarchy-aware global model for hierarchical text classification}. In \bibinfo{booktitle}{\emph{Proceedings of the 58th annual meeting of the association for computational linguistics}}. \bibinfo{pages}{1106--1117}.
\newblock


\bibitem[Zhu et~al\mbox{.}(2020)]%
        {zhu2020deformable}
\bibfield{author}{\bibinfo{person}{Xizhou Zhu}, \bibinfo{person}{Weijie Su}, \bibinfo{person}{Lewei Lu}, \bibinfo{person}{Bin Li}, \bibinfo{person}{Xiaogang Wang}, {and} \bibinfo{person}{Jifeng Dai}.} \bibinfo{year}{2020}\natexlab{}.
\newblock \showarticletitle{Deformable detr: Deformable transformers for end-to-end object detection}.
\newblock \bibinfo{journal}{\emph{arXiv preprint arXiv:2010.04159}} (\bibinfo{year}{2020}).
\newblock


\bibitem[Zou et~al\mbox{.}(2018)]%
        {zou2018unsupervised}
\bibfield{author}{\bibinfo{person}{Yang Zou}, \bibinfo{person}{Zhiding Yu}, \bibinfo{person}{BVK Kumar}, {and} \bibinfo{person}{Jinsong Wang}.} \bibinfo{year}{2018}\natexlab{}.
\newblock \showarticletitle{Unsupervised domain adaptation for semantic segmentation via class-balanced self-training}. In \bibinfo{booktitle}{\emph{ECCV}}. \bibinfo{pages}{289--305}.
\newblock


\bibitem[Zou et~al\mbox{.}(2019)]%
        {zou2019confidence}
\bibfield{author}{\bibinfo{person}{Yang Zou}, \bibinfo{person}{Zhiding Yu}, \bibinfo{person}{Xiaofeng Liu}, \bibinfo{person}{BVK Kumar}, {and} \bibinfo{person}{Jinsong Wang}.} \bibinfo{year}{2019}\natexlab{}.
\newblock \showarticletitle{Confidence regularized self-training}. In \bibinfo{booktitle}{\emph{ICCV}}. \bibinfo{pages}{5982--5991}.
\newblock


\end{thebibliography}

\newpage
\clearpage

\textbf{\huge Supplementary}
\appendix

\vspace{10mm}

In this supplementary document, we present: (i) more details about the structure of hierarchical task tokens and training pipeline, (ii) a more comprehensive explanation of experimental implementations, (iii) more quantitative and qualitative experimental results.

\section{Model Details}
\subsection{Training Pipeline}
\par We will discuss utilizing the discovered supervision signals by Hierarchical Task Tokens (HiTTs) for the training process in this section. We first train the multi-task model with HiTTs on all labeled task data. Then the weights of the network are fixed, and used to produce hierarchical supervision on both feature and prediction spaces. We produce the pseudo label in an offline manner according to~\cite{xie2020self}, which is labeling on clean image without data augmentation, and training on augmented images and pseudo labels to enforce consistent predictions. In~\cite{xie2020self}, this method is only applied to classification tasks while we extend the utilization to general dense prediction tasks. After the pseudo labels are produced, we use them along with the ground-truth labels to jointly train the multi-task model from scratch. Additionally, we also use the pseudo feature supervision produced by this pre-trained multi-task model for feature regularization during the optimization process. Both hierarchies of the discovered supervision signals ensure that all task predictions will obtain pixel-wise supervision for multi-task optimization to gain better generalization ability on unlabeled data.

\subsection{Global Task Token Learning}
In this section we are going to introduce the inter- and intra-task attention in detail. Given an RGB input $\boldsymbol{X} \in \mathbb{R}^{3 \times H \times W}$, a multi-task dense prediction framework firstly produces a task-generic representation $\boldsymbol{F_{s}} \in \mathbb{R}^{C \times h \times w}$ through a shared encoder. Considering we have $T$ tasks, and we target decoding task features $\{\boldsymbol{F_{1}}, \boldsymbol{F_{2}}, \cdots, \boldsymbol{F_{T}}\}$ from $\boldsymbol{F_{s}}$ as well as learning representative global task tokens $\{\boldsymbol{\theta_{1}}, \boldsymbol{\theta_{2}}, \cdots, \boldsymbol{\theta_{T}}\}$ for each task. 

As shown in Fig.~\ref{fig:GTTL_detail}: i) Inter-Task Learning, which aims to learn explicit global cross-task token affinities $\boldsymbol{A}$, and conduct cross-task interaction accordingly for the global token learning process. ii) Intra-task Learning, which learns task-specific information by globally conducting self-attention between task feature and token pairs.

Firstly, we flatten the shared feature $\boldsymbol{F_{s}}$ into feature tokens with shape $\mathbb{R}^{C \times (h w)}$, and use each global task token to query the shared feature to obtain each task feature $\boldsymbol{F_{i}}$ accordingly. Then, to conduct \textit{inter-task attention}, all global task tokens are used to produce cross-task affinities that explicitly guide the learning process. The cross-task affinity map  $\boldsymbol{A} \in \mathbb{R}^{T \times T}$ is calculated as:
\begin{equation}
\boldsymbol{A} = \operatorname{Softmax} ( \boldsymbol{Q} \times \boldsymbol{K}^{\top} ),
\end{equation}
\noindent where $\boldsymbol{Q}$ and $\boldsymbol{K}$ are individual linear projection of concatenated global tokens $\boldsymbol{\Theta} = [ \boldsymbol{\theta_{1}}; \boldsymbol{\theta_{2}}; \cdots; \boldsymbol{\theta_{T}}]^{\top}$, in which $[\cdot]$ indicates the concatenation. After affinity matrix $\boldsymbol{A}$ is calculated, it is used to conduct affine combinations of task features and global task tokens respectively:
\begin{equation}
\boldsymbol{\Theta^{\prime}} = \boldsymbol{A} \times \boldsymbol{\Theta}, \quad
\boldsymbol{\mathcal{F}^{\prime}} = \boldsymbol{A} \times \boldsymbol{\mathcal{F}},
\end{equation}
\noindent and similarly, $\boldsymbol{\mathcal{F}} = [\boldsymbol{F_{1}}; \boldsymbol{F_{2}}; \cdots; \boldsymbol{F_{T}}]^{\top}$, and $\boldsymbol{\Theta^{\prime}}$, $\boldsymbol{\mathcal{F}^{\prime}}$ represents all updated task tokens and features after the affine combinations. For each task feature $\boldsymbol{F_{i}}$, if it is not directly supervised by labels, the feature will be less representative and contain more noise. Thus, conducting affine combinations among all tasks ensures that the task-shared representations from labeled tasks are able to fertilize the unlabeled task features.

Afterward, the updated tokens and features with cross-task information are involved in the \textit{intra-task attention}, where we first concatenate every corresponding task token and feature, and perform self-attention on the spatial dimension among each token-feature pair $\left [ \boldsymbol{\theta_{i}^{\prime}}; \boldsymbol{F_{i}^{\prime}} \right ] \in \mathbb{R}^{C \times (h w + 1)}$. The global task tokens will further learn more specific and discriminative task representations during this process, and representative task tokens will in turn enhance the feature quality as well. Followingly, we discover pseudo feature supervision with the aid of well-learned global task tokens $\boldsymbol{\theta_{i}^{\prime}}$.

Additionally, for multi-scale backbone features, directly fusing them ignores the various granularity of task representations maintained at different scales. Thus, for multi-scale image backbone, we further propose Multi-scale Global Task Token Learning in order to learn comprehensive multi-scale task relations. The proposed method involves \textit{inter-task attention} separately at each scale, and then the multi-scale features and tokens are fused before \textit{intra-task attention}. In this way, the global task tokens gain richer cross-task relations at different scales and are able to maintain stronger representations. We will illustrate this part in detail in Sec.~\ref{sec:MS-GTTL}.

\begin{figure*}[t]
  \centering
  \includegraphics[width=0.95\linewidth]{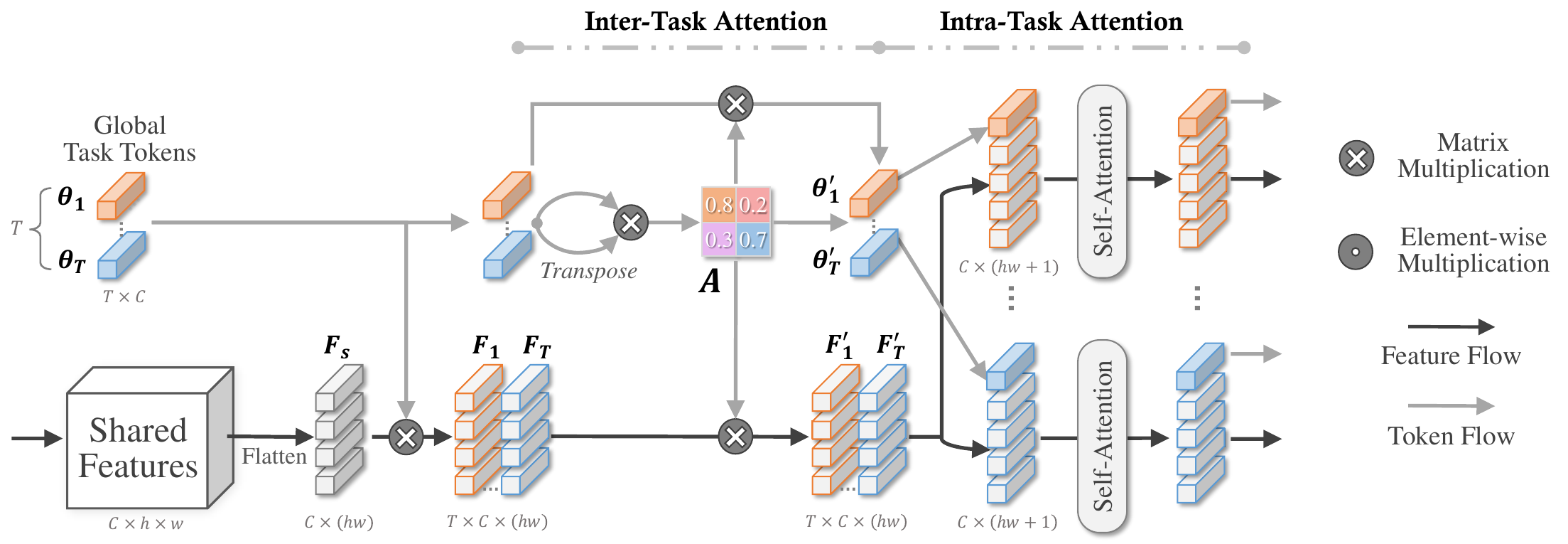}
  \caption{Illustration of detail designs of: (i) Inter-Task Attention: predicting cross-task token affinities $\boldsymbol{A}$ from the global task tokens. (ii) Intra-Task Attention: conducting self-attention between task features and tokens. }
  \label{fig:GTTL_detail}
\end{figure*}

\subsection{Discrete Quantization and Task Losses}
\par In Sec. 3 of the body part, we have discussed how to learn HiTTs. For continuous regression tasks, such as Depth and Normal, we first need to perform a discrete quantization of the label space to provide discriminative supervision for tokens. The goal for quantization is to assign meaningful category bins to each fine-grained token for classification. As analyzed in~\cite{bruggemann2021exploring}, this quantization only changes the way of predicting regression task, but does not contribute to the learning performance. For depth estimation, we follow the setting in~\cite{bruggemann2021exploring,li2018monocular}, and divide the range of depth values into several logarithmic bins. Our predicted task logits score $\boldsymbol{G_i^{\prime}}$ is used to calculate the soft-weighted sum with each bin and produce final task predictions accordingly. For surface normal estimation, we follow~\cite{bruggemann2021exploring,zeisl2014discriminatively} and use K-means to learn several unit normal vectors, which serve as clustering centers, and they are also used to generate predictions with $\boldsymbol{G_i^{\prime}}$. This process can be expressed as:

\begin{equation}
\boldsymbol{\hat{Y_{i}}} = \operatorname{Sum}\left (\boldsymbol{c_{i}}^{\top} \times \operatorname{Softmax} (\boldsymbol{G_{i}^{\prime}}) \right),
\end{equation}

\noindent where $\boldsymbol{c_{i}} \in \mathbb{R}^{C_p}$ represents the center of each bin. In our experiments, the numbers of depth bins and normal cluster centers on NYUD-v2 are $30$ and $20$, respectively. For Cityscapes, we consider $100$ depth bins as the Cityscapes dataset is captured from outdoor scenarios and have more significant changes in depth. For PASCAL-Context, we select $40$ different unit verctors uniformly from the space to serve as normal cluster centers.

To supervise the task predictions, we can directly impose regression losses on $\boldsymbol{\hat{Y_{i}}}$ for Depth. and Normal:

\begin{equation}
\operatorname{L_i^{reg}}\left( \boldsymbol{\hat{Y_{i}}}, {\boldsymbol{Y_{i}}} \right) ,
\end{equation}

\noindent where $\operatorname{L_i^{reg}}(\cdot)$ can be $\operatorname{L_1}$ loss or Angle loss for Depth. or Normal. respectively. Furthermore, in order to gain more discriminative task category information for each token, we also impose classification loss on $\boldsymbol{G_{i}^{\prime}}$. We first extract the corresponding one-hot label from $\boldsymbol{Y_{i}}$ by $\boldsymbol{c_{i}}$, denoting as $\boldsymbol{Y_{i}^{oh}}$, and the loss can be written as:

\begin{equation}
\operatorname{L_i^{cls}}\left( \boldsymbol{G_{i}}, {\boldsymbol{Y_{i}^{oh}}} \right) = - \boldsymbol{Y_{i}^{oh}} \cdot \operatorname{log} \left( \operatorname{Softmax}(\boldsymbol{G_{i}^{\prime}}) \right) ,
\end{equation}

\noindent and the overall \textbf{mixture loss} for each task can be written as:

\begin{equation}
\operatorname{L_i}(\cdot) = \lambda_i \operatorname{L_i^{reg}}\left( \boldsymbol{\hat{Y_{i}}}, {\boldsymbol{Y_{i}}} \right) + \operatorname{L_i^{cls}}\left( \boldsymbol{G_{i}^{\prime}}, {\boldsymbol{Y_{i}^{oh}}} \right), 
\end{equation}

\noindent where $\lambda_i = 0.1$ if task $i$ is a regression task itself, otherwise $\lambda_i = 0$.

\subsection{Multi-scale Global Task Token Learning}
\label{sec:MS-GTTL}
For the implementations on NYUD-v2 and Cityscapes, we use the SegNet~\cite{badrinarayanan2017segnet} to serve as the shared backbone, which produces single-scale shared features for per-task decoding. However, on PASCAL-Context, we use ResNet-18~\cite{he2016deep} as the shared backbone, which can produce multi-scale shared features for decoding. Directly fusing them ignores the various task information maintained in different scales. Thus, we propose to learn the global task tokens on different scales in order to learn more comprehensive task relations.

As shown in Fig.~\ref{fig:MS_GTTL}, compared with the single-scale global task token learning process, the multi-scale process involves \textit{inter-task learning} separately on each scale, and then the multi-scale features and tokens are fused before \textit{intra-task learning}. The multi-scale backbone features $\left\{\boldsymbol{F_{s}}^{(j)} \right\}, j = 0, 1, 2, 3$ are first flattened on each scale, and each global task token $\boldsymbol{\theta_{i}}$ is projected by a linear layer to produce multi-scale tokens:

\begin{equation}
\boldsymbol{\theta_{i}}^{(j)} = \boldsymbol{W^{s \to m}_{i}}^{(j)} \times \boldsymbol{\theta_{i}}, \quad j = 0, 1, 2, 3,
\end{equation}

\noindent where $\boldsymbol{W^{s \to m}_{i}}^{(j)}$
Then, for every feature $\boldsymbol{F_{s}}^{(j)}$ and token $\boldsymbol{\theta_{i}}^{(j)}$ on scale $j$ and task $i$, we query $\boldsymbol{F_{s}}^{(j)}$ to obtain task features $\left\{\boldsymbol{F_{i}}^{(j)} \right\}, i=1, 2, \cdots, T; j = 0, 1, 2, 3$. After that, on each scale, we concatenate features and tokens from every task for \textit{Inter-task Learning} similar to Sec. 3.1:

\vspace{-3mm}
\begin{align}
\boldsymbol{\mathcal{F}}^{(j)} &= \left[\boldsymbol{F_{1}}^{(j)}; \boldsymbol{F_{2}}^{(j)}; \cdots; \boldsymbol{F_{T}}^{(j)}\right]^{\top}, \\
\boldsymbol{\Theta}^{(j)} &= \left[\boldsymbol{\theta_{1}}^{(j)}; \boldsymbol{\theta_{2}}^{(j)}; \cdots; \boldsymbol{\theta_{T}}^{(j)}\right]^{\top}.
\end{align}

Subsequently, $\boldsymbol{\mathcal{F}}^{(j)}$ and $\boldsymbol{\Theta}^{(j)}$ are used for inter-task learning on each scale. We denote the features and tokens after intra-task learning as $\boldsymbol{\mathcal{F}^{\prime}}^{(j)}$ and $\boldsymbol{\Theta^{\prime}}^{(j)}$. We fused them to share cross-task information on each scale:

\begin{align}
\boldsymbol{F_{i}^{\prime}} &= \operatorname{Conv_{i}^{1 \times 1}} \left(\left[\boldsymbol{F_{i}^{\prime}}^{(0)}; \boldsymbol{F_{i}^{\prime}}^{(1)};
\boldsymbol{F_{i}^{\prime}}^{(2)}; \boldsymbol{F_{i}^{\prime}}^{(3)}\right]\right), \\
\boldsymbol{\theta_{i}^{\prime}} &= 
\sum_{j=0}^{3}\left( \boldsymbol{W^{m \to s}_{i}}^{(j)} \times \boldsymbol{\theta_{i}^{\prime}}^{(j)} \right).
\end{align}

Finally, on each task $i$, the updated task features $\boldsymbol{F_{i}^{\prime}}$ and global task tokens $\boldsymbol{\theta_{i}^{\prime}}$ are used for \textit{Intra-task Learning}. The process is the same as Sec. 3.1.

In this way, we achieve learning global task tokens on multi-scale task features, which gains richer cross-task relations on different scales and maintains stronger representations in the global task tokens.

\begin{figure*}[t]
  \centering
  \includegraphics[width=1.0\linewidth]{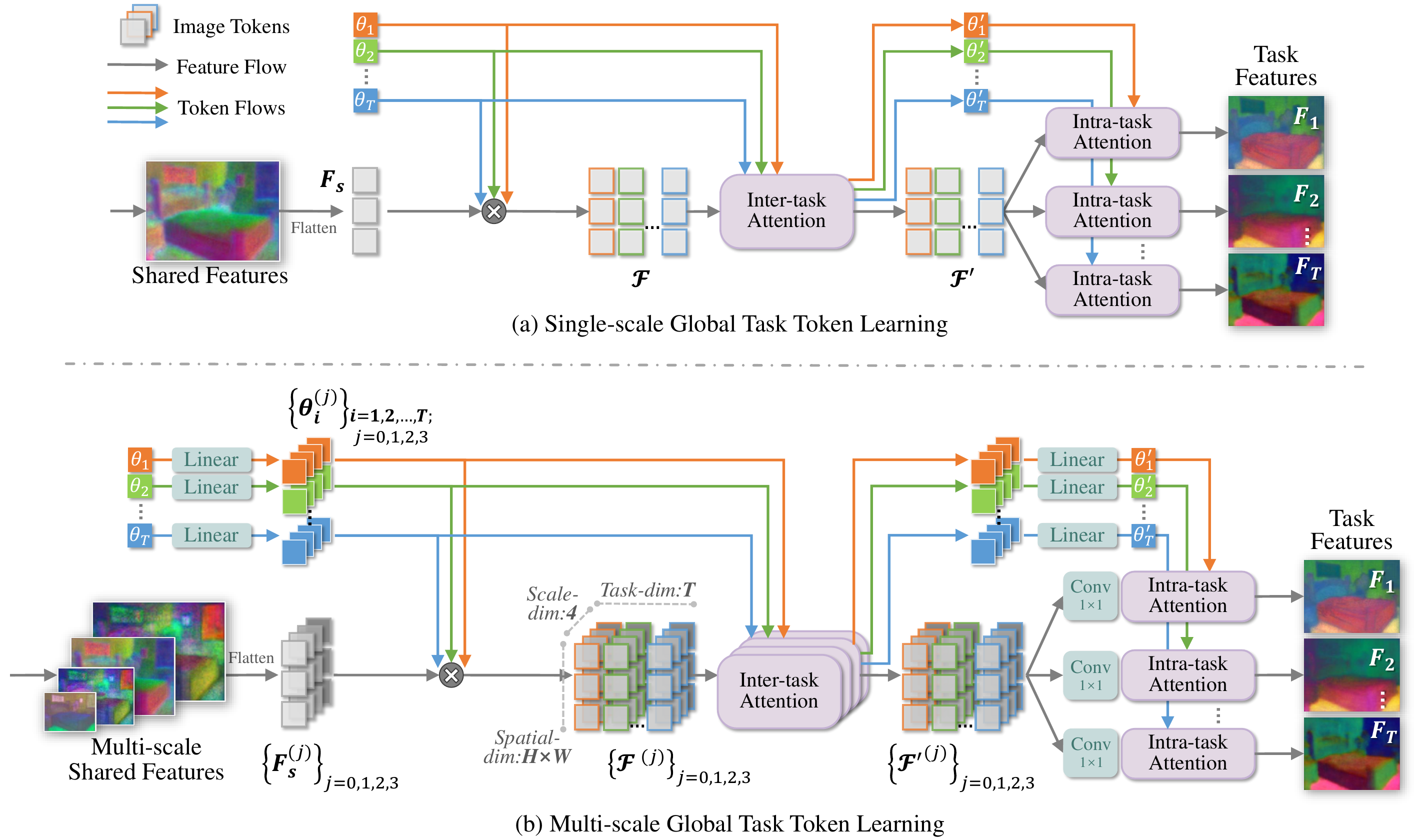}
  \vspace{-2mm}
  \caption{Illustrations of \textit{Single-scale} Global Task Token Learning (a) and \textit{Multi-scale} Global Task Token Learning (b). For the multi-scale features produced by the shared backbone, we use linear layers to produce corresponding task tokens for each scale. Then, in each scale, we query task features and conduct \textit{inter-task attention} to gain multi-task multi-scale representations. The multi-scale features and tokens are fused before \textit{intra-task attention} to transfer cross-scale information.}
  \label{fig:MS_GTTL}
\end{figure*}

\subsection{Broader Applicability and Future Directions}
The hierarchical design of our HiTTs is highly generalizable and not limited to the currently studied tasks. Its principles can be readily extended to other fundamental computer vision tasks, such as object detection~\cite{bochkovskiy2020yolov4,zhu2020deformable,cai2023rethinking,ren2024dino,wang2022making} and instance segmentation~\cite{chen2023instance,cheng2022masked,schult2022mask3d,gu2024diffusioninst}, which fundamentally rely on rich, multi-scale pixel representations. The experiment on the PASCAL-Context dataset partially demonstrates this potential, where HiTTs effectively learns from both coarse-grained semantic labels and fine-grained human part annotations simultaneously, showcasing its robust joint-learning capabilities across different granularities.

Furthermore, the core concepts of HiTTs extend beyond the visual domain. Its ability to build cross-task relations upon multi-level representations makes it a promising approach for tasks like hierarchical text classification~\cite{sun2001hierarchical,zhou2020hierarchy,stein2019analysis,wang2024sla,wang2022self} or 3D scene-level parsing~\cite{peng2023openscene,wang2025towards,jaritz2019multi}. Moreover, the unified token-feature interaction mechanism, based on the Transformer architecture, is inherently compatible with diverse data types, paving the way for future exploration in multi-modal learning environments~\cite{lin2023video,liu2023visual,Qwen-VL,wang2023vaquita,wang2025cautious}. This adaptability underscores the broad potential of our hierarchical token-based approach for complex multi-task and multi-modal problems.

\section{Implementation Details}
\subsection{Dataset}
\noindent \textbf{PASCAL-Context.} PASCAL-Context~\cite{everingham2009pascal} contains 4998 and 5105 images for training and testing respectively, which also have pixel-level annotations for semantic segmentation, human-parts segmentation and semantic edge detection. Additionally, we also consider surface normal estimation and saliency detection distilled by~\cite{maninis2019attentive}. We use Adam optimizer with learning rate $2\times 10^{-5}$, and weight decay $1\times 10^{-6}$, and train for $100$ epochs with batch size $6$. We update the learning rate with polynomial strategy and $\gamma = 0.9$ for the power factor.

\noindent \textbf{NYUD-v2.} NYUD-v2~\cite{silberman2012indoor} contains 795 and 654 RGB-D indoor scene images for training and testing respectively. We use the 13-class semantic annotations which is defined in~\cite{couprie2013indoor}, the truth depth annotations recorded by Microsoft Kinect depth camera, and surface normal annotations which are produced in~\cite{eigen2015predicting}. Following the setting in~\cite{liu2019end,li2022learning}, we use $288 \times 384$ image resolution to speed up training. We use Adam optimizer with a learning rate of $1\times 10^{-4}$, and train for 400 epochs with batch size $8$. We update the learning rate every $100$ epoch with $\gamma = 0.5$ as the multiplying factor.

\noindent \textbf{Cityscapes.} Cityscapes~\cite{cordts2016cityscapes} contains 2975 and 500 street-view images for training and testing respectively. We used the projected 7-class semantic annotations from~\cite{liu2019end}, and disparity maps to serve as depth annotations. Following the setting in~\cite{liu2019end,li2022learning}, we use $128 \times 256$ image resolution to speed up training. The optimizer and learning rate scheduler are set as the same as NYUD-v2.

\subsection{Data Preprocessing}
We follow the setting of~\cite{li2022learning,liu2019end,vandenhende2021multi,ye2024diffusionmtl} to process training data. For NYUD-v2 and Cityscapes, we use random scaling, cropping and horizon flipping for data augmentation following~\cite{li2022learning,liu2019end}. For PASCAL-Context, we follow~\cite{ye2024diffusionmtl} and use random scaling, cropping, horizon flipping and photometric distortion for data augmentation. Apart from the two partial settings (\textbf{one-label} and \textbf{random-labels}) we mentioned in Sec.4.1, we also consider two extra settings: \textbf{(iii) full-labels}: each image has labels on every task; \textbf{(iv) few-shot:} one task has only very few labels while other tasks are fully supervised. Both of the extra settings will furthermore show the effectiveness of our method.

\subsection{Model Setting}
We use SegNet~\cite{badrinarayanan2017segnet} as the image backbone for our experiments on NYUD-v2 and Cityscapes; and we use ResNet-18 as the image backbone, Atrous Spatial Pyramid Pooling (ASPP)~\cite{chen2018encoder} as the task-specific decoding heads. For the threshold $\tau_i$ which selects binary mask $\boldsymbol{M_i^p}$, we use \{Semseg: 0.9, Depth: 0.45, Normal: 0.6 \} for NYUD-v2 {one-label} setting, \{Semseg: 0.9, Depth: 0.7, Normal: 0.7\} for NYUD-v2 {random-labels} setting. \{Semseg: 0.9, Depth: 0.5\} for Cityscapes {one-label} setting, and \{Semseg: 0.9, Parsing: 0.85, Normal: 0.7, Sal: 0.7, Edge: 0.9\} for PASCAL-Context {one-label} and {random-labels} settings.

\begin{table*}[t]
\begin{center}
 \caption{Comparison on NYUD-v2 under \textit{one-label} and \textit{random-labels} settings. Our method shows clear performance gain over three tasks, which is consistent with the visualization results.}
 \label{tab:as}
\setlength{\tabcolsep}{4.2mm}{\scalebox{0.8}{
\begin{tabular}{clcccccccccc}
 \toprule
\multirow{2}{*}{Setting}&\multirow{2}{*}{Model}&\multicolumn{2}{c}{Semseg.} &\multicolumn{2}{c}{Depth.} & \multicolumn{5}{c}{Normal.}&$\Delta_{M T L}$\\
\cmidrule(lr){3-4}\cmidrule(lr){5-6}\cmidrule(lr){7-11}
 & &\textit{mIoU}↑ &\textit{pAcc}↑ &\textit{AbS}↓&\textit{rmse}↓ &\textit{mErr}↓ &\textit{rmse}↓ & \textit{$\eta_{1}$}↑& \textit{$\eta_{2}$}↑& \textit{$\eta_{3}$}↑&(\%)↑ \\ \midrule
\multirow{5}*{\rotatebox{90}{One-Label}}&STL & 29.28	&55.41	&0.7182	&1.0151 	&30.1971 &37.7115 &23.1532 &46.4046 &58.5216& - \\
&MTL \textit{baseline} & 30.92	&58.23	&0.5982	&0.8544	&31.8509 &38.6313 &19.7083 &41.2614 &53.6381	&0.11\\
&MTAN~\cite{liu2019end}& 30.92	&57.14	&0.6196	&0.8477	&30.0278 &36.7808 &21.4199 &44.7805 &57.5720	&3.26\\
&XTC~\cite{li2022learning} &33.46	&60.95	&0.5728	&0.8056	&31.1492 &37.8211 &19.8410 &42.2268 &54.9997	&3.60\\
&Ours &\textbf{35.81}	&\textbf{63.22}	&\textbf{0.5540}	&\textbf{0.7939}	&\textbf{28.5131}	&\textbf{36.1738}	&\textbf{26.4985}	&\textbf{50.2357}	&\textbf{61.8343}& \textbf{13.23}\\
\midrule
\multirow{5}*{\rotatebox{90}{\fontsize{7}{12}\selectfont Random-Labels}}&STL & 34.49	&60.52	&0.6272	&0.8824	&27.9681 &34.9293 &24.6011 &49.7888 &62.4425&-\\
&MTL \textit{baseline} & 35.49	&61.81	&0.5503	&0.7874	&29.9541 &36.7726 &21.6933 &45.0412 &57.7516	&-1.47\\
&MTAN~\cite{liu2019end}&35.96	&61.64	&0.6120	&0.8272	&28.6933 &35.3528 &23.0253 &47.2287 &60.1113	&-0.48\\
&XTC~\cite{li2022learning} & 38.11	&64.37	&0.5387	&0.7755	&29.6549 &36.3992 &21.7058 &45.4801 &58.4236	&0.66\\
&Ours &\textbf{41.78}	&\textbf{66.50}	&\textbf{0.5177}	&\textbf{0.7472}	&\textbf{27.3488} &\textbf{34.6820} &\textbf{27.1619} &\textbf{51.8924} &\textbf{63.7670}	&\textbf{9.28}\\
 \bottomrule
\end{tabular}
}}
\end{center}
\end{table*}

\begin{table}[t]
 \caption{Comparison on NYUD-v2 under \textit{full-labels} settings. Our method achieves significantly better performance compared with SoTA multi-task learning works on all of the three tasks.}
\begin{center}
\setlength{\tabcolsep}{3.8mm}{\scalebox{0.8}{
\begin{tabular}{lcccc}
     \toprule
    \multirow{2}{*}{Model}&{Semseg.} &{Depth.} & {Normal.}&$\Delta_{M T L}$\\
     &\textit{mIoU}↑ &\textit{AbS}↓ &\textit{mErr}↓ &(\%)↑ \\ \midrule
    STL & 37.45	&	0.6079	&	25.94 & - \\
    MTL \textit{baseline} & 36.95	&0.5510		&29.51		&-1.91\\
    MTAN~\cite{liu2019end}& 39.39	&0.5696		&28.89		&0.03\\
    X-Task~\cite{zamir2020robust} &38.91		&0.5342		&29.94	&0.20\\
    Uncertainty~\cite{kendall2018multi} &36.46		&0.5376		&27.58	&0.87 \\
    GradNorm~\cite{chen2018gradnorm} &37.19		&0.5775		&28.51		&-1.87 \\
    MGDA~\cite{desideri2012multiple} &38.65		&0.5572		&28.89		&0.06 \\
    DWA~\cite{liu2019end} &36.46		&0.5429		&29.45		&-1.83 \\
    XTC~\cite{li2022learning} & 41.00	&0.5148		&28.58		&4.87 \\
    XTC+Uncertainty~\cite{kendall2018multi} &41.09	&0.5090		&26.78		&7.58 \\
    CCR~\cite{yang2023contrastive} & 43.09		&0.4894		&27.87		&9.04 \\ \midrule
    Ours (HiTTs) &44.32		&0.4813		&25.76		&13.29\\
    Ours (HiTTs+$\mathcal{L}_f$) &\textbf{45.47}		&\textbf{0.4763}		&\textbf{25.72}		&\textbf{14.64}\\
     \bottomrule
\end{tabular}
}}
\end{center}
\label{tab:aps}
\end{table}

\subsection{Evaluation Metrics}
We have briefly introduced our evaluation metrics for multiple dense prediction tasks in Sec.4.1. We provide a more detailed description as follows: (i) {mIoU}: mean intersection over union; (ii) {pAcc}: per-pixel accuracy; (iii) {AbS or AbR}: absolute error or absolute-relative error; (iv) {rmse}: root mean square error (for Normal. we calculate the mean square error of the predicted angles with the ground-truths); (v) {mErr}: mean of angle error; (vi) odsF: optimal dataset F-measure~\cite{martin2004learning}; (vii) {threshold}: for surface normal estimation, we calculate the proportion of pixels with angle error smaller than three thresholds $\eta \in\left\{11.25^{\circ}, 22.50^{\circ}, 30^{\circ}\right\}$.

Additionally, to better evaluate the proposed method, we also consider using $\Delta_{M T L}$ proposed by~\cite{vandenhende2021multi} to evaluate the overall improvement of the multi-task performances of all the tasks, which is defined as:
\vspace{2mm}
\begin{equation}
 \Delta_{M T L}= \sum_{N} (-1)^{l_{t}}\left(M_{m, t}-M_{s, t}\right) / M_{s, t}
  \label{eq:13a}
\end{equation}
where $l_{t}=1$ if a lower evaluation value indicates a better performance measurement of $M_{t}$ for task $t$, and $l_{t}=0$ if a higher value is better. Footnote $s$ and $m$ represent the performance of the single-task learning and the multi-task learning respectively. We will show experimental results with all of these metrics to further show the effectiveness of our method.

\begin{table}[t]
\caption{Comparison on Cityscapes under \textit{one-label} setting.}
\begin{center}
\setlength{\tabcolsep}{2.4mm}{\scalebox{0.8}{
\begin{tabular}{clccccc}
 \toprule
\multirow{2}*{Setting} & \multirow{2}*{Model}&\multicolumn{2}{c}{Semseg.} &\multicolumn{2}{c}{Depth.}  &$\Delta_{M T L}$\\
\cmidrule(lr){3-4}\cmidrule(lr){5-6}
 & &\textit{mIoU}↑ &\textit{pAcc}↑ &\textit{AbS}↓&\textit{rmse}↓&(\%)↑ \\ \midrule
\multirow{5}*{\rotatebox{90}{One-Label}} & STL & 69.69	&91.91	&0.0142	&0.0271& - \\
& MTL \textit{baseline} &69.94	&91.62	&0.0159	&0.0292	&-4.92\\
& MTAN~\cite{liu2019end} & 71.12	&92.35	&0.0146	&0.0278	&-0.72 \\
& XTC~\cite{li2022learning} & 73.23	&92.73	&0.0159	&0.0293	&-3.53\\
& Ours & \textbf{73.65}	&\textbf{92.81}	&\textbf{0.0135}	&\textbf{0.0265}	&\textbf{3.45}\\
 \bottomrule
\end{tabular}
}}
\end{center}
\label{tab:bs}
\end{table}

\begin{table*}[t]
\begin{center}
 \caption{Investigate the effectiveness of different components on NYUD-v2 testing set under \textit{one-label} setting.}
\vspace{-4mm}
\label{tab:ds}
\setlength{\tabcolsep}{4.4mm}{\scalebox{0.8}{
\begin{tabular}{lcccccccccc}
 \toprule
\multirow{2}{*}{Method}&\multicolumn{2}{c}{Semseg.} &\multicolumn{2}{c}{Depth.} & \multicolumn{5}{c}{Normal.}&$\Delta_{M T L}$\\
\cmidrule(lr){2-3}\cmidrule(lr){4-5}\cmidrule(lr){6-10}
 &\textit{mIoU}↑ &\textit{pAcc}↑ &\textit{AbS}↓&\textit{rmse}↓ &\textit{mErr}↓ &\textit{rmse}↓ & \textit{$\eta_{1}$}↑& \textit{$\eta_{2}$}↑& \textit{$\eta_{3}$}↑&(\%)↑ \\ \midrule
STL & 29.28	&55.41	&0.7182	&1.0151 	&30.1971 &37.7115 &23.1532 &46.4046 &58.5216& - \\
MTL \textit{baseline} & 30.92	&58.23	&0.5982	&0.8544	&31.8509 &38.6313 &19.7083 &41.2614 &53.6381	&0.11\\ \midrule
+HiTTs \textit{w/o.} OE &27.38	&55.08	&0.6049	&0.8626	&30.5904 &38.1046 &23.5233 &45.7012 &57.1902	&2.13\\
+HiTTs \textit{w/o.} Inter & 31.26	&57.99	&0.5966	&0.8592	&30.2911 &37.7714 &22.7802 &46.2590 &58.0880	&4.51\\
+HiTTs \textit{w/o.} Intra & 31.44	&58.50	&0.5910	&0.8533	&30.1432 &37.6719 &23.3251 &46.4354 &58.1508	&5.23\\
+HiTTs \textit{w/o.} $\boldsymbol{\theta_{i}}$ & 30.03 & 58.21 & 0.5823 & 0.8389 & 30.0005 & 37.3750 & 23.4160 & 46.1824 & 58.2909 & 5.08\\
+HiTTs \textit{w/o.} $\boldsymbol{\varphi_{i}}$ & 30.53 & 57.18 & 0.5842 & 0.8565 & 30.0891 & 37.4465 & 23.2297 & 45.9603 & 58.0509 & 4.60\\
+HiTTs &32.48	&59.61	&0.5844	&0.8382	&30.0847 &37.5827 &23.9975 &46.4790 &58.2146	&6.51 \\ \midrule
STL \textit{w.} $\mathcal{L}_p$ & 30.78	&58.94	&0.6693	&0.9362	&30.2420 &37.8601 &23.5830 &46.4743 &58.3739	&3.03\\
MTL \textit{w.} $\mathcal{L}_p$ &  33.59	&61.79	&0.5882	&0.8554	&29.8174 &36.9781 &23.2875 &45.8803 &58.1061	&6.89\\ \midrule
+HiTTs \textit{w.} $\mathcal{L}_f$ &33.24	&60.74 	&0.5708	&0.8200	&29.2227 &36.9305 &25.7968 &48.8173 &60.2608	&9.75\\
+HiTTs \textit{w.} $\mathcal{L}_p$ &35.22   &62.93  	&0.5613  &0.8014  &28.8852 &36.4316 &25.3873 &49.1251	&60.9806	&11.57\\
+HiTTs \textit{w.} $\mathcal{L}_p+\mathcal{L}_f$ &\textbf{35.81}	&\textbf{63.22}	&\textbf{0.5540}	&\textbf{0.7939}	&\textbf{28.5131}	&\textbf{36.1738}	&\textbf{26.4985}	&\textbf{50.2357}	&\textbf{61.8343}	&\textbf{13.23}\\
\bottomrule
\end{tabular}
}}
\end{center}
\end{table*}

\begin{table*}[t]
\begin{center}
 \caption{Investigate the performance on labeled and unlabeled data of NYUD-v2 training set under \textit{one-label} setting.}
 \vspace{-3mm}
\label{tab:fs}
\setlength{\tabcolsep}{3.9mm}{\scalebox{0.8}{
\begin{tabular}{lccccccccccc}
 \toprule
\multirow{2}{*}{Method}&\multicolumn{2}{c}{Supervision}&\multicolumn{2}{c}{Semseg.} &\multicolumn{2}{c}{Depth.} & \multicolumn{5}{c}{Normal.}\\
\cmidrule(lr){2-3}\cmidrule(lr){4-5}\cmidrule(lr){6-7}\cmidrule(lr){8-12}
 &\textit{GT}&\textit{Pseudo} &\textit{mIoU}↑ &\textit{pAcc}↑ &\textit{AbS}↓&\textit{rmse}↓ &\textit{mErr}↓ &\textit{rmse}↓ & \textit{$\eta_{1}$}↑& \textit{$\eta_{2}$}↑& \textit{$\eta_{3}$}↑ \\ \midrule
\multirow{2}{*}{MTL \textit{baseline}} & \Checkmark & \XSolidBrush & 89.00	&96.04	&0.2041	&0.3434	&25.9280	&32.0824	&25.8276	&52.0758	&65.4135 \\
& \XSolidBrush & \XSolidBrush & 34.31	&61.56	&0.5823	&0.8375	&31.7697	&38.5071	&19.5042	&41.2401	&53.8490 \\ \midrule
\multirow{2}{*}{HiTTs} & \Checkmark & \XSolidBrush & 86.04    &	95.00&	0.3016	&0.4625	&21.7911	&28.9368	&38.8125	&63.6280	&73.8326 \\
& \XSolidBrush & \XSolidBrush & 34.69	&61.48	&0.5699	&0.8319	&29.8920&	37.3331	&23.9788	&46.4809	&58.4659 \\ \midrule
\multirow{2}{*}{HiTTs \textit{w.} $\mathcal{L}_p+\mathcal{L}_f$} & \Checkmark & \XSolidBrush & 86.63	&94.89	&0.3173	&0.4770	&20.9220	&28.1260	&41.2846	&66.0712	&75.6904 \\
& \XSolidBrush & \Checkmark & \textbf{37.25}	&\textbf{63.72}	&\textbf{0.5563}	&\textbf{0.8074}	&\textbf{28.5169}	&\textbf{36.1536}	&\textbf{26.4528}	&\textbf{50.0344}	&\textbf{61.5778} \\
\bottomrule
\end{tabular}
}}
\end{center}
\end{table*}

\begin{table*}[t]
\begin{center}
 \caption{Investigate the cross-task learning effect on NYUD-v2 under the \textit{few-shot} setting.}
 \vspace{-4mm}
 \label{tab:gs}
\setlength{\tabcolsep}{5.3mm}{\scalebox{0.8}{
\begin{tabular}{lccccccccc}
 \toprule
\multirow{2}{*}{Method}&\multicolumn{2}{c}{Few-Shot-Semseg} &\multicolumn{2}{c}{Few-Shot-Depth} & \multicolumn{5}{c}{Few-Shot-Normal}\\
\cmidrule(lr){2-3}\cmidrule(lr){4-5}\cmidrule(lr){6-10}
 &\textit{mIoU}↑ &\textit{pAcc}↑ &\textit{AbS}↓&\textit{rmse}↓ &\textit{mErr}↓ &\textit{rmse}↓ & \textit{$\eta_{1}$}↑& \textit{$\eta_{2}$}↑& \textit{$\eta_{3}$}↑\\ \cmidrule(lr){1-1} \cmidrule(lr){2-3}\cmidrule(lr){4-5}\cmidrule(lr){6-10}
STL & 5.80	&26.06	&0.9533	&1.2907	&47.5281 &53.6422 &5.4343	&17.8888 &27.9915	\\
MTL \textit{baseline}& 16.75	&41.01	&0.9165	&1.2968	&40.0456 &\textbf{46.3370} &12.0348 &26.8520 &37.2863 \\
HiTTs &18.05	&44.69	&0.7803	&1.1272	&39.8508 &47.1113 &14.1108 &29.4719 &39.5924 \\
HiTTs \textit{w.} $\mathcal{L}_p+\mathcal{L}_f$& \textbf{20.07} 	&\textbf{45.62}	&\textbf{0.7366}	&\textbf{1.0206}	&\textbf{38.5029} &{46.9907} &\textbf{17.1082} &\textbf{34.7171} &\textbf{45.1765} \\
\bottomrule
\end{tabular}
}}
\end{center}
\end{table*}

\begin{table}[t]
 \caption{Comparison of HiTTs with \textit{Single-scale} (SS) and \textit{Multi-scale} (MS) Global Task Token Learning on PASCAL-Context under the \textit{one-label} and \textit{random-labels} setting.}
\begin{center}
\setlength{\tabcolsep}{1.5mm}{\scalebox{0.8}{
\begin{tabular}{clccccccc}
 \toprule
\multirow{2}{*}{Setting}&\multirow{2}*{Model}& Semseg. & Parsing. & Norm. & Sal. & Edge. &$\Delta_{M T L}$\\
&  &\textit{mIoU}↑  &\textit{mIoU}↑ & \textit{mErr}↓ & \textit{mIoU}↑   & \textit{odsF} ↑ &(\%)↑\\ \midrule
\multirow{4}*{\rotatebox{90}{One-} \rotatebox{90}{Label}}&STL & 47.7 &56.2 &16.0 &61.9 &64.0 & -\\
&MTL \textit{baseline} & 48.4 &55.1 &16.0 &61.6 &66.5 & 0.59\\
&HiTTs (SS) & 51.0	&54.7	&16.2	&61.7	&66.1 &1.19\\
&HiTTs (MS) &{52.3}	&56.2	&15.8	&62.0     &67.9 &3.43\\ \midrule
\multirow{4}*{\rotatebox{90}{Random-}  \rotatebox{90}{Labels}}&STL &  60.9 &55.3 &14.7 &64.8 &66.8& -\\
&MTL \textit{baseline} &58.4		&55.3		&16.0		&63.9	&67.8 & -2.57\\
&HiTTs (SS) &59.1		&53.4		&15.0		&64.1		&67.8 & -1.60\\
&HiTTs (MS) &60.3		&55.3		&14.7		&64.6		&70.2 & 0.76\\
 \bottomrule
\end{tabular}
}}
\end{center}
\label{tab:hs}
\end{table}

\subsection{More Quantitative Results}

\subsubsection{State-of-the-art Comparison}
\noindent \\ \textbf{Comparison on NYUD-v2.}
We compare our method with~\cite{li2022learning,liu2019end} on NYUD-v2 under both the one-label and random-labels settings, and the quantitative results are shown in Table~\ref{tab:as}. XTC~\cite{li2022learning} is the first work designed for partially annotated multi-task dense prediction. MTAN~\cite{li2022learning} is an attention-based MTL network designed for the fully supervised setting, and we train it with our setup. The quantitative results show that our method surpasses them by a large margin on all the metrics of the three tasks. More specifically, ours achieves $+9.63\%~\Delta_{MTL}$ and $+8.62\%~\Delta_{MTL}$ compared with~\cite{li2022learning} under the two partial-label settings, respectively.

Our HiTTs can also be applied on full-labels setting, which utilizes cross-task relations and task-token interactions to fertilize the multi-task learning process. We compare with some of the recent works, including multi-task interaction works like MTAN~\cite{liu2019end}, X-Task~\cite{zamir2020robust} and CCR~\cite{yang2023contrastive}; and multi-task loss weighting strategies like Uncertainty~\cite{kendall2018multi}, GradNorm~\cite{chen2018gradnorm}, MGDA~\cite{desideri2012multiple} and DWA~\cite{liu2019end}. As shown in Table~\ref{tab:aps}, our method still clearly surpasses all of the SOTA works ($13.29\%~\Delta_{MTL}$ overall), indicating the effective cross-task interaction brought by HiTTs. Additionally, with the aid of feature-level supervision loss $\mathcal{L}_f$, which is supported by global task tokens, our method can achieve $14.64\%~\Delta_{MTL}$ overall on the three tasks.

\medskip
\noindent \textbf{Comparison on Cityscapes.}
We also compare our results with~\cite{li2022learning,liu2019end} on Cityscapes, under the \textit{one-label} setting with both Semseg. and Depth. tasks. As shown in Table~\ref{tab:bs}, our method achieves SOTA performance on both tasks, and significantly better performance on Depth, resulting in an average gain of $+6.98\%$ in terms of $\Delta_{MTL}$. Additionally, it’s worth mentioning that our work is the only one that achieves balanced performance gain on both tasks compared with STL.

\subsubsection{Model Analysis}
\noindent \\ \textbf{Effect of Hierarchical Task Tokens.}
As shown in Table~\ref{tab:ds}, under the {one-label} setting on NYUD-v2, we give a more-detailed ablation study on the key components of hierarchical task tokens (HiTTs), including both hierarchies of the task tokens, and the inter-task (Inter) and intra-task (Intra) learning of global task tokens, and orthogonal embeddings (OE) of fine-grained task tokens, and two types of pseudo supervision losses ($\mathcal{L}_p$ for pseudo label loss and $\mathcal{L}_f$ for feature supervision loss) on different hierarchy. The quantitative results clearly show that every component contributes to the multi-task performance on all the metrics and on all the tasks over the MTL baseline. HiTTs boost the model performance by conducting cross-task interaction and encouraging high-confidence predictions, and leads to an overall $+9.64\%~\Delta_{MTL}$ on all tasks compared with baseline. However, for the learning process of HiTTs, the orthogonal embeddings (OE) are essential for generating representative fine-grained task tokens, and without OE, the performance will significantly drop ($-4.38\%~\Delta_{MTL}$), especially on Semseg. which requires more discriminative category information. The inter-task and intra-task learning processes are also important since without either of them, the learning of cross-task relations and task representations will be affected, resulting in $-2.00\%~\Delta_{MTL}$ and $-1.28\%~\Delta_{MTL}$ performance after removing them respectively from the learning process of HiTTs.

\medskip
\par\noindent\textbf{Effect of Hierarchical Feature Supervision and Label Discovery.} We analyze the contributions from both feature-level and prediction-level supervision with all of the metrics, as shown in Table~\ref{tab:ds}, both methods boost multi-task performance, and $\mathcal{L}_p$ contributes more since fine-grained task tokens contain more specific and discriminative task information and are directly involved in the formation process of task predictions. The combination of both methods achieves better performance than applying them separately, which validates the importance of consistently discovering supervision signals in both hierarchies. Comparing $\mathcal{L}_p$ imposed on different models, including STL, MTL baselines, and our HiTTs, our token-based pseudo-label discovery is much better. Since MTL shares a backbone that learns stronger representations on all tasks, MTL produces pseudo labels with better quality and surpasses STL a lot in performance ($+3.86\%~\Delta_{MTL}$). Our HiTTs perform i) consistent label discovery in both feature and prediction space; ii) effective cross-task feature-token interactions, which furthermore enhance the quality of pseudo labels, and bring extra $+4.68\%~\Delta_{MTL}$ overall. 

We also study the performance of our method on the labeled and unlabeled data separately on NYUD-v2 training set under the one-label setting. As shown in Table~\ref{tab:fs}, for data without labels, the model with HiTTs generalizes better on them, especially on Depth. and Normal, and adding hierarchical supervision will more significantly boost the performance of unlabeled data.

\medskip
\par\noindent\textbf{Effect of Cross-Task Interactions.} To further show the effect of cross-task learning brought by HiTTs, we develop new \textbf{few-shot} settings, under which one task has only a few labels while other tasks are fully labeled. We apply this setting respectively on the three tasks of NYUD-v2, namely \textbf{few-shot-semseg},  \textbf{few-shot-depth} and  \textbf{few-shot-normal}. For each few-shot task, we have 10 shots for the model to learn. As shown in Table~\ref{tab:gs}, we only show the performance of the few-shot tasks in the table, and due to the lack of label supervision, the STL performs poorly on each few-shot task: $5.80$ \textit{mIoU} on Semseg, $0.9633$ \textit{AbS} on Depth, and $47.5281$ \textit{mErr} on Normal. Benefiting from the sharing backbone, MTL \textit{baseline} performs much better, since the backbone can be fully supervised on the other two tasks, and gain stronger representations from other tasks. With the aid of HiTTs, the multi-task model can achieve an extra performance gain, since the cross-task interactions brought by intra-task learning can fertilize the unlabeled tasks in the decoding stage, which introduces more task-relevant information and discriminative representations to task features without label supervision. The gain brought by HiTTs is $+7.76 \%$ on Semseg, $+14.86 \%$ on Depth, and $+0.49 \%$ on Normal respectively. Additionally, if we add the pseudo supervision signals to aid the learning process, the performance will be further improved: $+19.82 \%$ on Semseg, $+19.63 \%$ on Depth, and $+3.85 \%$ on Normal compared with MTL \textit{baseline}.

\medskip
\par\noindent\textbf{Analysis of Multi-Scale Global Task Token Learning.} As we illustrated in Sec.~\ref{sec:MS-GTTL}, we adopt multi-scale global task learning with ResNet-18 backbone on PASCAL-Context. To validate the effectiveness of learning global task tokens on multi-scale features, we compare the performance of Global Task Token Learning with single-scale and multi-scale backbone features respectively in Table~\ref{tab:hs}. As shown in the table, HiTTs with single-scale (SS) Global Token Learning surpass the MTL \textit{baseline} on both one-label and random-labels settings, with overall $+0.60\%~\Delta_{MTL}$ and $+0.97\%~\Delta_{MTL}$ on all tasks respectively, and the multi-scale (MS) Global Token Learning further enhances the performance to $+2.84\%~\Delta_{MTL}$ and $+3.33\%~\Delta_{MTL}$ on all tasks, which indicates the effectiveness of applying global token learning on multi-scale features.

\medskip
\par\noindent\textbf{Analysis of Threshold Hyperparameter $\tau_i$.} We discuss the effect of the confidence threshold hyperparameter, $\tau_i$. This threshold is chosen to ensure that high-confidence pixel predictions are masked out to serve as pseudo-labels. To analyze the sensitivity of our method to this hyperparameter, we conduct an ablation study on the NYUD-v2 dataset under the \textit{one-label} setting. As shown in Table~\ref{tab:is}, we vary $\tau_i$ across a wide range from $0.3$ to $0.95$. The results demonstrate that the performance on all tasks remains remarkably stable with only minor fluctuations. Furthermore, the overall multi-task learning gain ($\Delta_{MTL}$) is consistently and significantly positive across all tested values. This demonstrates that our method is not sensitive to the choice of $\tau_i$, showcasing its robustness.

\medskip

\par\noindent\textbf{Comparison under DiffusionMTL~\cite{ye2024diffusionmtl} settings on NYUD-v2.}
Since DiffusionMTL~\cite{ye2024diffusionmtl} adopts different experimental settings from our main experiments, we re-implement our method under the specific setup of DiffusionMTL, using a ResNet-18 backbone to ensure a fair comparison. The quantitative results on NYUD-v2 under both the \textit{one-label} and \textit{random-labels} settings are shown in Table~\ref{tab:js}. As the results indicate, our method significantly outperforms DiffusionMTL across all metrics in the \textit{one-label} setting. More specifically, compared to the best-performing DiffusionMTL variant, our method achieves a $+6.26\%$ higher $\Delta_m$ in the \textit{one-label} setting and a $+2.09\%$ higher $\Delta_m$ in the \textit{random-labels} setting. This demonstrates the effectiveness and superior performance of our approach when compared directly with other methods.

\medskip

\par\noindent\textbf{Implementations with Different Backbones.}
To demonstrate that our method can be flexibly implemented on different image backbones, we perform additional experiments implementing our HiTTs with a Vision Transformer (ViT~\cite{dosovitskiy2020image}) backbone. The results on NYUD-v2 are presented in Table~\ref{tab:ks}. The quantitative results clearly show that when equipped with the more powerful ViT-base backbone, our method achieves a substantial performance improvement across all tasks under both the \textit{one-label} and \textit{random-labels} settings. For instance, in the \textit{random-labels} setting, using ViT-base boosts the Semantic Segmentation mIoU from 47.73 to 61.88. This not only confirms the flexibility of our approach but also highlights its potential to achieve even greater performance when paired with more advanced backbone architectures.

\medskip

\par\noindent\textbf{Analysis of Incorporating Dense FixMatch~\cite{pieropan2022dense} on NYUD-v2.}
Our method can also incorporate general semi-supervised dense prediction strategies, e.g. Dense FixMatch~\cite{pieropan2022dense}. The quantitative results on the NYUD-v2 \textit{one-label} setting are shown in Table~\ref{tab:ls}. The results indicate that Dense FixMatch provides consistent performance improvements. This demonstrates the efficacy of Dense FixMatch as a versatile component for enhancing various dense prediction tasks in a semi-supervised context.

\begin{table}[t]
\caption{Ablation study on the hyperparameter $\tau_i$.}
\vspace{-2mm}
\label{tab:is}
\begin{center}
\setlength{\tabcolsep}{5mm}{\scalebox{0.8}{
\begin{tabular}{ccccc}
\toprule
\multirow{2}{*}{$\tau_i$} & Semseg. & Depth. & Normal. & $\Delta_{MTL}$ \\
& \textit{mIoU}↑ & \textit{mErr}↓ & \textit{mErr}↓ & (\%)↑ \\ \midrule
0.95 & 46.67 & 0.4585 & 25.48 & 3.11 \\
0.90 & 47.10 & 0.4573 & 25.43 & 3.57 \\
0.70 & 46.89 & 0.4621 & 25.36 & 3.17 \\
0.50 & 47.00 & 0.4690 & 25.35 & 2.79 \\
0.30 & 47.11 & 0.4706 & 25.33 & 2.78 \\
\bottomrule
\end{tabular}
}}
\end{center}
\end{table}

\begin{table}[t]
\caption{Comparisons with incorporating general semi-supervised dense prediction method~\cite{pieropan2022dense} on NYUD-v2 \textit{one-label} setting.}
\label{tab:ls}
\begin{center}
\setlength{\tabcolsep}{6mm}{\scalebox{0.8}{
\begin{tabular}{cccc}
\toprule
\multirow{2}{*}{Method} & Semseg. & Depth. & Normal. \\
& \textit{mIoU}↑ & \textit{mErr}↓ & \textit{mErr}↓ \\ \midrule
Ours & 47.30 & 0.4539 & 25.40 \\
Ours+\cite{pieropan2022dense} & 47.72 & 0.4502 & 25.33 \\
\bottomrule
\end{tabular}
}}
\end{center}
\end{table}

\begin{table}[t]
\caption{Comparison on NYUD-v2 under the settings of DiffusionMTL~\cite{ye2024diffusionmtl}. (P) and (F) represent the Prediction Diffusion and Feature Diffusion modes for DiffusionMTL~\cite{ye2024diffusionmtl}. ``*'' denotes the re-implemented results to align the settings. Our method performance outperforms previous methods.}
\label{tab:js}
\begin{center}
\setlength{\tabcolsep}{2.4mm}{\scalebox{0.8}{
\begin{tabular}{clcccc}
\toprule
\multirow{2}{*}{Setting} & \multirow{2}{*}{Model} & {Semseg.} & {Depth.} & {Normal.} & {$\Delta_m$} \\
 & & \textit{mIoU} $\uparrow$ & \textit{absErr} $\downarrow$ & \textit{mErr} $\downarrow$ & (\%) $\uparrow$ \\ \midrule
\multirow{8}{*}{\rotatebox{90}{One-Label}} 
 & STL & 45.28 & 0.4802 & 25.93 & - \\
 & MTL \textit{baseline} & 43.92 & 0.5138 & 26.44 & -3.99 \\
 & SS~\cite{li2022learning} & 27.52 & 0.6499 & 33.58 & - \\
 & XTC~\cite{li2022learning} & 30.36 & 0.6088 & 32.08 & - \\
 & XTC*~\cite{li2022learning} & 43.97 & 0.5140 & 26.30 & -3.79 \\
 & DiffusionMTL (P)~\cite{ye2024diffusionmtl} & 44.97 & 0.5137 & 26.17 & -2.86 \\
 & DiffusionMTL (F)~\cite{ye2024diffusionmtl} & 44.47 & 0.5059 & 25.84 & -2.27 \\
 & Ours* & \textbf{47.30} & \textbf{0.4539} & \textbf{25.40} & \textbf{3.99} \\ \midrule
\multirow{8}{*}{\rotatebox{90}{Random-Labels}} 
 & STL & 48.25 & 0.4792 & 24.65 & - \\
 & MTL \textit{baseline} & 45.93 & 0.4839 & 25.53 & -3.12 \\
 & SS~\cite{li2022learning} & 29.50 & 0.6224 & 33.31 & - \\
 & XTC~\cite{li2022learning} & 34.26 & 0.5787 & 31.06 & - \\
 & XTC*~\cite{li2022learning} & 46.03 & 0.4811 & 25.97 & -3.44 \\
 & DiffusionMTL (P)~\cite{ye2024diffusionmtl} & 47.44 & 0.4803 & 25.26 & -1.45 \\
 & DiffusionMTL (F)~\cite{ye2024diffusionmtl} & 46.82 & 0.4743 & \textbf{24.75} & -0.77 \\
 & Ours* & \textbf{47.73} & \textbf{0.4510} & 24.86 & \textbf{1.32} \\
\bottomrule
\end{tabular}
}}
\end{center}
\end{table}

\begin{table}[t]
\caption{Ablation study of our method with different backbones on NYUD-v2. The results show that our method is flexible and achieves significant gains when paired with a more powerful ViT backbone.}
\label{tab:ks}
\begin{center}
\setlength{\tabcolsep}{3.8mm}{\scalebox{0.8}{
\begin{tabular}{clccc}
\toprule
\multirow{2}{*}{Setting} & \multirow{2}{*}{Backbone} & {Semseg.} & {Depth.} & {Normal.} \\
 & & \textit{mIoU} $\uparrow$ & \textit{absErr} $\downarrow$ & \textit{mErr} $\downarrow$ \\ \midrule
\multirow{2}{*}{One-Label}
 & ResNet-18 & 47.30 & 0.4539 & 25.40 \\
 & ViT-base & \textbf{58.38} & \textbf{0.3740} & \textbf{23.65} \\ \midrule
\multirow{2}{*}{Random-Labels} 
 & ResNet-18 & 47.73 & 0.4510 & 24.86 \\
 & ViT-base & \textbf{61.88} & \textbf{0.3979} & \textbf{23.03} \\
\bottomrule
\end{tabular}
}}
\end{center}
\end{table}

\begin{figure}[t]
  \centering
  \includegraphics[width=1.0\linewidth]{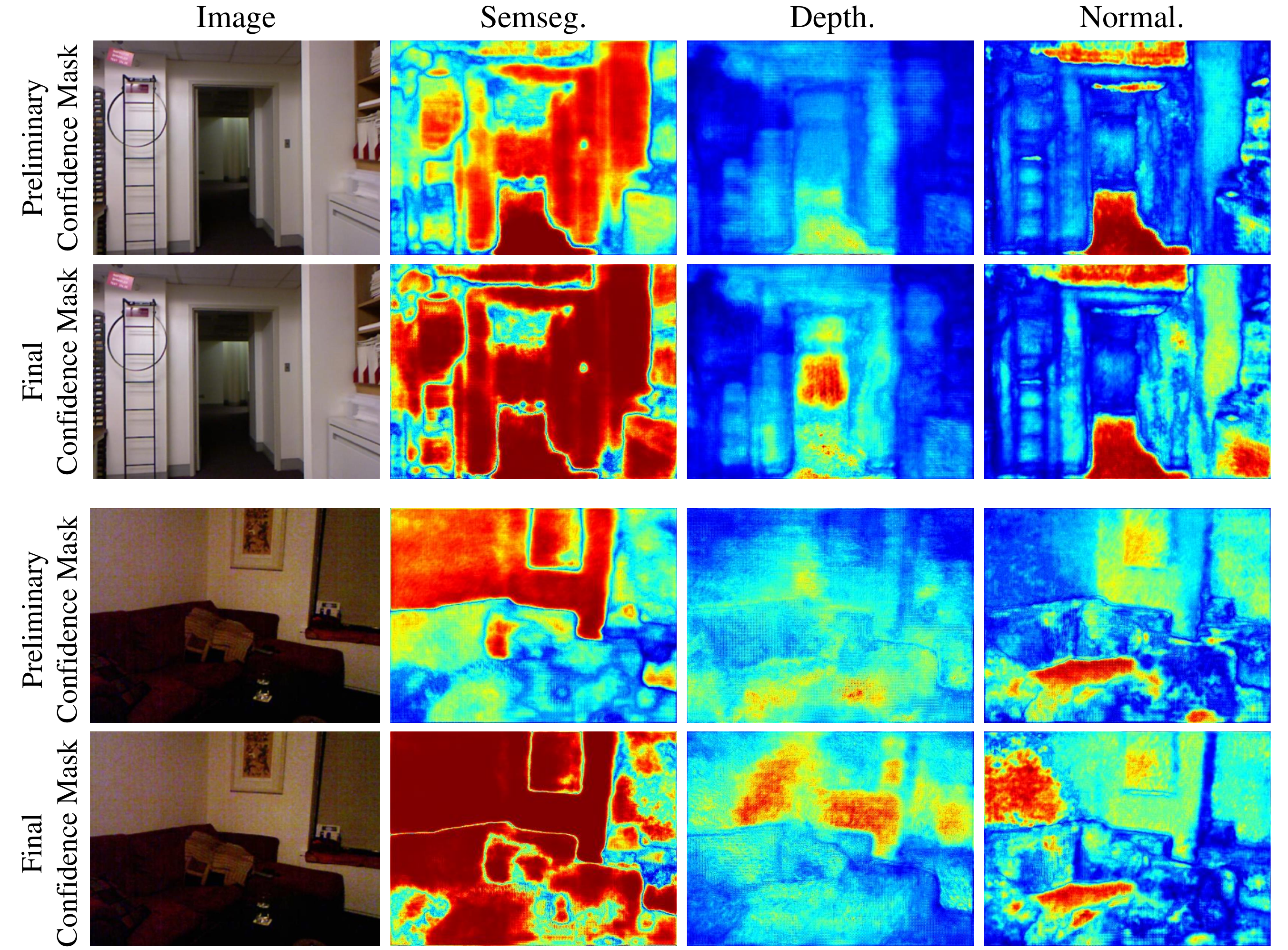}
  \vspace{-7mm}
  \caption{Comparison of the task confidence map before and after refined by the fine-grained task tokens, which greatly encourage high-confidence predictions on all tasks (red color represents high-confidence areas). For noisy data like the second photo taken in a dark environment, this enhancement are more significant.}
  \vspace{-5mm}
  \label{fig:comFG}
\end{figure}

\subsection{More Qualitative Results}
We provide more qualitative results mainly from four parts: more visualization of the fine-grained token distributions, more comparisons of task score maps produced by HiTTs, more qualitative prediction comparisons, and the quality of generated pseudo labels.

\medskip
\noindent \textbf{Role of Fine-grained Task Tokens.} To illustrate the role of our fine-grained task tokens, we visualize task confidence maps before and after the refinement process in Fig.~\ref{fig:comFG}. The fine-grained tokens significantly enhance prediction confidence across all tasks, as indicated by the expansion of high-confidence areas (represented in red). This enhancement is particularly evident in challenging scenarios, such as the noisy image captured in a dark environment (second row), where the refined map shows a marked improvement in clarity and confidence.

\medskip
\noindent \textbf{Visualization of token distributions.} We also provide visualization analysis to show the distributions of fine-grained task tokens on Cityscapes. In Fig.~\ref{fig:comOE_all}, with the aid of OE, the self-correlation map of tokens will be more diagonal, and the distributions after PCA have better clusters in 3-dimensional feature space.

\medskip
\noindent \textbf{Comparisons of Task Score Maps.} We visualize score maps produced by global task tokens and fine-grained task tokens respectively on each task. As shown in Fig.~\ref{fig:comp_score}, we conduct visualization on both NYUD-v2 and Cityscapes datasets. The score maps indicate the response of task features to task tokens, and the response patterns of feature maps on different tasks are very different, e.g. Semseg. features highlight areas with distinguish semantics, Depth. features focus on areas with a certain depth range and Normal. features focus on surfaces with the same orientation.

Comparing the score maps produced by tokens from different hierarchies, we find that score maps produced by global task tokens are relatively rough and noisy, while those generated by fine-grained task tokens have finer granularity and less noise, which shows the hierarchy of the HiTTs learning process. Also, we observed that the high-light areas of global task tokens are monotonous, while fine-grained task tokens can highlight more details. This phenomenon is clearly observed in Cityscapes, since the ground truths of this dataset follow the long-tail distribution, thus the global task-tokens tend to learn the category with more pixel samples, and consequently always highlight the {road} area as shown in Fig.~\ref{fig:comp_score}. However, the fine-grained tokens can give attention to more details, including the vehicles and pedestrians with fewer pixel samples. Thus, it is necessary to design a hierarchical structure for tokens to learn representations with different granularity.

\medskip
\noindent \textbf{Qualitative prediction comparisons with SOTA works.} We additionally provide comparisons with SOTA works on NYUD-v2 and Cityscapes. As shown in Fig.~\ref{fig:comSOTA_more}, we compare with MTL \textit{baseline}, XTC~\cite{li2022learning} on NYUD-v2 three tasks, and with MTL \textit{baseline}, MTAN~\cite{liu2019end}, XTC~\cite{li2022learning} on Cityscapes two tasks. Our method shows clearly better performance in semantic understanding and accurate geometry estimations (including depth and normal estimation), indicating the effectiveness of our method.

\medskip
\noindent \textbf{Visualization of Pseudo Labels.} In Fig.~\ref{fig:vis_plabel}, we show pseudo task label maps generated by fine-grained task tokens. The pseudo label on different tasks has good quality without ground-truth supervision, which proves the effective cross-task learning and strong generalization ability brought by HiTTs.

\begin{figure*}[!t]
  \centering
  \includegraphics[width=0.95\linewidth]{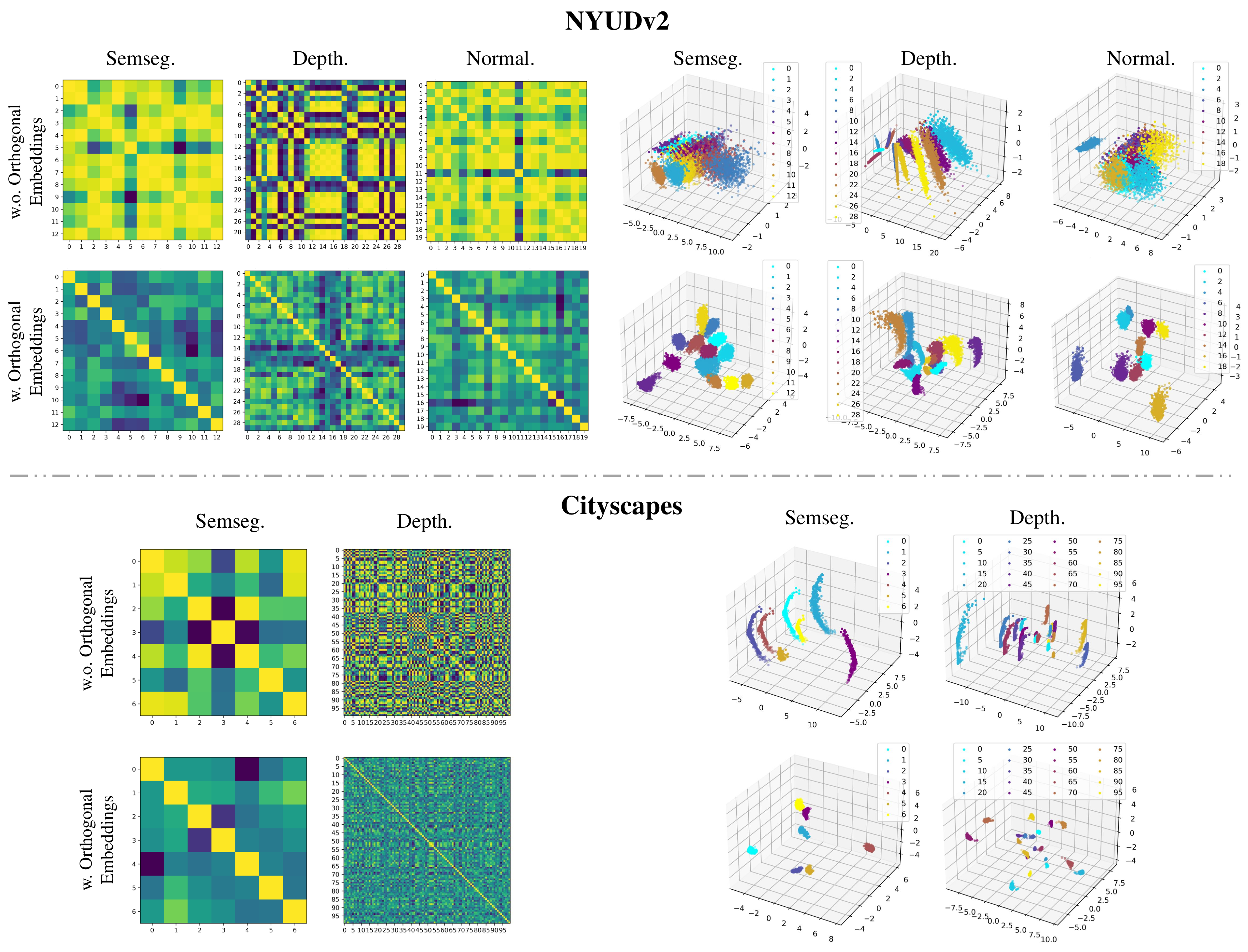}
  \vspace{-2mm}
  \caption{Visualization of self-affinities heatmap (\textit{left}) and PCA for distributions (\textit{right}) of fine-grained task tokens of the two tasks on NYUDv2 and Cityscapes validation sets. With orthogonal embeddings, the affinities between different tokens are low and the clustering of token distributions on each category is better.}
  \label{fig:comOE_all}
\end{figure*}

\begin{figure*}[!t]
  \centering
  \includegraphics[width=0.98\linewidth]{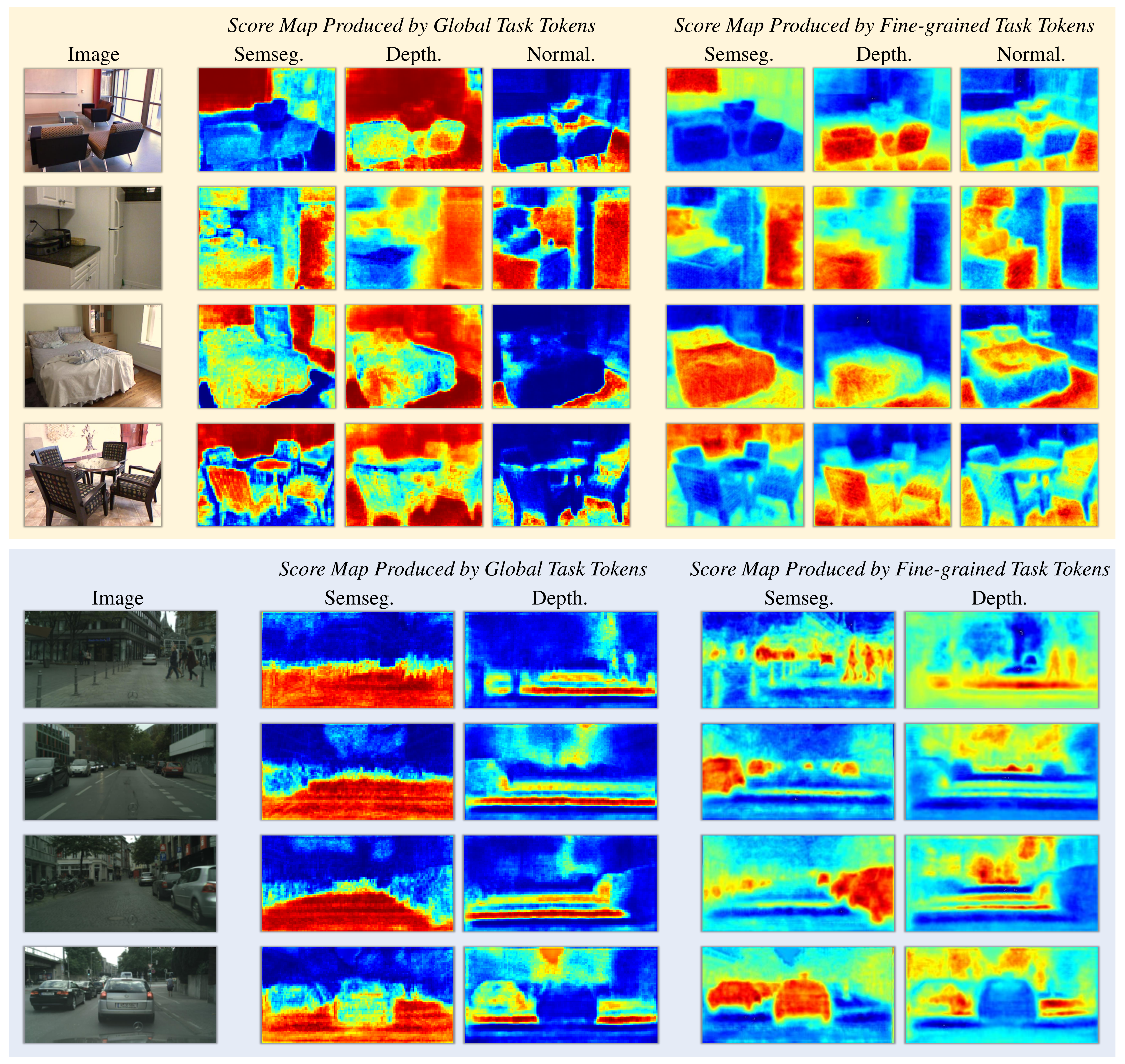}
    \vspace{-2mm}
  \caption{Comparisons of task score maps produced by global task tokens and fine-grained task tokens. The upper part is the visualization of samples on NYUD-v2 while the lower part is on Cityscapes.} 
  \label{fig:comp_score}
\end{figure*}

\begin{figure*}[!t]
  \centering
  \includegraphics[width=0.98\linewidth]{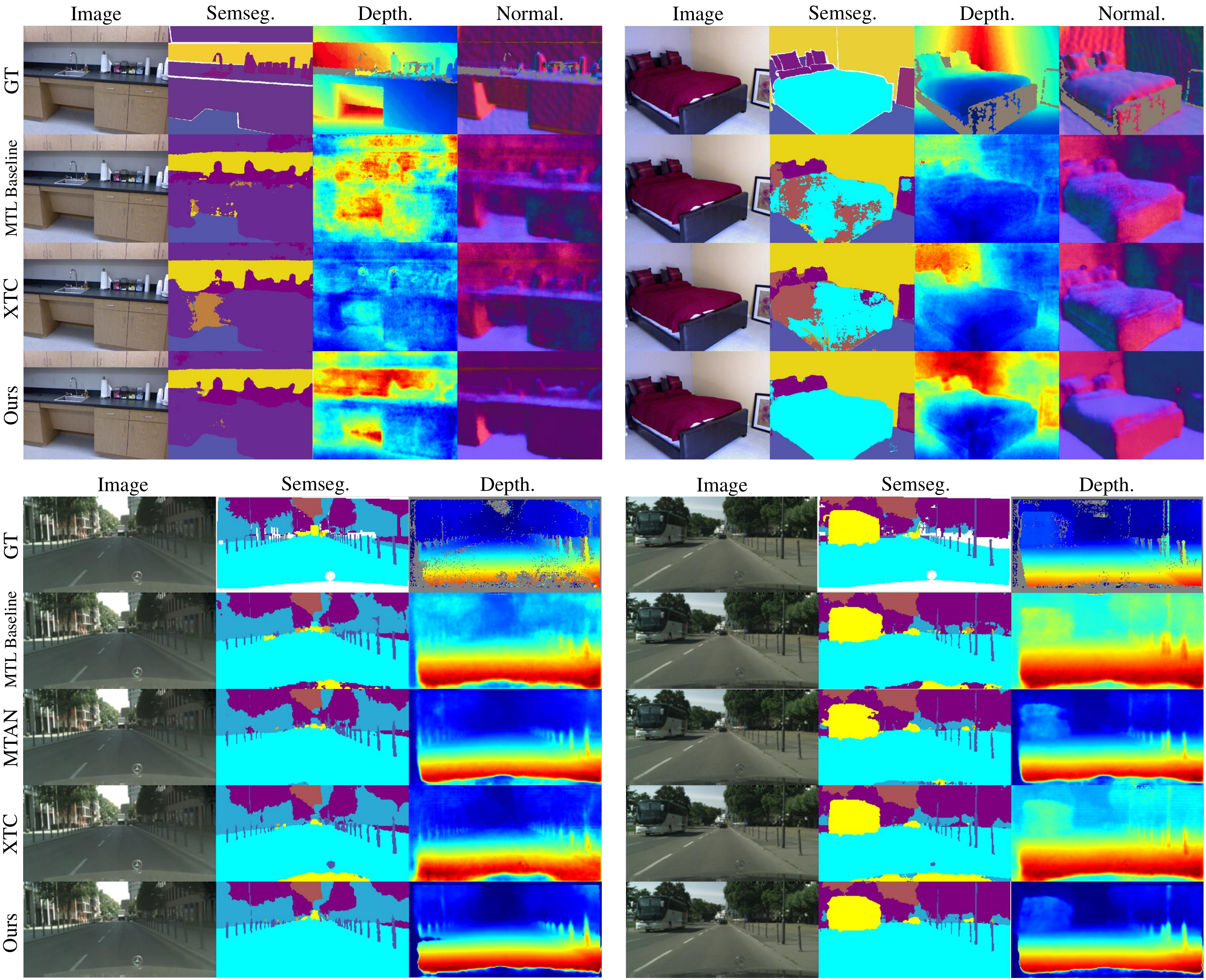}
    \vspace{-2mm}
  \caption{Comparisons with SOTA works on NYUD-v2 (upper part) and Cityscapes (lower part).}
  \label{fig:comSOTA_more}
\end{figure*}

\begin{figure*}[!t]
  \centering
  \includegraphics[width=0.95\linewidth]{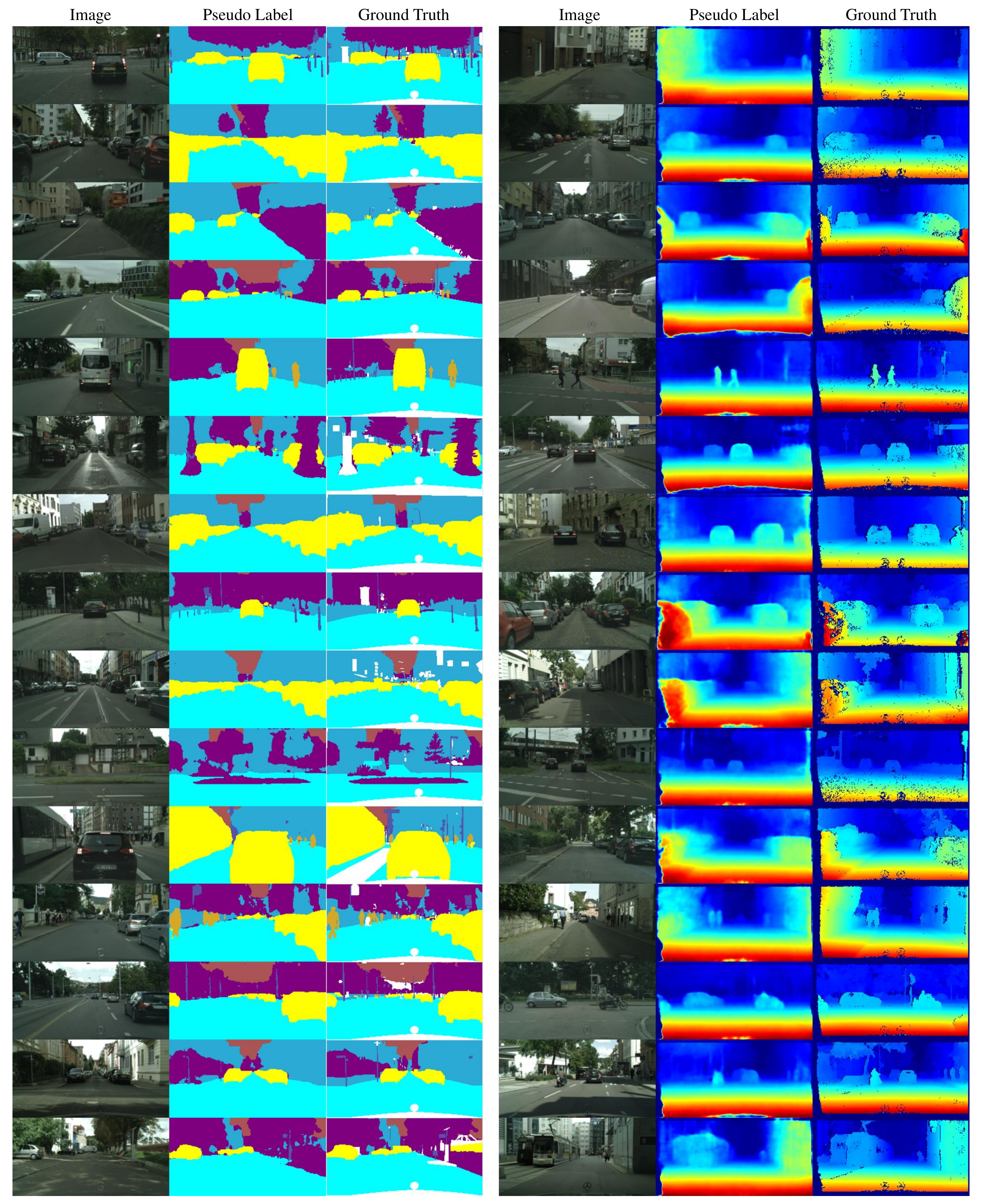}
      \vspace{-2mm}
  \caption{Quantitative analysis of the quality of pseudo labels generated by gloabl task tokens.}
  \label{fig:vis_plabel}
\end{figure*}

\end{document}